\documentclass[12pt]{article}
\usepackage{graphicx}
\usepackage[dvipsnames]{xcolor}
\usepackage{geometry}
\usepackage{subcaption}
\usepackage{booktabs}
\usepackage{tablefootnote}
\usepackage{multirow}
\usepackage{amsmath}
\usepackage{float}
\usepackage{hyperref}
\usepackage{xurl}
\hypersetup{
  colorlinks   = true, 
  urlcolor     = blue, 
  linkcolor    = blue, 
  citecolor   = black 
}
\usepackage[utf8]{inputenc}
\usepackage{natbib}
\usepackage{cleveref}
\usepackage{siunitx}
\usepackage{setspace}
\usepackage{lineno}
\usepackage{wrapfig}
\title{Domain-informed explainable boosting machines for trustworthy lateral spread predictions}

\author{Cheng-Hsi Hsiao, Krishna Kumar, Ellen M. Rathje}
\date{}

\begin{document}

\maketitle

\doublespacing

\begin{abstract}
Explainable Boosting Machines (EBMs) provide transparent predictions through additive shape functions, enabling direct inspection of feature contributions. However, EBMs can learn non-physical relationships that reduce their reliability in natural hazard applications. This study presents a domain-informed framework to improve the physical consistency of EBMs for lateral spreading prediction. Our approach modifies learned shape functions based on domain knowledge. These modifications correct non-physical behavior while maintaining data-driven patterns. We apply the method to the 2011 Christchurch earthquake dataset and correct non-physical trends observed in the original EBM. The resulting model produces more physically consistent global and local explanations, with an acceptable tradeoff in accuracy (4–5\%).
\end{abstract}

\section{Introduction}
Machine learning (ML) models are increasingly used to predict earthquake-induced hazards, including liquefaction triggering~\citep{rateria-2022,geyin_ai_2022,demir-2022} and lateral spreading~\citep{hsiao2024explainable,durante-2021}.
These models can capture complex, non-linear relationships among input features that traditional empirical methods struggle to represent.
Ensemble methods like random forests~\citep{breiman-2001} and gradient boosting~\citep{Chen_2016} capture complex ground responses by aggregating weak predictors that fit local data patterns, collectively covering the entire data distribution.
However, this aggregation obscures the isolated contribution of each feature, limiting physical interpretability.

Without interpretability, we cannot identify when models trained on biased or sparse data learn non-physical relationships.
\citet{hsiao2024explainable} developed an XGBoost model that predicted the probability of liquefaction-induced lateral spreading decreased with increasing PGA, inverting the physical expectation that stronger shaking is more likely to generate lateral spreading.
Sparse or imbalanced training data enables such artifacts, allowing statistically plausible but physically incorrect patterns to emerge.

To identify such artifacts, researchers use post-hoc methods like SHAP~\citep{lundberg-2017} and LIME~\citep{ribeiro-2016}.
These methods quantify feature contributions and their directional effects by analyzing input-output relationships, rather than inspecting the model's internal structure~\citep{can-2021,hsiao2024explainable}.
Such approximations can diverge from the model's actual logic~\citep{huang_failings_2024}.
Post-hoc methods can reveal non-physical relationships but cannot correct them.

Correcting non-physical relationships requires models whose decision logic is directly accessible and editable.
The Explainable Boosting Machine (EBM;~\cite{nori-2019}), a generalized additive model~\citep{hastie-1987}, represents predictions as a sum of univariate shape functions and pairwise interaction terms.
Unlike post-hoc explanations, these functions directly reveal the model's decision logic, showing how each feature contributes to predictions.
Non-physical patterns in these functions can be directly identified and modified.
EBM discloses the decision logic quite simply without sacrificing accuracy, performing comparably to black-box ensembles~\citep{maxwell-2021}.

We propose a framework to embed domain knowledge into trained EBM models through targeted modification of shape functions.
For univariate functions, we fit monotonic curves to regions where the learned relationship is physically plausible, replacing segments that contradict domain understanding.
For bivariate interaction functions, we synthesize alternatives from the modified univariate functions and selectively replace regions that deviate significantly from physical expectations.
This approach preserves data-driven patterns where they are trustworthy while correcting non-physical artifacts.

We apply this framework to lateral spreading prediction using the 2011 Christchurch earthquake dataset~\citep{LateralSpreadingDataset}.
The trained EBM exhibits non-physical behavior in specific input regions—for example, assigning reduced lateral spreading likelihood to sites with very shallow groundwater.
We demonstrate that these artifacts arise from data imbalance and can be corrected through our domain-informed modifications.
The resulting model sacrifices modest predictive accuracy but produces physically consistent explanations across all input ranges.

This study addresses two research questions: How can domain knowledge be systematically incorporated into a trained EBM to correct non-physical learned relationships?
What trade-offs emerge between predictive accuracy and physical consistency when imposing such constraints?

The remainder of this paper is organized as follows.
We first describe the Christchurch lateral spreading dataset.
\Cref{sect:meth} presents the EBM methodology and our framework for domain-informed modification.
\Cref{sect:results} evaluates the original and domain-informed models through performance metrics, global feature importance, and local explanations.
Finally, \Cref{sect:conclusion} discusses implications, limitations, and future research directions.

\section{Dataset Description: 2011 Christchurch Earthquake Lateral Spreading}

\begin{figure}[h]
    \centering
    \includegraphics[width=\textwidth]{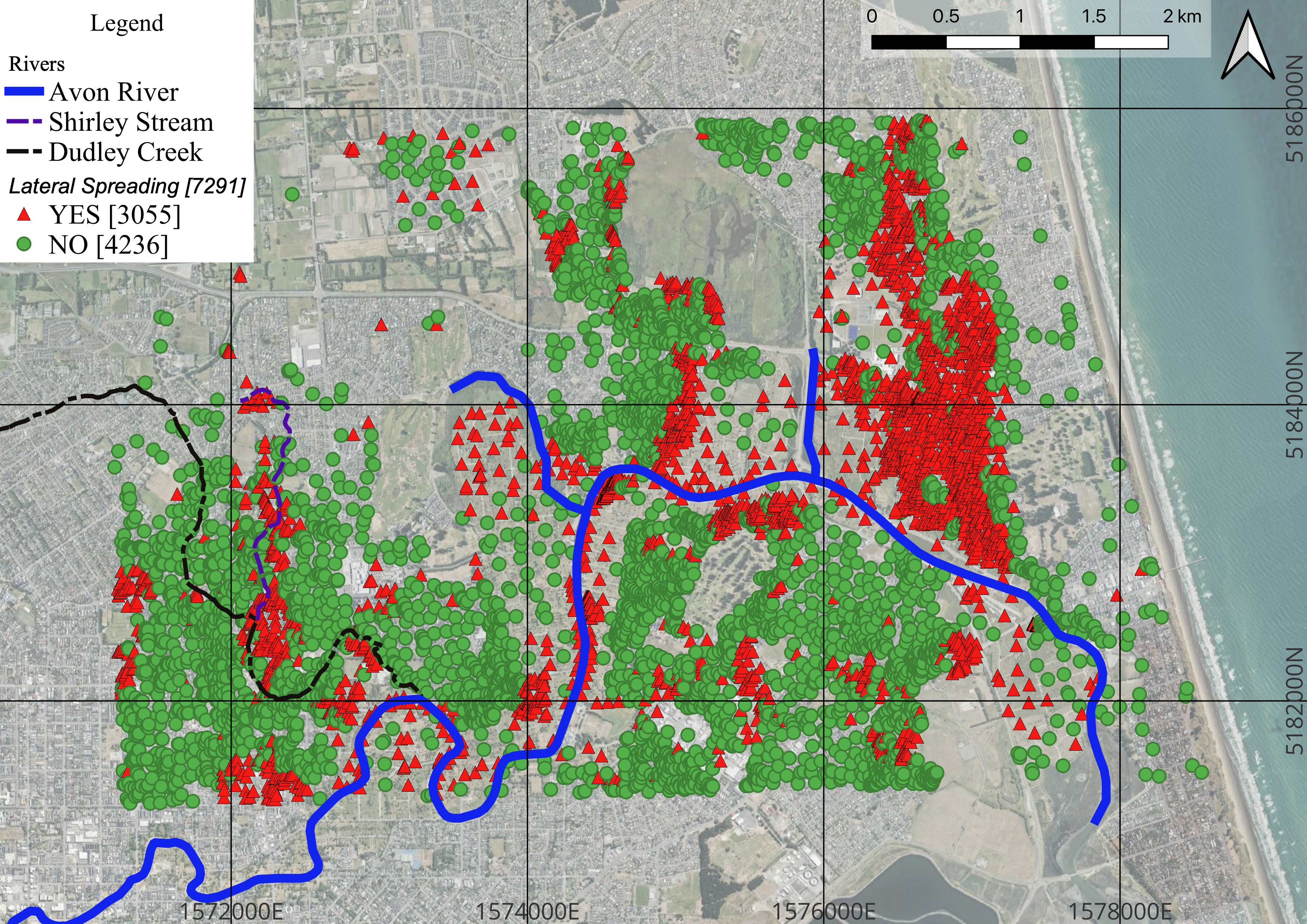}
    \caption{Spatial distribution of 2011 Christchurch earthquake showing lateral spreading Yes/No sites.}
    \label{fig:LS-map}
\end{figure}

We train and evaluate our EBM model using the 2011 Christchurch earthquake lateral spreading dataset compiled by~\citet{LateralSpreadingDataset}. 
This dataset includes 3,055 sites that exhibited lateral spreading and 4,236 sites with no lateral spreading, classified based on horizontal surface displacements measured through remote sensing image correlation~\citep{rathje-2017}.
Sites with displacements under 0.3 m are labeled as non-lateral spreading, while sites above this threshold are labeled as lateral spreading instances. \Cref{fig:LS-map} shows the spatial distribution of the dataset.

We consider five input features: groundwater depth (GWD), peak ground acceleration (PGA), elevation, distance to the closest river ($L$), and slope angle to predict lateral spreading.  
The GWD, PGA, and elevation are extracted from the New Zealand Geotechnical Database (NZGD~\cite{nzgd-gwd,nzgd-pga,nzgd-dem}). 
$L$ is the distance between a site and the closest point to the Avon River or its tributaries. 
The slope angle is computed using a preprocessed 15 m Digital Elevation Model.

\begin{figure}[h]
    \centering
    \begin{subfigure}[b]{0.3\textwidth}
        \centering
        \includegraphics[width=\textwidth]{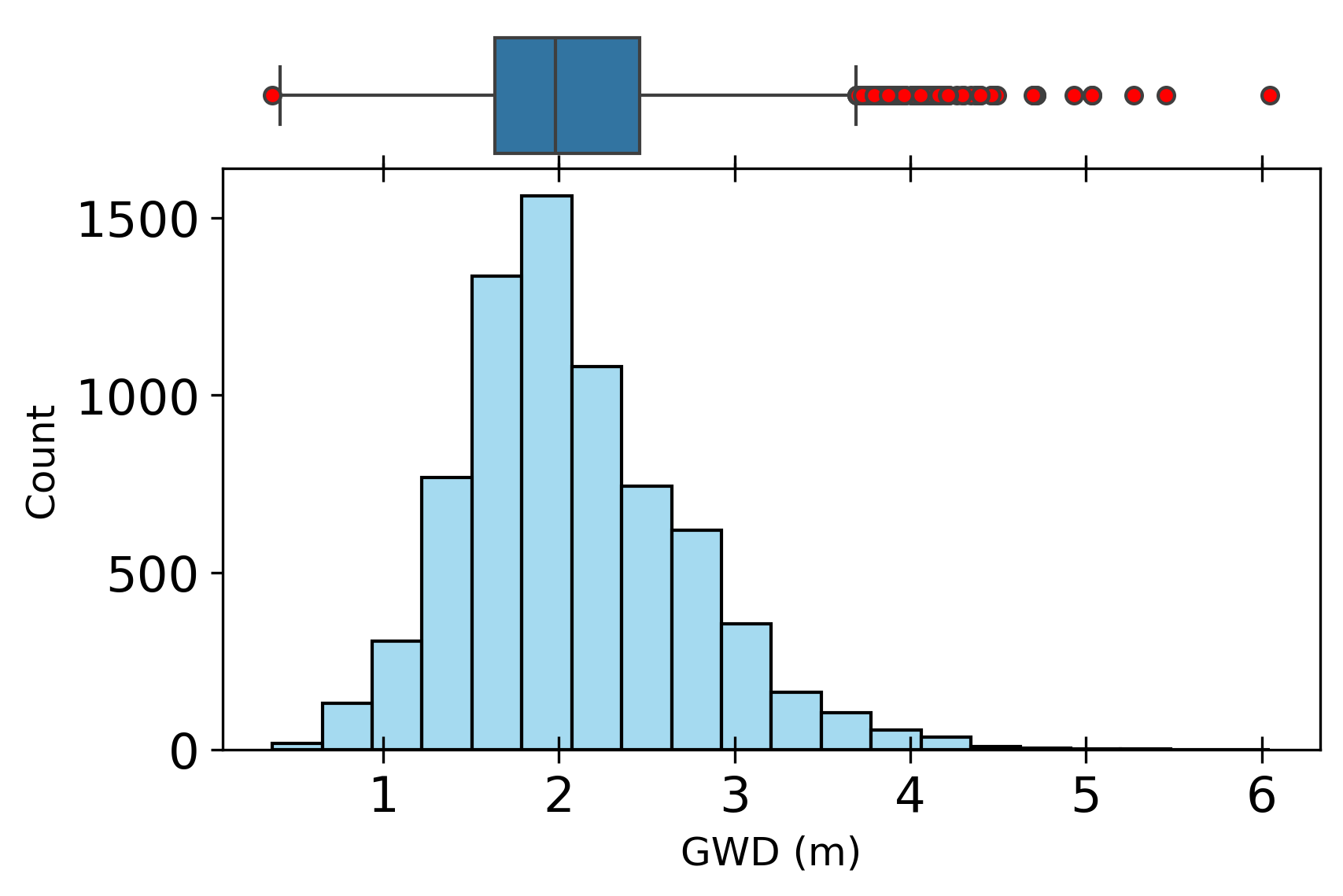}
        \caption{GWD.}
        \label{fig:hist-gwd}
    \end{subfigure}
    \hfill
    \begin{subfigure}[b]{0.3\textwidth}
        \centering
        \includegraphics[width=\textwidth]{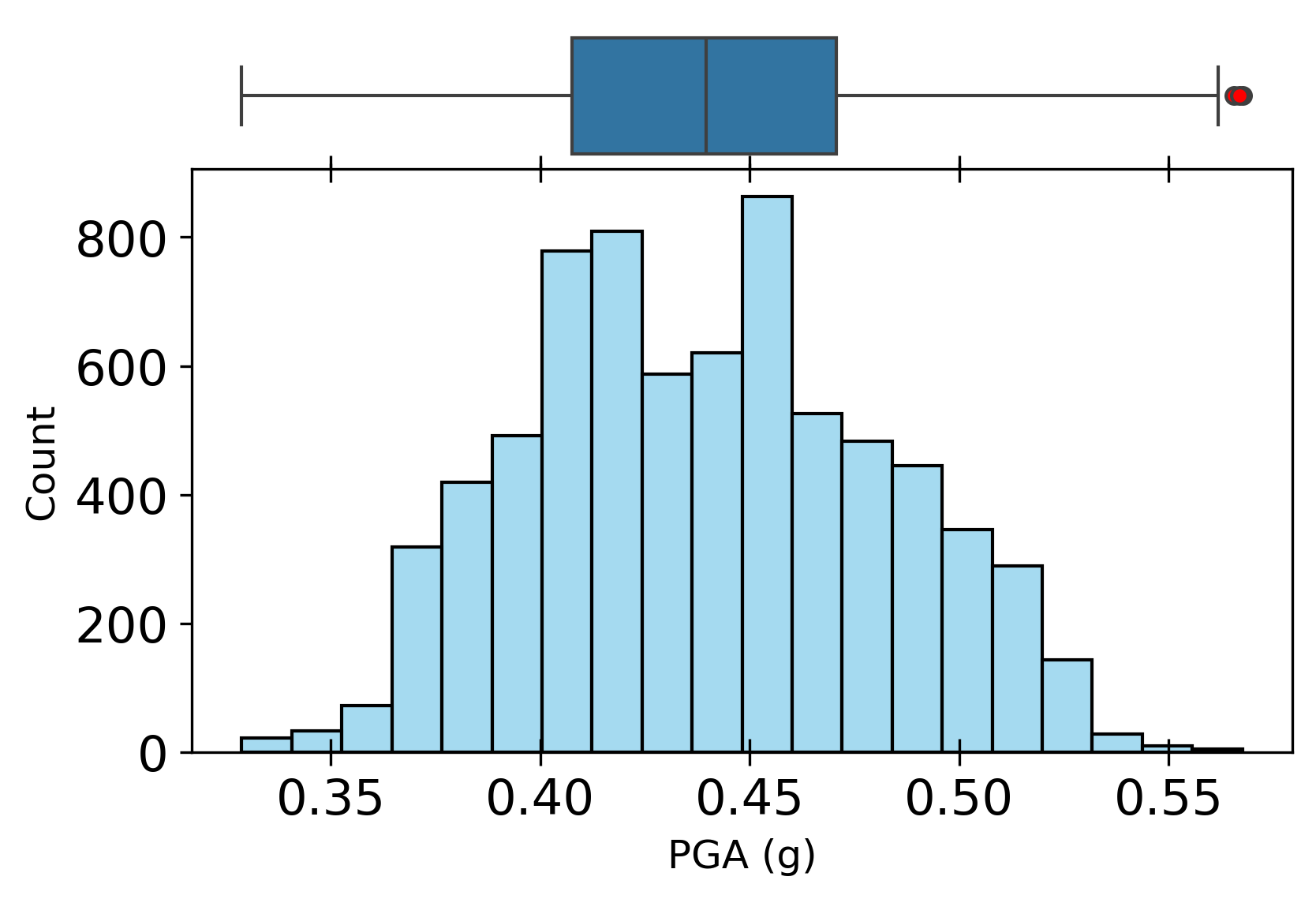}
        \caption{PGA.}
        \label{fig:hist-pga}
    \end{subfigure}
    \hfill
    \begin{subfigure}[b]{0.3\textwidth}
        \centering
        \includegraphics[width=\textwidth]{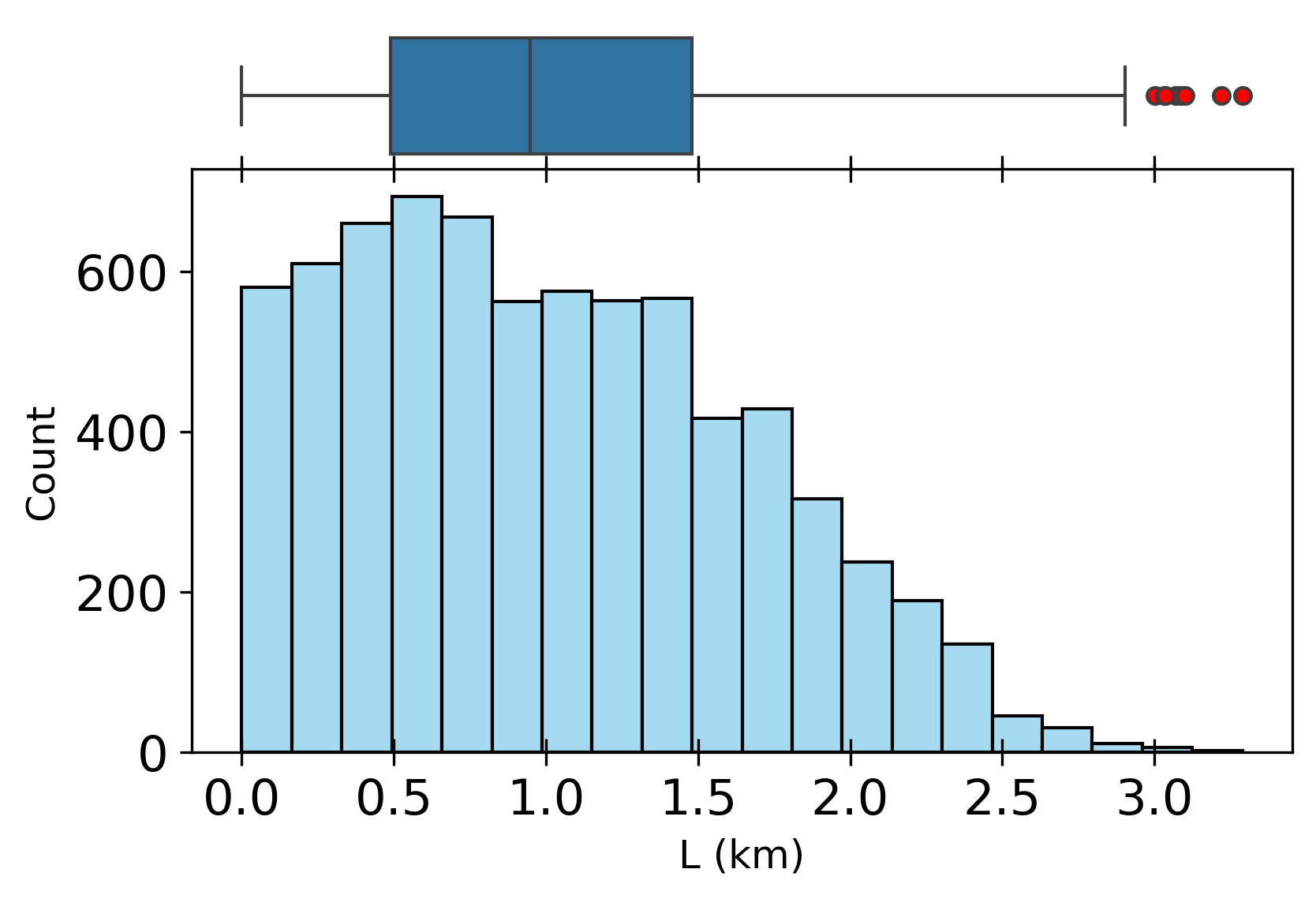}
        \caption{Distance to the rivers.}
        \label{fig:hist-L}
    \end{subfigure}
    \vfill
    \begin{subfigure}[b]{0.3\textwidth}
        \centering
        \includegraphics[width=\textwidth]{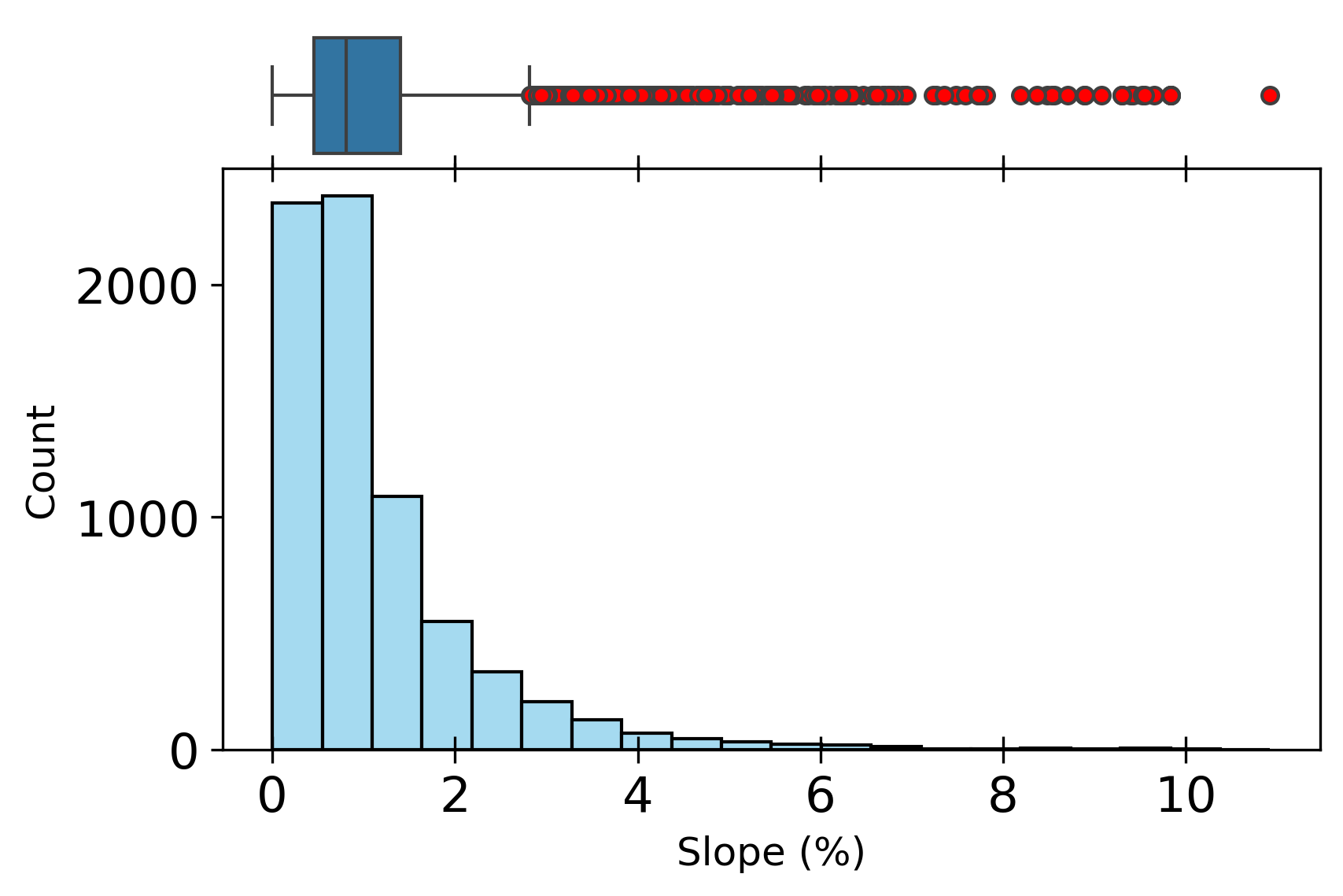}
        \caption{Slope.}
        \label{fig:hist-slope}
    \end{subfigure}
    \begin{subfigure}[b]{0.3\textwidth}
        \centering
        \includegraphics[width=\textwidth]{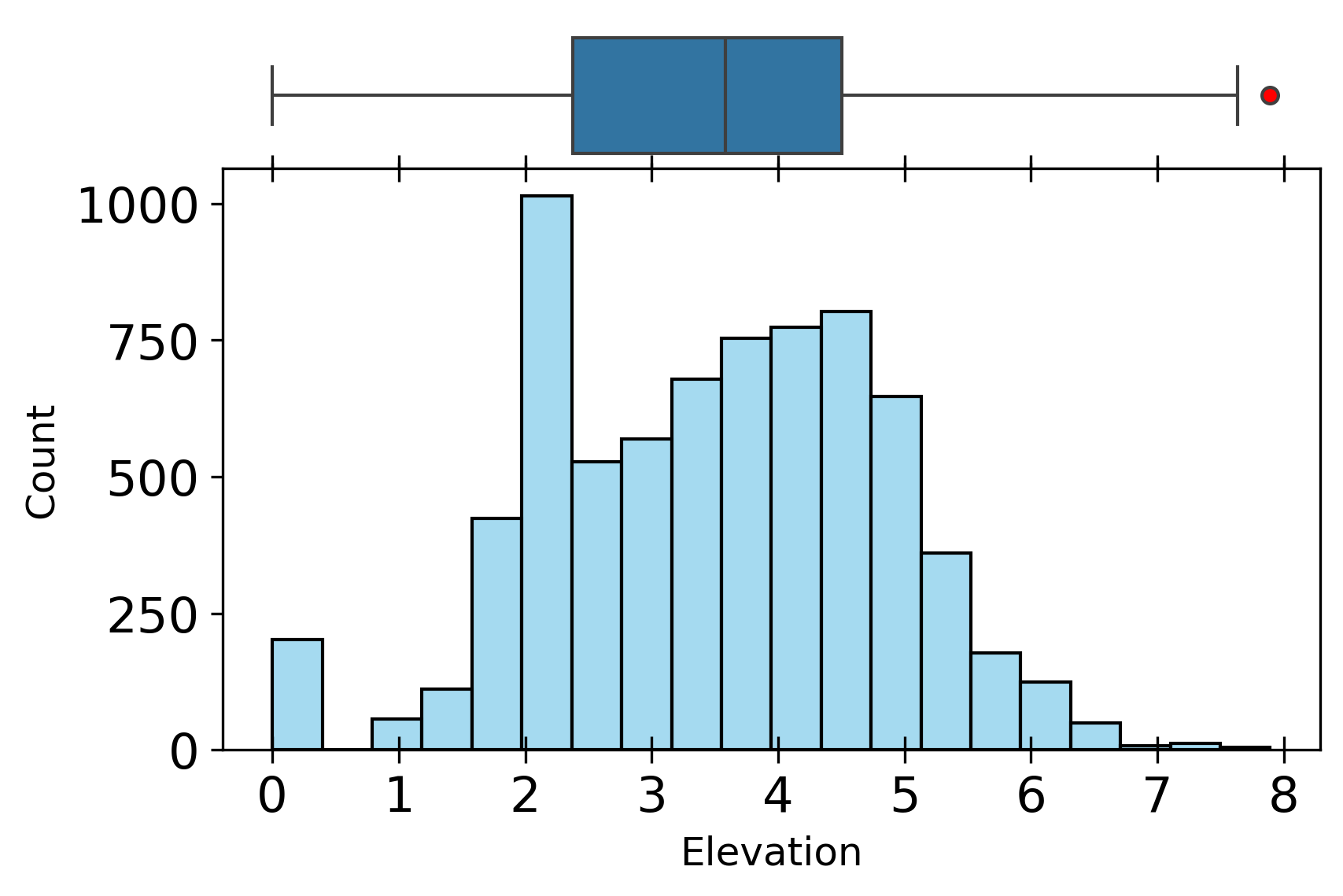}
        \caption{Elevation.}
        \label{fig:hist-elevation}
    \end{subfigure}
    \caption{Distribution of input features and outliers analysis.}
    \label{fig:hist}
\end{figure}

\Cref{fig:hist} shows the distribution of five input features with histograms and box plots with outliers marked as red dots. 
Outliers are identified using the interquartile range (IQR) method: values below $Q1 - 1.5 \times IQR$ or above $Q3 + 1.5 \times IQR$. 
We observe that the PGA, L, and elevation data distribute evenly with few outliers, while the GWD and slope data have long-tail distributions with noticeable outliers. 
All data points were retained in the analysis to preserve the full range of site conditions in the dataset.

\section{Methodology} \label{sect:meth}
This section outlines our process for training an Explainable Boosting Machine (EBM) using the lateral spreading dataset. We aim to understand the relationship between lateral spreading and various input features through an interpretable glass-box model. 

\subsection{Explainable Boosting Machine (EBM)} \label{sect:meth-ebm}
\begin{figure}[h]
    \centering
    \includegraphics[width=0.5\textwidth]{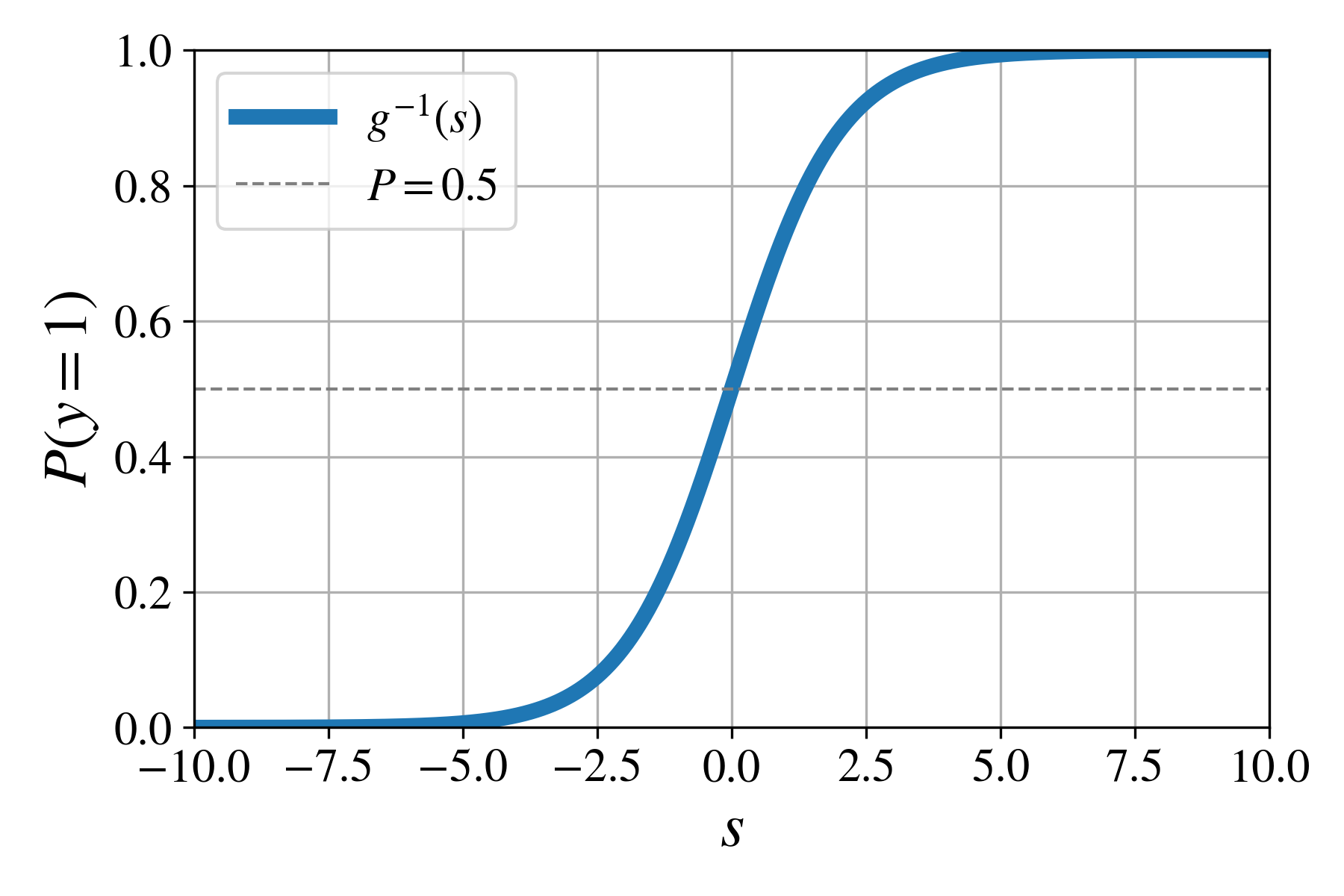}
    \caption{Relationship between the score $s$ and probability $P(y=1)$ through the logistic function. The classification threshold $P=0.5$ corresponds to $s=0$.}
    \label{fig:g-func}
\end{figure}

The Explainable Boosting Machine (EBM; \cite{nori-2019}) is a generalized additive model (GAM) that uses gradient boosting to learn shape functions.
A GAM~\citep{hastie-1987} predicts an outcome $y$ from input features $x_i$ by summing univariate shape functions:
\begin{equation}
\label{eq:gam}
E[y] = g^{-1}(\beta_0 + \sum_{i=1}^{n} f_i(x_i)) \,,
\end{equation}
where $\beta_0$ is the intercept and each $f_i$ is a shape function capturing the contribution of feature $x_i$. The quantity $s=\beta_0 + \sum_{i=1}^{n} f_i(x_i)$, denotes the score, obtained by summing the intercept and the outputs of all shape functions. The function $g$ is the link function connecting the score to the expected response $E(y)$. For binary classification where $y\in\{0,1\}$, the expected value corresponds to the probability of the positive class, i.e., $E[y] = P(y=1)$. In this case, the score $s$ is in logit (log-odds) space, and the inverse link function converts it to a probability:
\begin{equation}
\label{eq:gam-prob}
E[y]= g^{-1}(s) = \frac{1}{1 + \exp(-s)} \,.
\end{equation}
\Cref{fig:g-func} illustrates the relationship between the score $s$ and the predicted probability. When score $s>0$, the predicted probability exceeds 50\%, corresponding to a prediction of the positive class. Conversely, when $s<0$, the  predicted probability is below 50\%, corresponding to a prediction of the negative class. Throughout this paper, we refer to $s$ as the score, and we also use the term score to describe the output of individual shape functions.

EBMs extend GAMs by incorporating pairwise interaction terms $f_{i,j}(x_i, x_j)$:
\begin{equation}
\label{eq:ebm}
s = \beta_0 + \sum_{i=1}^{n} f_i(x_i) + \sum_{i<j} f_{i,j}(x_i, x_j) \,.
\end{equation}
These bivariate shape functions capture interactions between feature pairs that univariate terms cannot represent. 

\begin{figure}[h]
    \centering
    \includegraphics[width=0.7\linewidth]{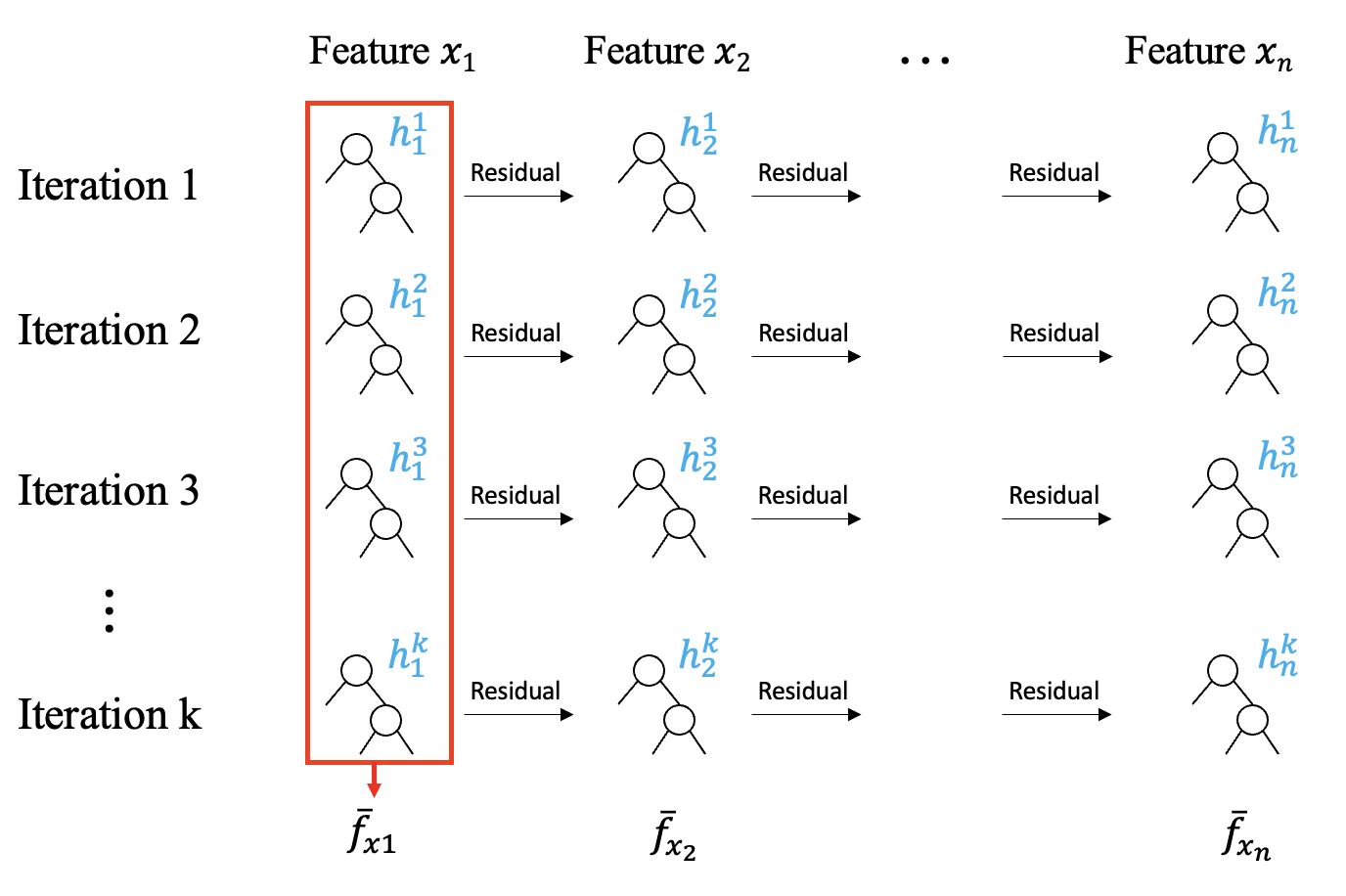}
    \caption{Round-robin training for univariate shape functions.}
    \label{fig:ebm-framework}
\end{figure}

EBMs learn the univariate shape functions $f_i$ (and similarly, the bivariate functions $f_{i,j}$) through an iterative process using gradient boosting, with decision trees as base learners (as illustrated in~\Cref{fig:ebm-framework}). The training process begins by initializing the model to the log-odds of positive cases in the training dataset:

\begin{equation}
\label{eq:training-init}
s=\beta_0^{init}=\log\frac{p}{1-p}\,,
\end{equation}
where $p$ is the positive case ratio in the training dataset.
At this stage, the model contains no shape functions $f_i$ or interaction terms $f_{i,j}$.
Because the model is a constant, all predictions on the training dataset are identical.
The residuals represent the error with respect to the ground truth.

In the next step, a shallow decision tree $h_1(x_1)$ is trained on a single feature $x_1$ to reduce the residuals.
This tree is then added to the model with a small learning rate $\eta$:

\begin{equation}
\label{eq:training-add_tree}
s=\beta_0^{init}+\eta\cdot h_1(x_1) \,,
\end{equation}

The residuals of the updated model are then used to train a new tree on a different feature.
This process repeats until all features have been cycled through, completing one iteration.
The small learning rate ensures that the order of features within an iteration does not significantly affect the result.
Because the learning rate is small, the number of iterations is typically large (e.g., 5000).

After $k$ iterations, each feature $x_i$ has $k$ associated trees.
The updated model can be expressed as:
\begin{equation}
\label{eq:training-after_iter}
s=\beta_0^{init}+\sum_{i=1}^n\sum_{t=1}^k\eta\cdot h_i^t(x_i) \,,
\end{equation}
where $n$ is the number of features, $k$ is the total number of iterations, and $h_i^t$ is the tree trained on feature $i$ at iteration $t$.
Since each tree depends on only one feature, the $k$ trees for each feature can be summed into a single shape function:

\begin{equation}
\label{eq:training-merge}
\bar{f}_i(x_i) = \sum_{t=1}^k \eta \cdot h_i^t(x_i) \,,
\end{equation}

In addition to the univariate terms $f_i$, EBMs selectively incorporate bivariate interaction terms $f_{i,j}$ to capture pairwise feature dependencies.
Let $U=\{(i,j) \mid 1\leq i< j\leq n\}$ denote the set of all candidate feature pairs.
After the model in \cref{eq:training-merge} is built using only univariate terms, the EBM evaluates potential interaction terms.
For each pair $(i,j) \in U$, a bivariate tree $h_{i,j}$ is trained on the residuals of the current model. The bivairate tree is used to fast select the interaction pairs with highest potential by evaluating the residual from each tree. The best m, a user-defined number, pairs are selected to the round-robin cyclic training stage, similar to univariate term training. The model after adding this stage training becomes:

\begin{equation}
\label{eq:training-bivariate}
s=\beta_0^{init}+\sum_{i=1}^n\bar{f}_i(x_i)+\sum \bar{f}_{i,j}(x_i, x_j)\,,
\end{equation}

Additionally, EBMs center each shape function to zero mean and the final model is written:
\begin{align}
\label{eq:training-centering}
s&=\beta_0^{init}+\sum_{i=1}^n(\bar{f}_i(x_i)-E[\bar{f}_i])+\sum(\bar{f}_{i,j}(x_i, x_j)-E[\bar{f}_{i,j}])+\Big(\sum_{i=1}^nE[\bar{f}_i]+\sum E[\bar{f}_{i,j}]\Big) \\
&=\beta_0+\sum_{i=1}^nf_i(x_i)+\sum f_{i,j}(x_i, x_j)
\,,
\end{align}
where $E[\bar{f}_i]$ and $E[\bar{f}_{i,j}]$ are the average output of a shape function for the training dataset.

The reason for centering is to make the model identifiable. Consider two models: $s_a = 5 + \bar{f}_1^a + \bar{f}_{1,2}^a$ and $s_b = 8 + \bar{f}_1^b + \bar{f}_{1,2}^b=8+(\bar{f}_1^a-1)+(\bar{f}_{1,2}^a-2)$. Both models have the same total score but different scores in each term. This would lead confusion in explanation. By centering each shape function, it enforces a zero-mean constraint, ensuring both models have same score in each term. 

Centering does not affect predictions.
However, it changes the intercept's meaning from the log-odds of the training data to the average model output in logit space.

With the training procedure and centering explained, we now describe how EBMs derive predictions and how to interpret them.
Consider a hypothetical dataset with features $x_1$ and $x_2$ and binary output $y$.
An EBM trained on these features computes the score:
\begin{equation}
\label{eq:ebm-example}
s = \beta_0+f_{1}(x_1)+f_{2}(x_2)+f_{1,2}(x_1, x_2)\,.
\end{equation}

\begin{figure}[h]
    \begin{subfigure}[b]{0.32\textwidth}
        \includegraphics[width=\textwidth]{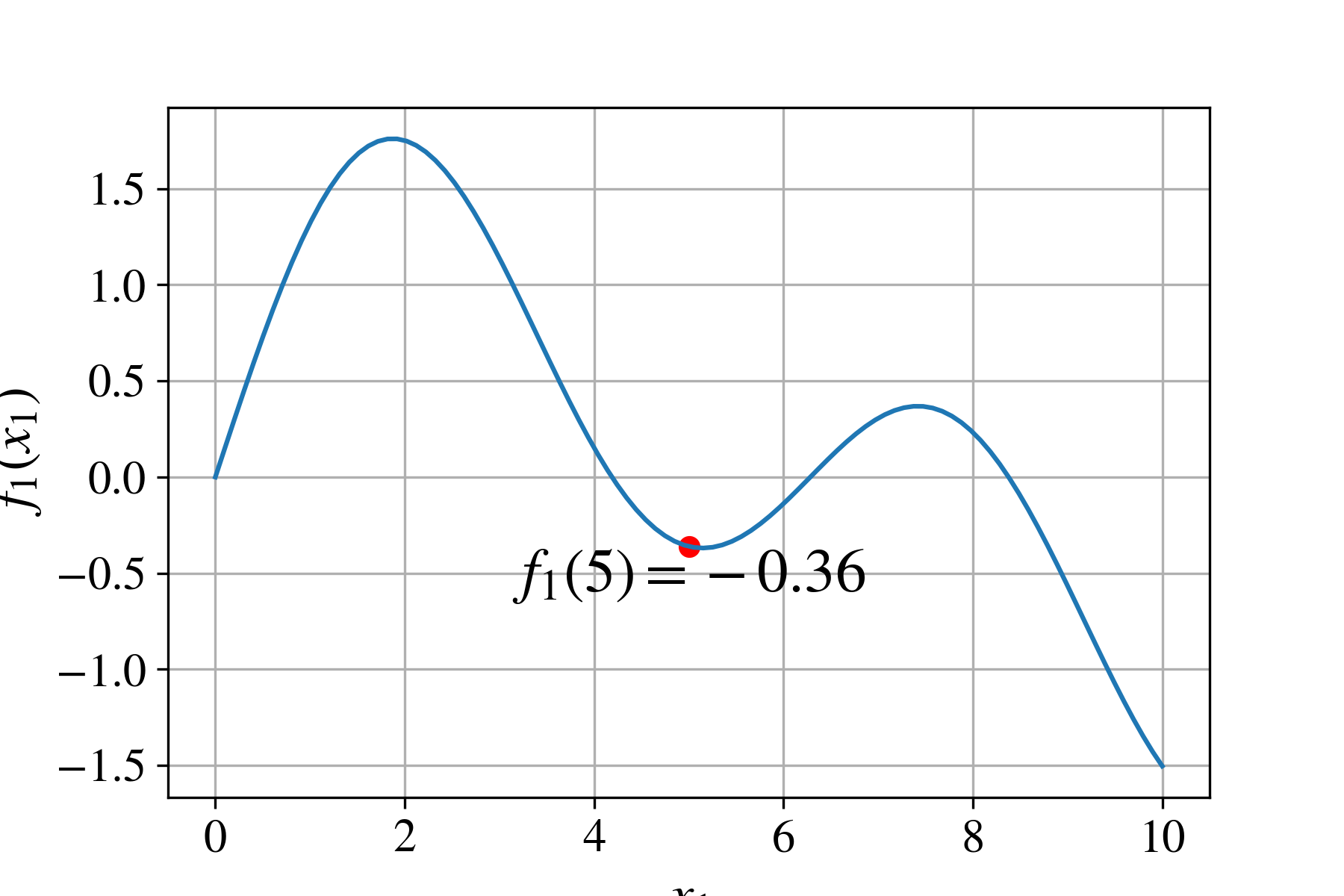}
        \caption{$f_1(x_1)$.}
        \label{fig:1d-shape-function_x1}
    \end{subfigure}
    \hfill
    \begin{subfigure}[b]{0.32\textwidth}
        \includegraphics[width=\textwidth]{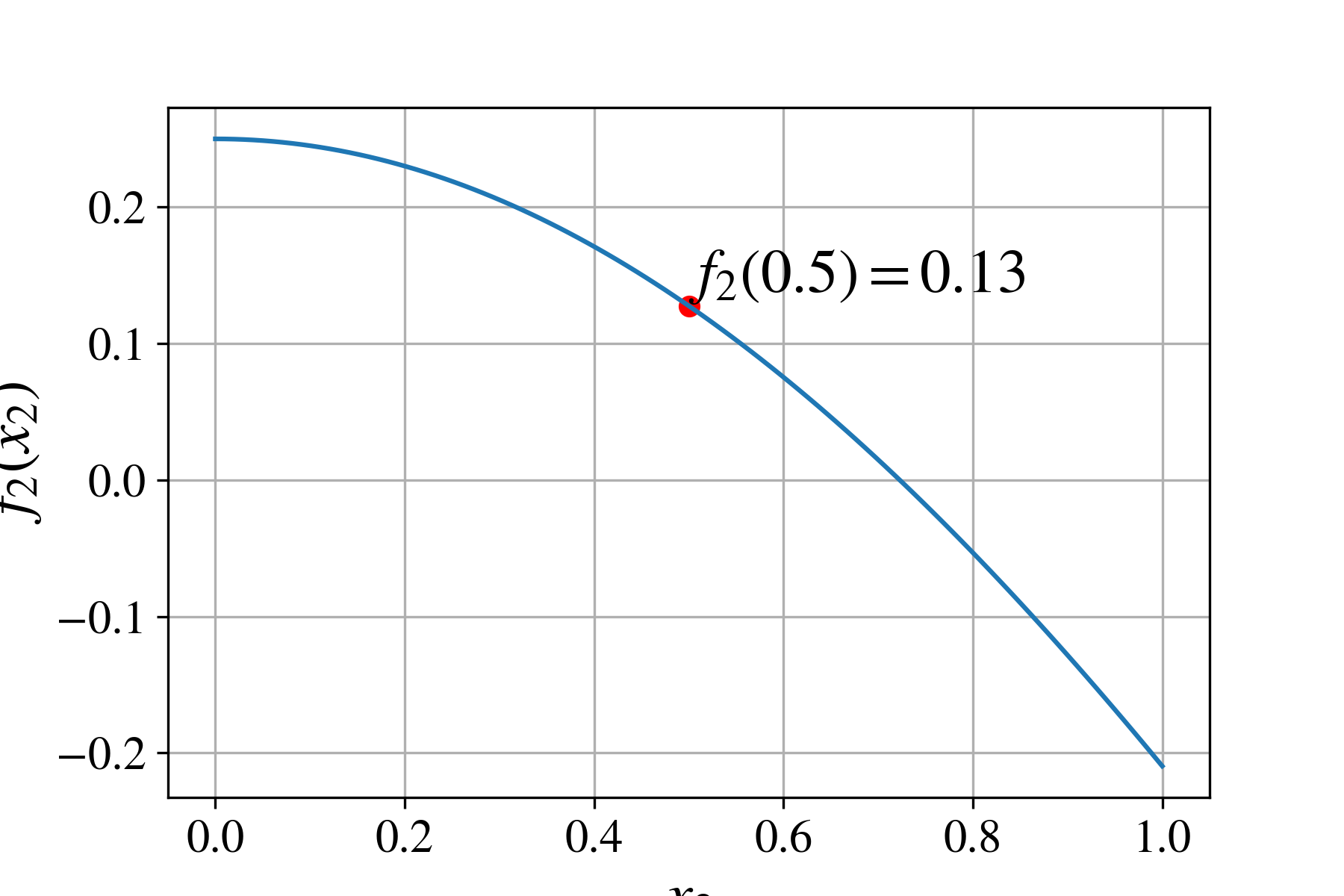}
        \caption{$f_2(x_2)$.}
        \label{fig:1d-shape-function_x2}
    \end{subfigure}
    \hfill
    \begin{subfigure}[b]{0.32\textwidth}
        \includegraphics[width=\textwidth]{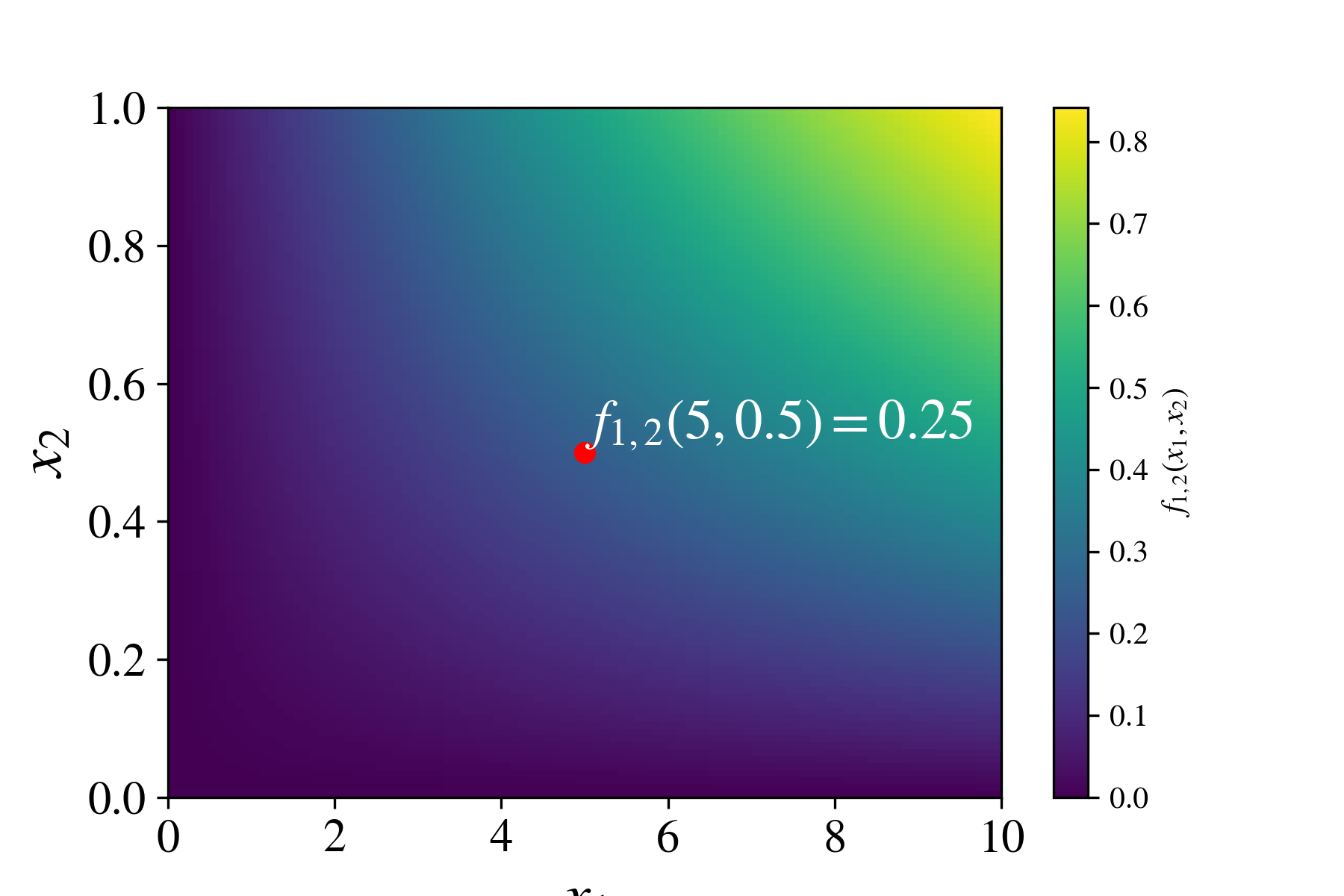}
        \caption{$f_{1,2}(x_1, x_2)$.}
        \label{fig:2d-shape-function}
    \end{subfigure}
    \caption{Shape function examples. (a)(b) 1D shape function examples for $f_1$ and $f_2$. The x axis is the feature value and the y axis is the score. (d) 2D shape function example for $f_{1,2}$. The color denotes the score. The red point denotes the example point with feature value $x_1=5$ and $x_2=0.5$.}
\end{figure}

Each term in~\cref{eq:ebm-example} can be visualized, as they are either univariate ($f_1, f_2$) or bivariate ($f_{1,2}$) functions.
Suppose~\Cref{fig:1d-shape-function_x1,fig:1d-shape-function_x2,fig:2d-shape-function} depict the learned functions $f_1$, $f_2$, and $f_{1,2}$, respectively.
The output of each function represents that feature's contribution to the total score $s$.

Consider an example data point where $x_1=5$ and $x_2=0.5$.
We look up the contribution for $x_1=5$ from $f_1$ (\Cref{fig:1d-shape-function_x1}), which returns $-0.36$.
Similarly, $f_2(0.5) = 0.13$ (\Cref{fig:1d-shape-function_x2}), and $f_{1,2}(5, 0.5) = 0.25$ (\Cref{fig:2d-shape-function}).
Assuming $\beta_0=0$ for simplicity, the total score is:
\begin{equation}
s = 0 + (-0.36) + 0.13 + 0.25 = -0.02\,.
\end{equation}

To convert this score to a probability, we apply the inverse link function $g^{-1}$:
\begin{equation}
\label{eq:logistic}
P(y=1) = \frac{1}{1+\exp(-s)} = \frac{1}{1+\exp(0.02)} \approx 0.495 = 49.5\%\,.
\end{equation}
Since this probability is less than $0.5$, the model predicts class $0$ for this example.

The score $s$ provides an intuitive decision rule: if $s > 0$, then $P(y=1) > 50\%$, predicting class $1$; if $s < 0$, then $P(y=1) < 50\%$, predicting class $0$.

Because $s$ is a sum of individual contributions, each term directly reflects that feature's impact on the prediction.
For example, $f_1(5)=-0.36$ indicates that $x_1=5$ decreases the log-odds by $0.36$, while $f_2(0.5)=0.13$ indicates that $x_2=0.5$ increases the log-odds by $0.13$.

The shape functions themselves are also interpretable.
If~\Cref{fig:1d-shape-function_x2} shows that $f_2(x_2)$ decreases as $x_2$ increases, this indicates that higher values of $x_2$ reduce the probability of class $1$, holding other features constant.

\subsection{EBM model for lateral spreading} \label{sect:meth-eval}
In this study, we train an EBM model on the lateral spreading dataset to predict binary occurrence outcomes. 
The dataset is split into training (70\%), validation (15\%), and testing (15\%) sets.

We use the default hyperparameter settings provided by the EBM python library (interpretML v0.5.0; \cite{nori-2019}), which are designed to prioritize interpretability and control overfitting. Specifically, the model uses a small learning rate of 0.01 and allows up to 5,000 boosting rounds. Each tree is shallow, with a maximum of 3 leaves, ensuring that individual trees remain simple and interpretable. Early stopping is set to 50 iterations, meaning that training halts if validation performance does not improve over 50 consecutive boosting rounds.

To evaluate EBM’s performance, we compare it with two widely used tree-based models: Random Forest (RF;\cite{breiman-2001}) and eXtreme Gradient Boosting (XGB;\cite{Chen_2016}). The RF model is configured with a maximum depth of 9 and 50 estimators, following the setup presented by \citet{durante-2021}. For the XGB model, we perform hyperparameter tuning using grid search with 5-fold cross-validation. The optimal configuration for XGB includes a learning rate of 0.1, a maximum depth of 8, and 100 estimators. Full model configurations are listed in~\Cref{tab:rf-xgb-ebm-compare}.

We assess the performance of all three models on the testing dataset using accuracy, precision, recall, F1-score, and the area under the receiver operating characteristic curve (AUC-ROC).
As shown in~\Cref{tab:rf-xgb-ebm-compare}, EBM achieves an accuracy of 0.80, an F1-score of 0.75, and an AUC of 0.88.
This performance is comparable to RF, while XGB outperforms both models across all metrics.

The superior performance of XGB is generally expected, as it benefits from boosting, allowing it to incrementally improve the model through sequential learning. Additionally, XGB utilizes deeper trees that can learn complex interactions involving multiple features simultaneously. In contrast, RF does not apply boosting. EBM also employs boosting but with shallow trees and univariate inputs to maintain simplicity. This design choice emphasizes interpretability in EBM, prioritizing model transparency over maximizing predictive complexity.

\begin{table}[]
\centering
\caption{Model configuration and testing performance for lateral spreading in Christchurch.}
\label{tab:rf-xgb-ebm-compare}
\begin{tabular}{lccc}
\toprule
 & RF & XGB & EBM  \\ \midrule
Learning rate & - & 0.1 & 0.01 \\
Max iteration & - & - & 5,000 \\
Max leaves & - & - & 3 \\
Early stopping & - & - & 50 \\
\# Estimators & 50 & 100 & - \\
Max depth & 9 & 8 & -\\ \midrule
Accuracy & 0.82 & \textbf{0.84} & 0.80  \\
Precision & \textbf{0.86} & 0.82 & 0.82  \\
Recall & 0.69 & \textbf{0.76} & 0.70  \\
F1-score & 0.77 & \textbf{0.81} & 0.75  \\
AUC & 0.89 & \textbf{0.92} & 0.88  \\
\bottomrule
\end{tabular}
\end{table}

Having evaluated EBM's predictive performance, we now turn to examining the interpretability of its learned shape functions. Unlike black-box models, EBM enables direct visualization of feature contributions, allowing inspection of model behavior across the input space.

\begin{figure}[h]
    \centering
    \begin{subfigure}[t]{0.3\textwidth}
        \includegraphics[width=\linewidth]{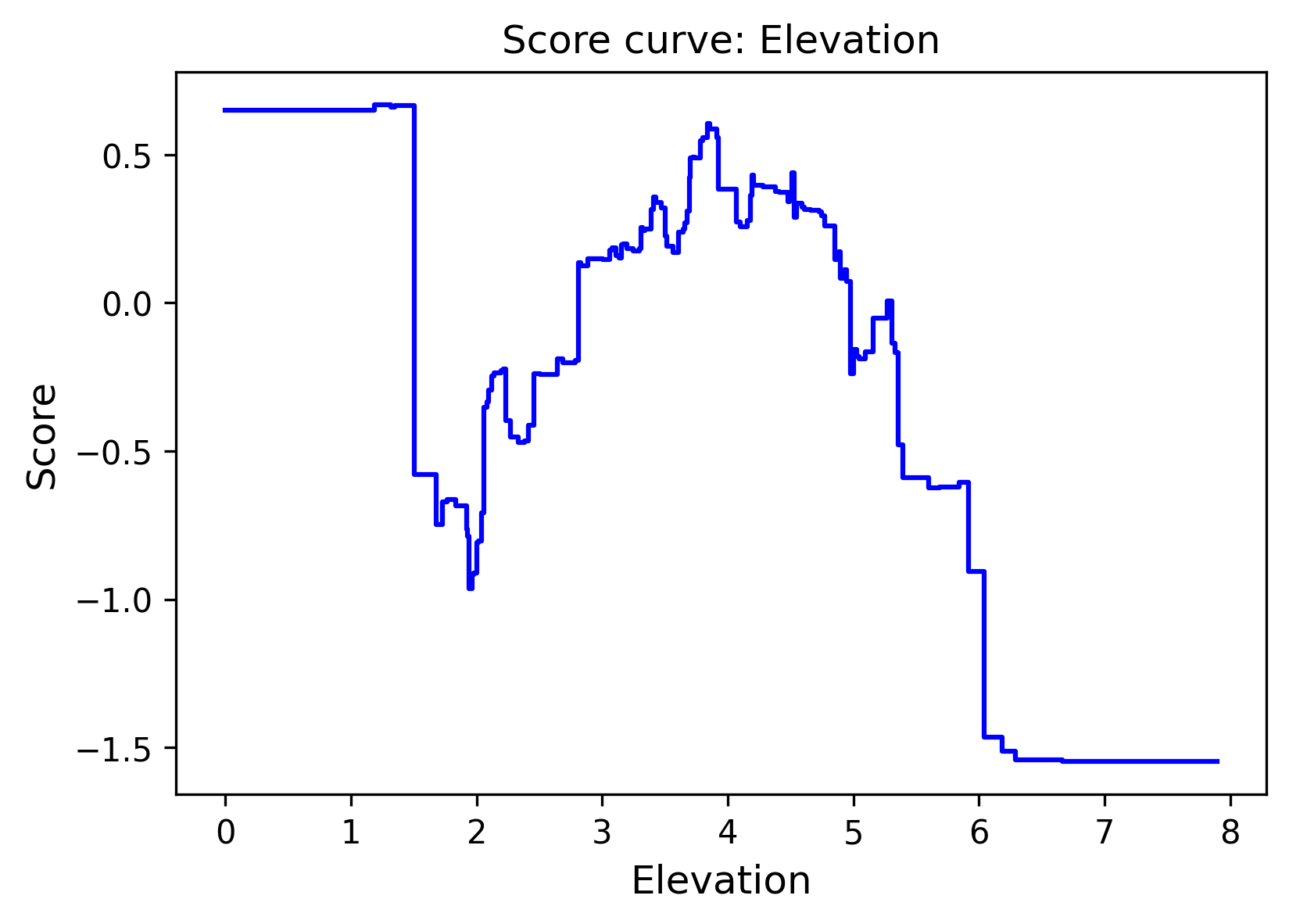}
        \caption{Elevation}
        \label{fig:uni-elevation}
    \end{subfigure}
    \hfill
    \begin{subfigure}[t]{0.3\textwidth}
        \includegraphics[width=\linewidth]{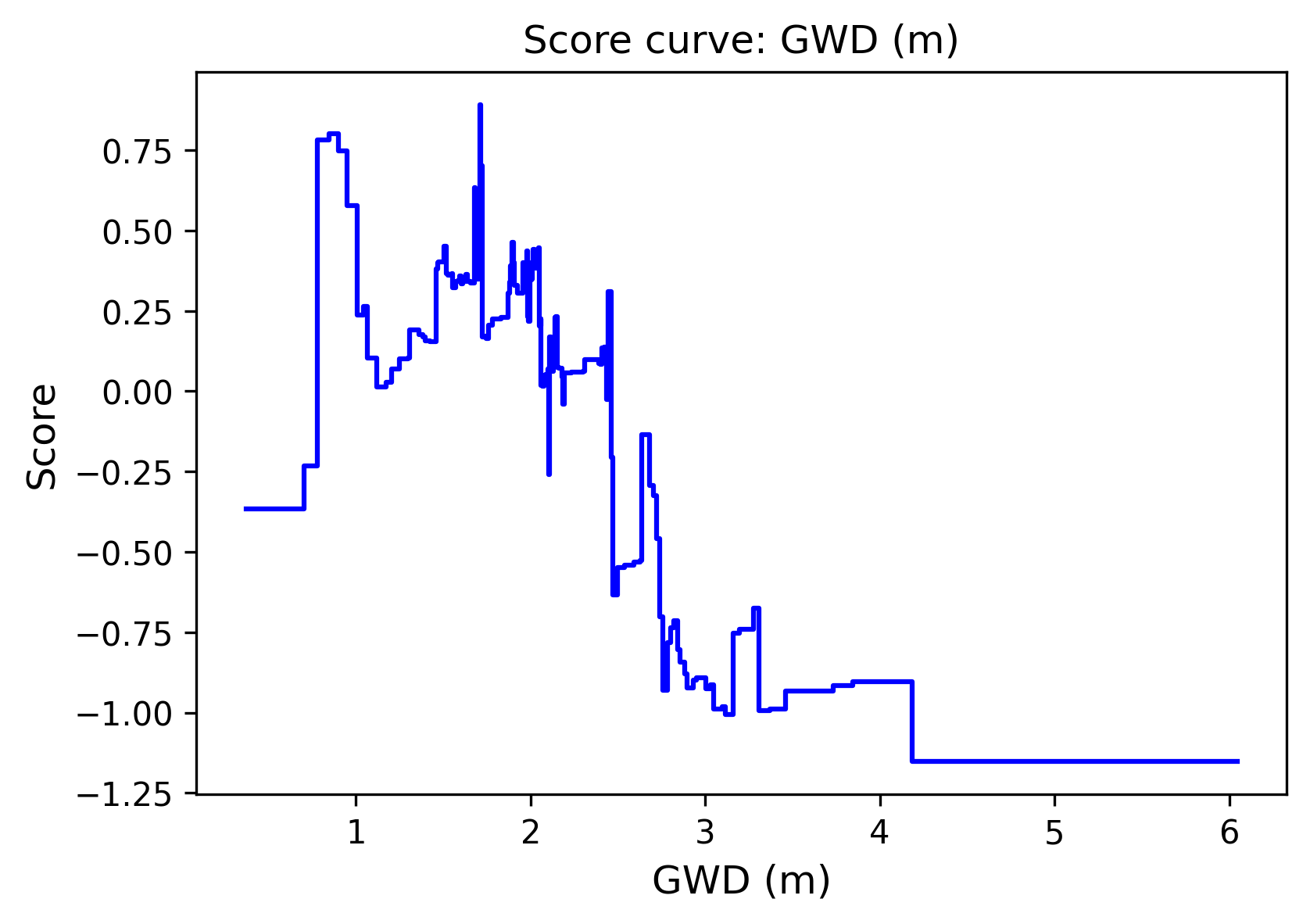}
        \caption{GWD}
        \label{fig:uni-gwd}
    \end{subfigure}
    \hfill
    \begin{subfigure}[t]{0.3\textwidth}
        \includegraphics[width=\linewidth]{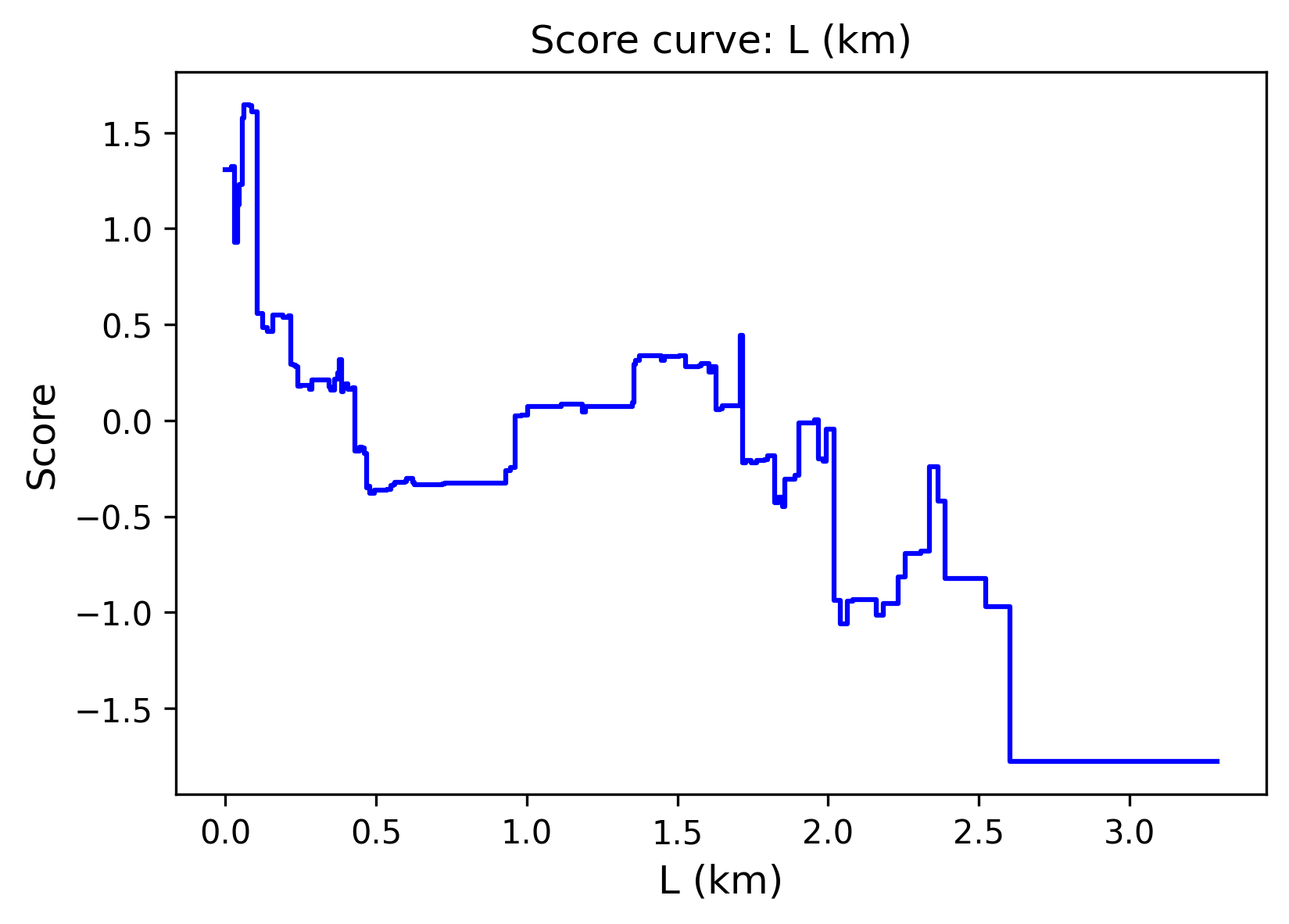}
        \caption{L}
        \label{fig:uni-l}
    \end{subfigure}
    \vfill
    \begin{subfigure}[t]{0.3\textwidth}
        \includegraphics[width=\linewidth]{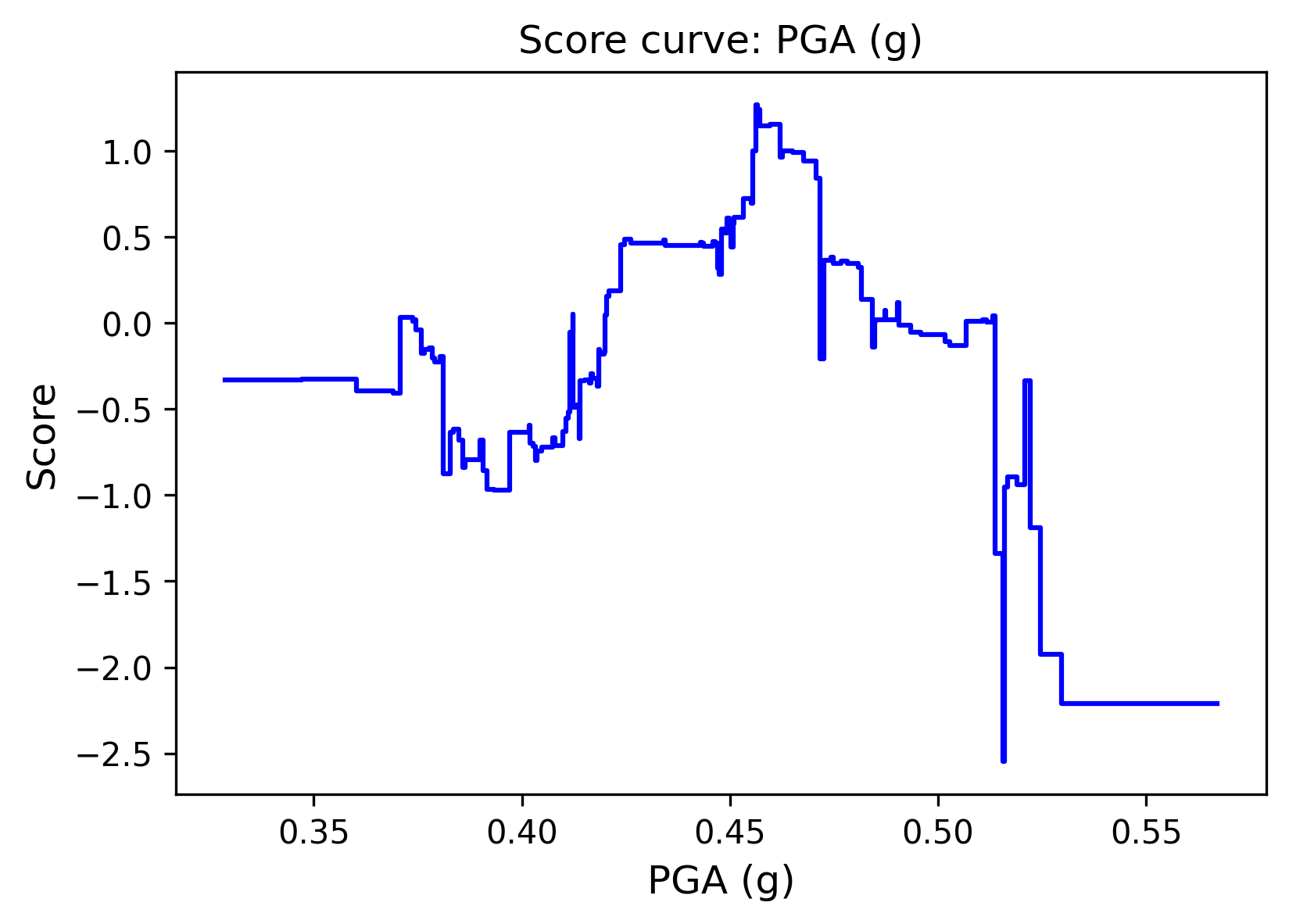}
        \caption{PGA}
        \label{fig:uni-pga}
    \end{subfigure}
    \hspace{0.5em}
    \begin{subfigure}[t]{0.3\textwidth}
        \includegraphics[width=\linewidth]{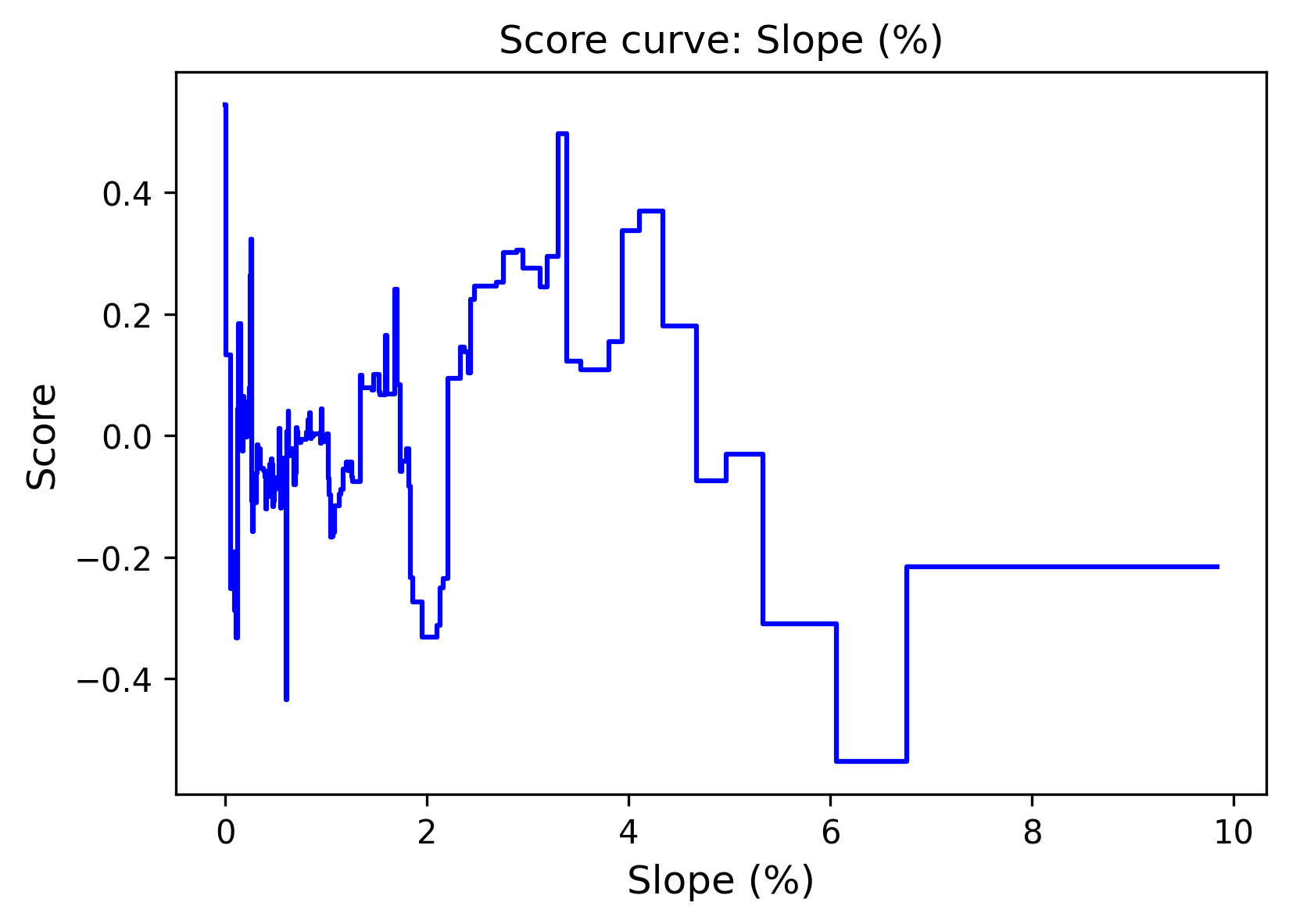}
        \caption{Slope}
        \label{fig:uni-slope}
    \end{subfigure}
    \caption{Univariate score curves from the EBM}
    \label{fig:univariate}
\end{figure}

In this study, the trained EBM consists of 5 univariate shape functions (shown in~\Cref{fig:univariate}) and 10 bivariate interaction functions (one for each of the $\binom{5}{2}=10$ feature pairs, see~\Cref{sect:appendix-bivariate}). These 15 learned functions collectively represent the full structure of the EBM model. The shape functions visualize how each predictor contributes to the susceptibility of lateral spreading.

In~\Cref{fig:uni-elevation}, the learned shape function indicates positive contributions when elevation is below approximately 1.3 m, followed by a sudden drop to negative scores at around 1.3 m. The score then increases between roughly 2 m and 4 m and gradually decrease beyond 4 m. Although no direct relationship between elevation and lateral spreading susceptibility is observed, we argue that elevation may act as a proxy for proximity to rivers (associated with lower elevations) or variations in soil conditions.

In~\Cref{fig:uni-gwd,fig:uni-l}, both GWD and $L$ show an overall decreasing trend in score as feature values increase. It suggests that the lateral spreading susceptibility reduces at deep groundwater depths or large distances from the rivers.

In~\Cref{fig:uni-pga}, the score is generally negative for PGA values below 0.42 g. The score becomes positive between about 0.42 g and 0.51 g with a peak near 0.46 g, and then drops to strongly negative values for PGA exceeding 0.51 g. The trend of relation between PGA and lateral spreading susceptibility is unclear.

Similarly, the slope shape function in~\Cref{fig:uni-slope} shows that the relation between slope and score is unclear. The function fluctuates within a relatively narrow range (approximately -0.5 to +0.5), which implies that slope is a weak predictor of lateral spreading susceptibility within the trained EBM model.

Since each shape function defines how individual features or feature interactions contribute to the prediction, we can directly assess their consistency with domain knowledge.

\begin{figure}[h]
    \centering
    \begin{subfigure}[b]{0.50\textwidth}
        \centering
        \includegraphics[width=\textwidth]{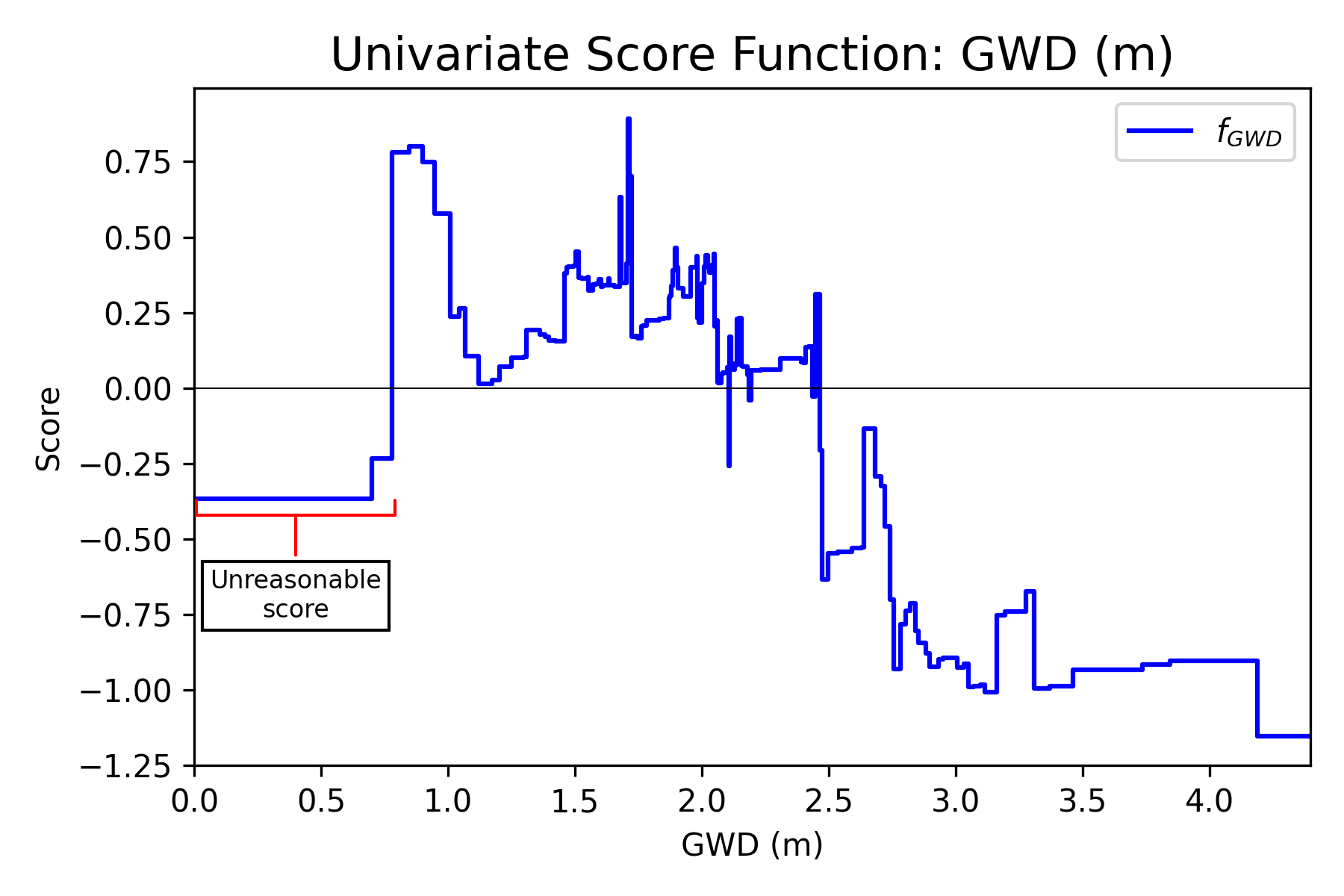}
        \caption{Shape function for single feature term ($f_{GWD}$).}
        \label{fig:score-example}
    \end{subfigure}
    \hfill
    \begin{subfigure}[b]{0.40\textwidth}
        \centering
        \includegraphics[width=\textwidth]{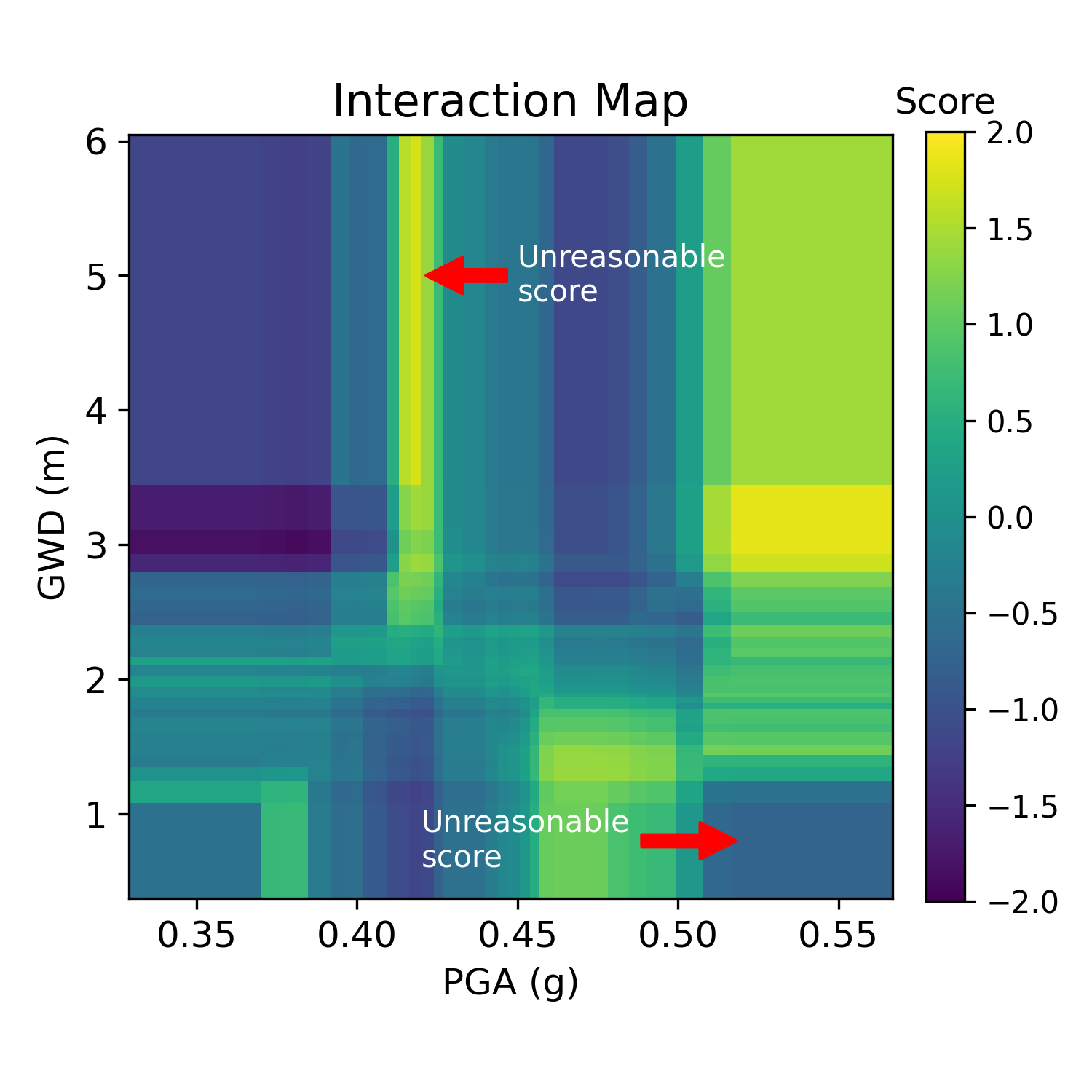}
        \caption{Shape functions for paired features term ($f_{GWD, PGA}$).}
        \label{fig:interaction-example}
    \end{subfigure}
    \caption{Examples of look-up graphs generated by EBM, with arrows highlighting non-physical behaviors.}
    \label{fig:feature-score-example}
\end{figure}

In~\Cref{fig:feature-score-example}, we examine the learned univariate function for GWD and the bivariate interaction between GWD and PGA. We identify inconsistencies between the model's learned patterns and established domain knowledge. For instance:

\begin{enumerate} 
    
    \item In~\Cref{fig:score-example}, the shape function for GWD assigns a negative score at very shallow depths ($<$0.7 m), implying that shallow GWD reduces the likelihood of lateral spreading. This contradicts physical understanding: shallow groundwater typically increases the susceptibility to lateral spreading because a larger thickness of liquefiable soil is saturated, and the likelihood should decrease as GWD becomes deeper.
    \item We also observe an unphysical pattern in the interaction function shown in~\Cref{fig:interaction-example}. In the region where PGA exceeds 0.51 g and GWD is less than 1.2 m, the model assigns a negative score of -0.7. This suggests that high ground motion combined with shallow groundwater reduces the likelihood of lateral spreading, which conflicts with our domain knowledge.
\end{enumerate}

With further investigation, we found that these inconsistencies in the model's learned behavior are likely due to bias in the training dataset. To explore this, we examined the distribution of training samples in the input regions where the model exhibited unphysical predictions. Specifically, we found that only 21 sites in the training set have GWD less than 0.7 meters, and among them, only 2 experienced lateral spreading. This imbalance likely led the model to associate shallow GWD with a lower likelihood of lateral spreading. Similarly, there are 35 cases where GWD is shallower than 1.2 m and PGA exceeds 0.51 g, yet only two of these cases experience lateral spreading. 

The predominance of negative cases in the shallow GWD and high PGA region suggests that the absence of lateral spreading is governed by additional factors not included in the model, such as soil type. When trained on such data, the EBM can only use the available features to compensate for the effects of missing factor, which may lead to statistically reasonable but physically incorrect negative scores, as shown in~\Cref{fig:feature-score-example}. While EBM’s transparency exposes these non-physical artifacts, enhancing the model's trustworthiness, it is not sufficient to merely identify where the model behaves unexpectedly. To further improve model reliability, we propose incorporating domain knowledge into the learned shape functions, as discussed in the following section.

\subsection{Incorporating domain knowledge in EBM}
The key advantage of EBM over traditional black-box models is its fully transparent decision-making process. Predictions in EBM are constructed by summing interpretable univariate and bivariate shape functions, allowing users to examine and reason about how each feature contributes to the output. This structure not only supports interpretability but also makes it possible to directly modify the model based on domain knowledge. In contrast, models like Random Forests and XGBoost rely on complex ensembles of decision trees, where feature contributions are deeply entangled and not easily accessible or adjustable. As a result, EBM provides a unique opportunity to incorporate expert knowledge by refining or constraining individual shape functions without retraining the entire model.

To incorporate domain knowledge into the EBM model, we develop a two-stage post-processing approach that modifies learned shape functions for a subset of physically understood features. This process includes:
\begin{enumerate}
    \item Univariate function modification:
    \begin{enumerate}
        \item We first examine the learned shape function $f_i(x_i)$ and identify regions where the model behavior contradicts established domain knowledge.
        \item We fit a new function $f'_i(x_i)$ to the “trustworthy” region using constraints aligned with physical expectations (e.g., monotonicity)
        \item The original shape function is then replaced with the fitted one.
    \end{enumerate}
    \item Bivariate function modification:
    \begin{enumerate}
        \item We synthesize a new interaction function $f'_{i,j}(x_i,x_j)$ using the modified univariate functions.
        \item This synthesized function is rescaled to match the min and max values of the original interaction $f_{i,j}$ to keep the range of influence similar.
        \item A threshold is applied to identify and replace only those regions of the original bivariate function that deviate significantly from domain knowledge.
    \end{enumerate}
\end{enumerate}
We apply modification only to the GWD, PGA, and $L$ features and their associated interaction terms (i.e., GWD-PGA, GWD-$L$, and $L$-PGA).

This selection is based on three considerations. First, the relationships between these variables and lateral spreading are qualitatively understood from physical principles. For example, shallower GWD and higher PGA are expected to increase the likelihood of lateral spreading. Second, these features show substantial influence in the original EBM. Correcting their shape functions will be more meaningfully and effectively enhance model trustworthy. In contrast, although slope has a physically interpretable relationship with lateral spreading, its associated shape functions contribute little to the model prediction, suggesting that it functions as a weak predictor within the learned representation. Third, highly influential variables, such as elevation, lack a clear direct physical interpretation and may instead act as proxies for other factors (e.g., soil type or proximity to rivers as we mentioned before).

For these reasons, post-processing is restricted to features that are both physically interpretable and influential in the original model.

\subsubsection{Univariate function modification}
For the univariate function modification, we replace the learned shape functions for GWD, PGA, and  $L$ with fitted sigmoid-like functions of the form:
\begin{equation}
\label{eq:monotonic}
f(x)= \frac{c}{1+exp(-a\cdot (x-b))}+d \,,
\end{equation}
where $a$, $b$, $c$, and $d$ are the parameters optimized to fit the “trusted” region of the original shape function. We choose this functional form for two main reasons:
(1) it enforces monotonicity, which aligns with domain knowledge about the directionality of feature influence, and (2) it introduces saturation at both ends, preventing extreme feature values from applying disproportionate influence on the model's output.

\begin{figure}[h]
    \centering
    \begin{subfigure}[b]{0.32\textwidth}
        \centering
        \includegraphics[width=\textwidth]{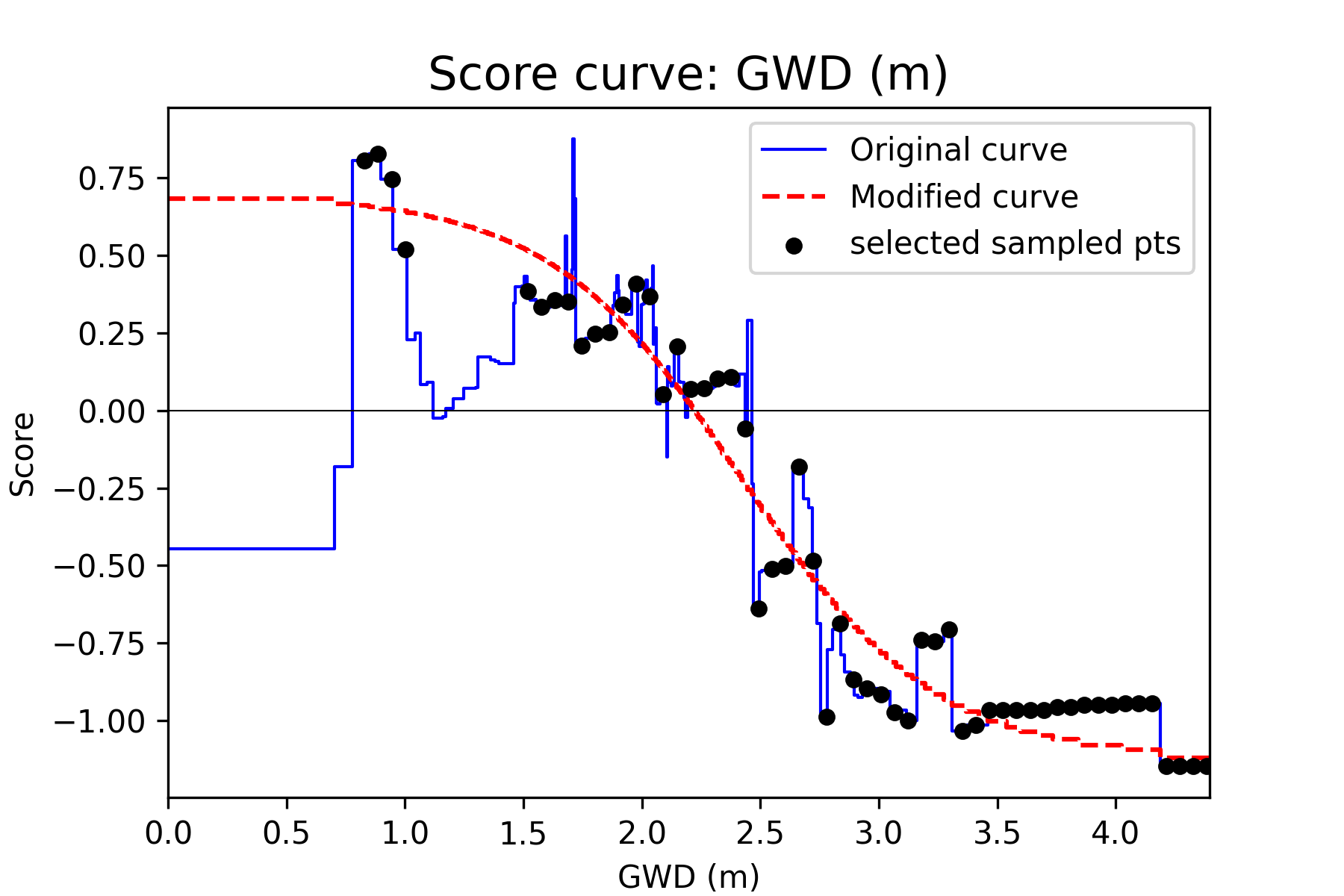}
        \caption{GWD.}
        \label{fig:modify-gwd}
    \end{subfigure}
    \begin{subfigure}[b]{0.32\textwidth}
        \centering
        \includegraphics[width=\textwidth]{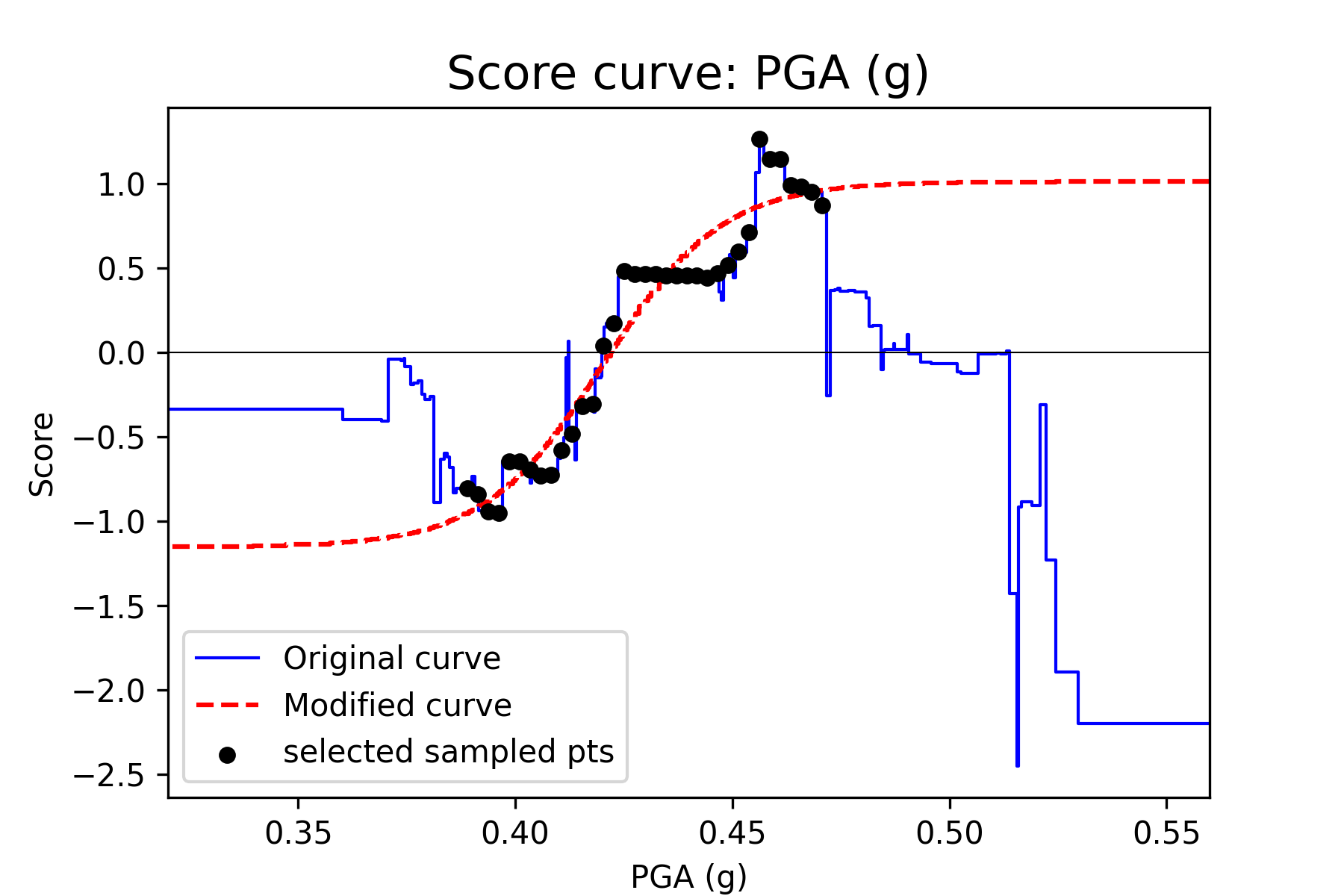}
        \caption{PGA.}
        \label{fig:modify-pga}
    \end{subfigure}
    \begin{subfigure}[b]{0.32\textwidth}
        \centering
        \includegraphics[width=\textwidth]{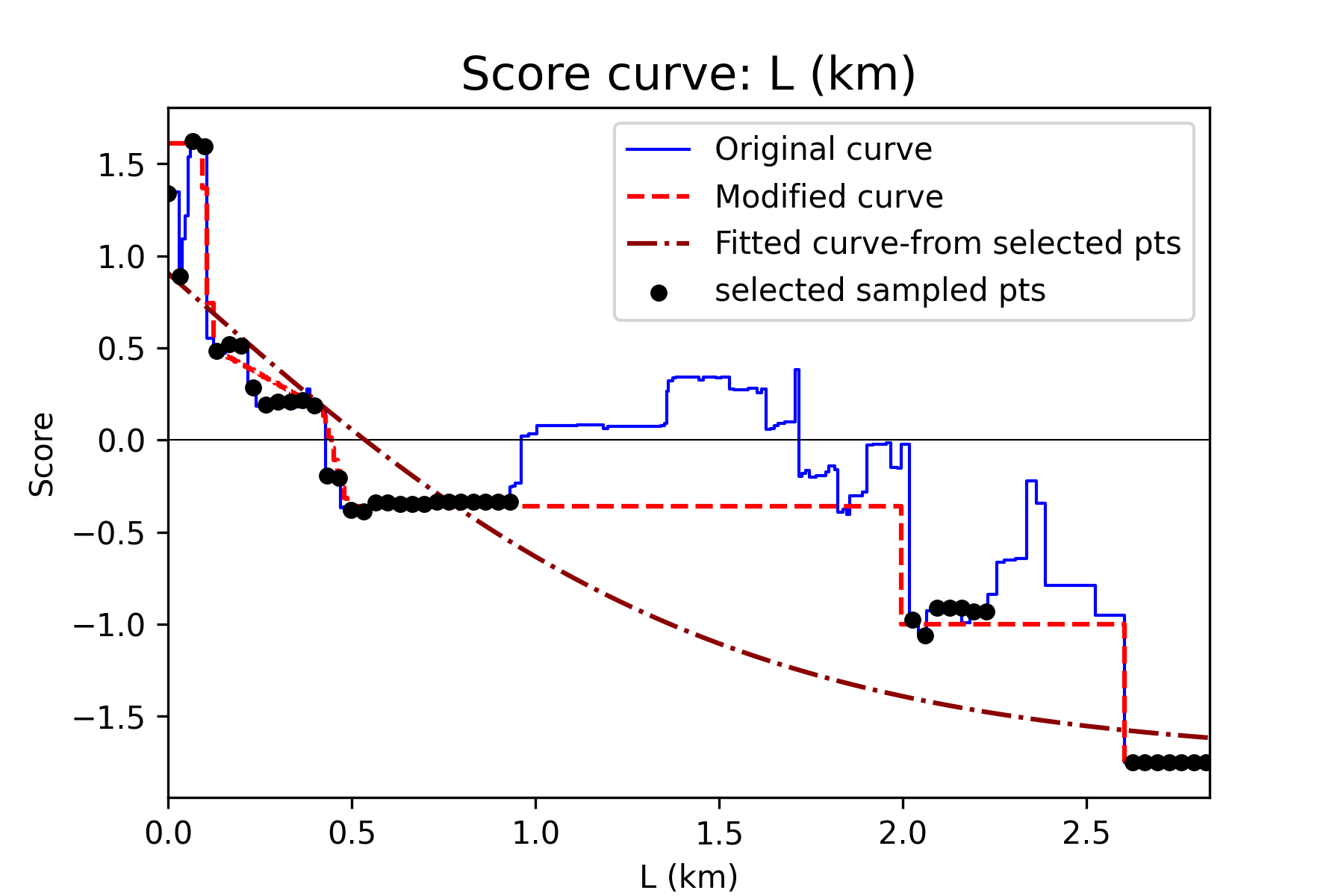}
        \caption{Distance to rivers, $L$.}
        \label{fig:modify-l}
    \end{subfigure}    
    \caption{The original and modified univariate function.}
    \label{fig:modify-score-curve}
\end{figure}

To optimize the parameters of the sigmoid-like function, we first sample 100 evenly spaced points across the range of the feature, using its minimum and maximum values in the training dataset. We then evaluate the original EBM scores at each of these sampled points. Next, we manually select a subset of sample points whose scores are consistent with domain expectations. For example, in~\Cref{fig:modify-gwd}, we show the original shape function for GWD as a blue line, with the selected sample points highlighted as black dots. Specifically, sample points in the ranges $GWD<0.7 m$ and $1.0m<GWD<1.5 m$ are excluded due to physically implausible scores. We then fit the sigmoid function by minimizing the least squares error between the selected points and the fitted curve. The resulting function, shown as a red dashed line in~\Cref{fig:modify-gwd}, is used to replace the original EBM shape function. 

This curve fitting process is also applied to PGA and $L$, as shown in~\Cref{fig:modify-pga,fig:modify-l}. However, in the case of $L$, the fitted sigmoid function does not align well with the selected sample points (see the dash-dotted line in~\Cref{fig:modify-l}). Specifically, we observe a peak score of 1.61 around L=0.1 km, followed by a sharp drop to 0.5 and a gradual decline to -0.36 at L=0.49 km. We believe this abrupt change in the shape function reflects the physical nature of the $L$ feature. 

Rivers typically serve as free faces that promote lateral spreading. Therefore, the distance to river behaves more like a binary indicator: when the distance is very small, the presence of a free face increases the likelihood of spreading; beyond a certain distance, the absence of a free face leads to a sharply reduced likelihood. As a result, unlike PGA or GWD, the relationship between $L$ and lateral spreading is more threshold-like, and its shape function may exhibit sharper transitions. Based on this interpretation, we manually define the modified shape function for L as a step function, capturing the abrupt transition in lateral spreading likelihood beyond a certain distance from the river.

\subsubsection{Bivariate function modification}
In addition to modifying univariate functions, we also identified non-physical relationships in bivariate functions (e.g., see~\Cref{fig:interaction-example}). To address this, we synthesize a new bivariate function based on the modified univariate functions. Specifically, we define:
\begin{equation}
    f^{\text{syn}}_{i,j}(x_i,x_j) = f'_i(x_i) + f'_j(x_j),
\end{equation}
where $f'_i$ and $f'_j$ are the modified univariate shape functions described in the previous section. This additive form inherits the monotonic trends and preserves the feature-specific characteristics encoded in the univariate functions. For example, in the case of $L$, the threshold-like behavior is retained in the synthesized interaction.

To ensure that the synthesized function maintains a similar influence on model predictions as the original interaction term, we rescale it to match the original score range:
\begin{equation}
    f^{\text{syn, res}}_{i,j} = f^{\min}_{i,j} + \frac{(f^{\text{syn}}_{i,j} - f^{\text{syn, min}}_{i,j}) \cdot (f^{\max}_{i,j} - f^{\min}_{i,j})}{f^{\text{syn, max}}_{i,j} - f^{\text{syn, min}}_{i,j}}\,,
\end{equation}
where $f^{\min}_{i,j}$ and $f^{\max}_{i,j}$ are the minimum and maximum values of the original bivariate function, and $f^{\text{syn}}_{\min}$ and $f^{\text{syn}}_{\max}$ are the corresponding bounds of the synthesized function. This rescaling ensures that the contribution of the new function remains consistent with the original model, preventing disproportionate shifts in prediction scores.

Bivariate functions in EBM are represented as $30 \times 30$ score matrices, with each cell corresponding to a bin defined by the discretized ranges of the feature pair. To generate the synthesized matrix, we evaluate the modified univariate functions at the center of each bin and compute $f^{\text{syn}}_{i,j} = f'_i + f'_j$. This process preserves the original bin boundaries and makes the synthesized function fully compatible with the EBM framework.

\begin{figure}[h]
    \centering
    \begin{subfigure}[b]{0.30\textwidth}
        \centering
        \includegraphics[width=\textwidth]{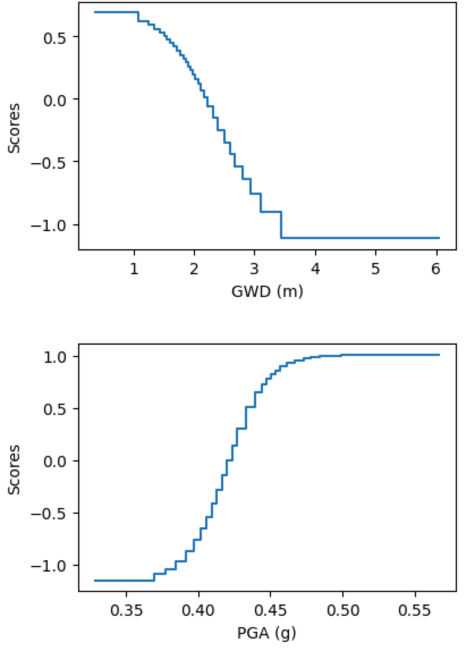}
        \caption{Binned $f'_{GWD}$ and $f'_{PGA}$.}
        \label{fig:gwd_pga_binned}
    \end{subfigure}
    \begin{subfigure}[b]{0.34\textwidth}
        \centering
        \includegraphics[width=\textwidth]{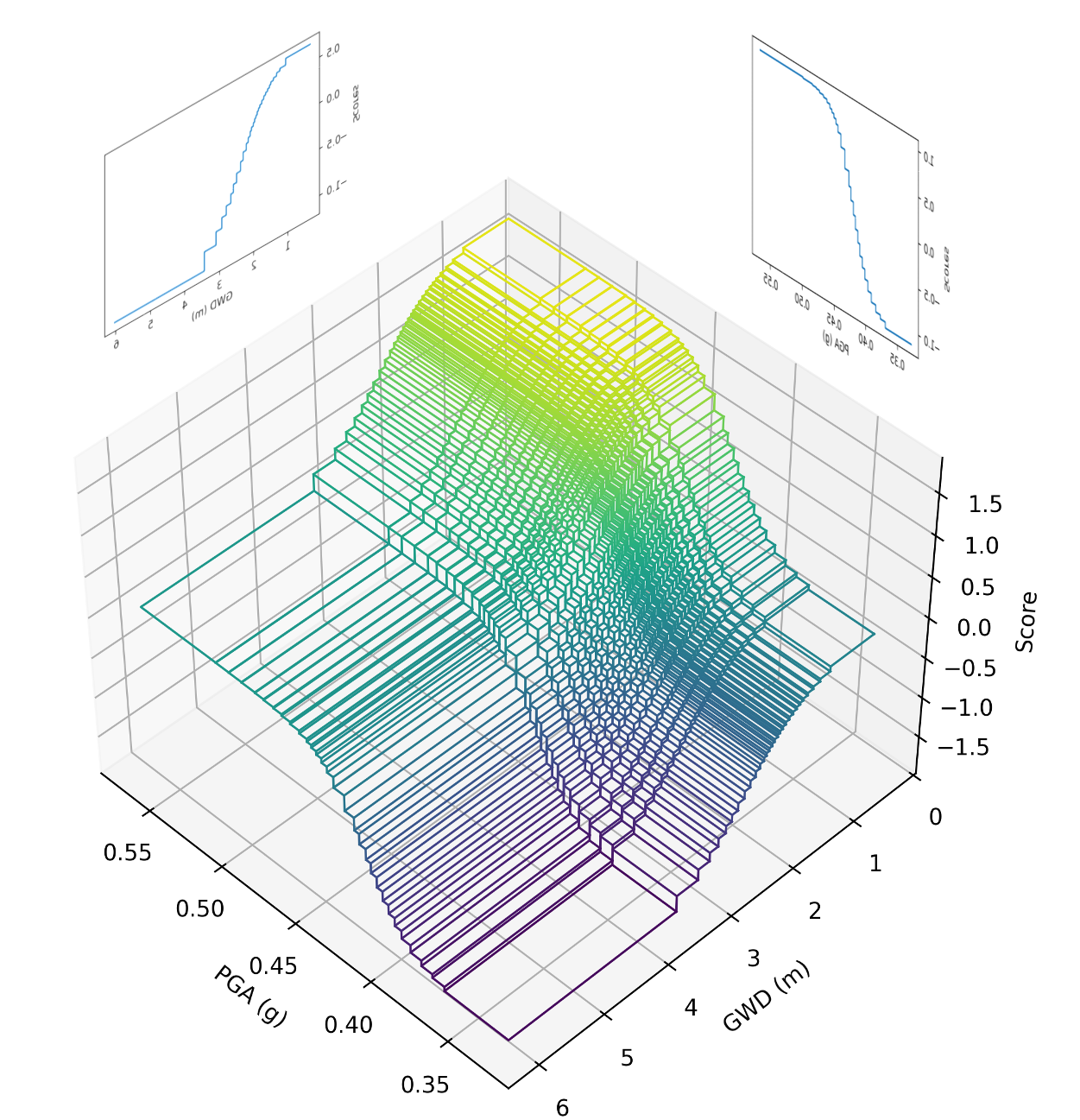}
        \caption{$f^{\text{syn}}_{GWD,PGA}=f'_{GWD}+f'_{PGA}$.}
        \label{fig:gwd_pga_wireframe}
    \end{subfigure}
    \begin{subfigure}[b]{0.34\textwidth}
        \centering
        \includegraphics[width=\textwidth]{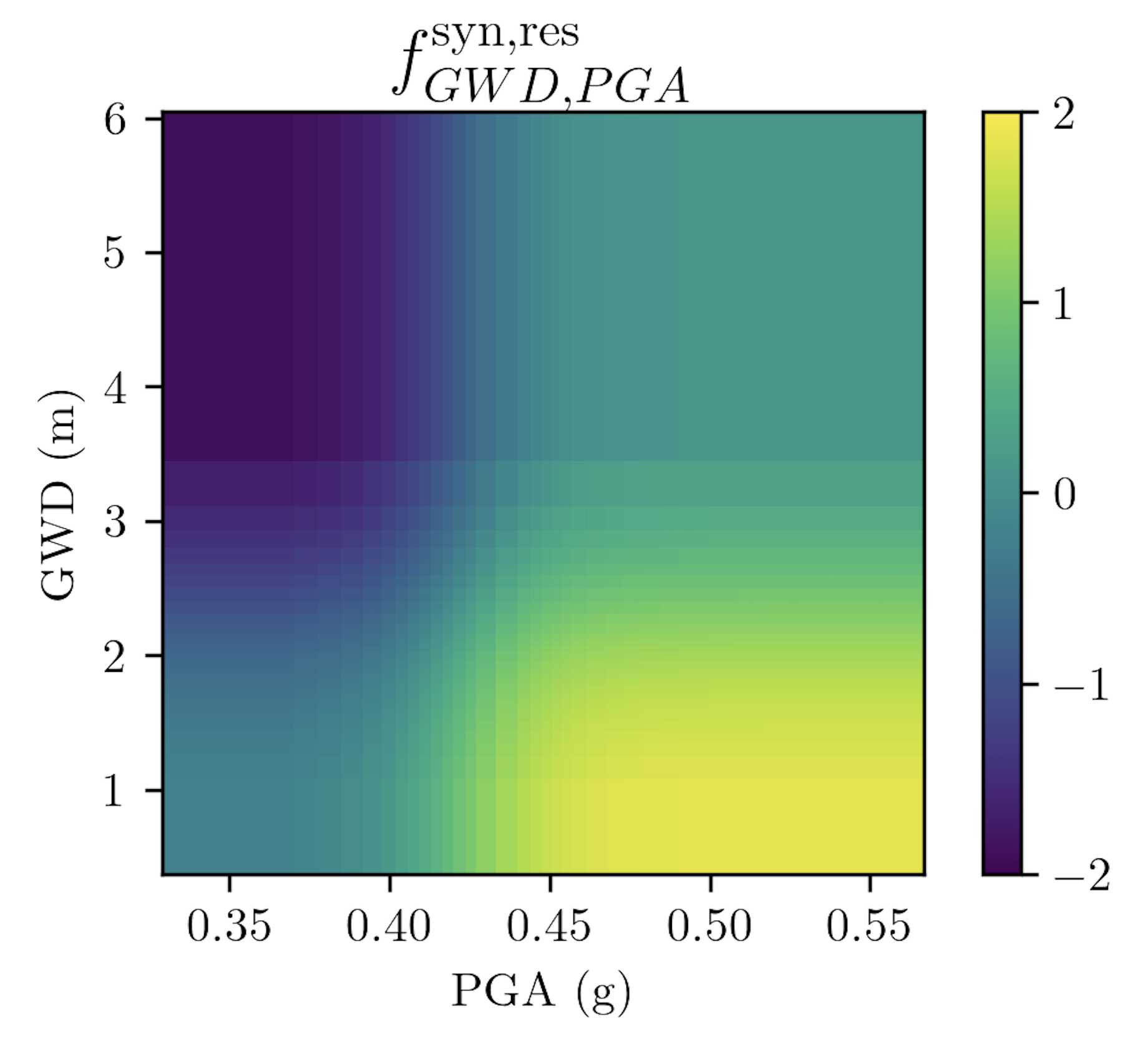}
        \caption{$f^{\text{syn, res}}_{GWD,PGA}$}
        \label{fig:gwd_pga_synthesized}
    \end{subfigure}    
    \caption{Synthesis of the modified bivariate function for GWD and PGA.}
    \label{fig:modify-interaction}
\end{figure}

\Cref{fig:modify-interaction} provides an example of this synthesis for GWD and PGA. The modified univariate functions are first discretized (as shown in~\Cref{fig:gwd_pga_binned}), and then summed to construct the bivariate function (shown in~\Cref{fig:gwd_pga_wireframe}). The final result is expressed as a rescaled $30 \times 30$ score matrix (see~\Cref{fig:gwd_pga_synthesized}).

This synthesized function aligns with physical expectations. For example, shallow GWD combined with high PGA correctly yields a high positive score. However, directly replacing bivariate function with the synthesized one would eliminate all original interaction effects learned by the EBM, resulting in a model that purely depends on univariate functions. To preserve some of the useful data-driven patterns, we develop a selective replacement strategy that only modifies regions of the original interaction function that significantly deviate from the synthesized version. The replacement rule is defined as
\begin{equation}
    f'_{i,j}[m,n]=
    \begin{cases}
        f^{\text{syn,res}}_{i,j}[m,n], & \text{if } \Delta_{i,j}[m,n] > \epsilon \\
        f_{i,j}[m,n], & \text{otherwise}
    \end{cases} \,,
\end{equation}
where $m$ and $n$ denote matrix indices, $\Delta_{i,j}[m,n]$ is an element-wise discrepancy measure between synthesized and original interaction values, and $\epsilon$ is a user-defined threshold. We consider two different definitions of $\Delta_{i,j}[m,n]$ which lead to different replacement patterns:
\begin{enumerate}
    \item \textbf{Relative error}: percent difference between the original and synthesized values relative to the original value:
    \begin{equation}
    \label{eq:rel_diff}
    \Delta^{\text{rel}}_{i,j}[m,n] = \left|\frac{f^{\text{syn,res}}_{i,j}[m,n] - f_{i,j}[m,n]}{f_{i,j}[m,n]}\right|,
    \end{equation}
    where $m$ and $n$ denote matrix indices.
    
    \item \textbf{Range-based error}: percent difference between the original and synthesized values relative to the range of the original function:
    \begin{equation}
    \label{eq:abs_diff}
    \Delta^{\text{rng}}_{i,j}[m,n]=\left|\frac{f^{\text{syn,res}}_{i,j}[m,n] - f_{i,j}[m,n]}{f^{\max}_{i,j}-f^{\min}_{i,j}}\right|
    \end{equation}
\end{enumerate}

The value of $\epsilon$ controls the sensitivity of the replacement: a lower threshold results in more aggressive correction of the original function, replacing a larger number of elements that deviate even slightly from the synthesized version. On the other hand, a higher $\epsilon$ makes the method more conservative, preserving more of the original EBM interaction and only replacing values with large deviations.

By employing these thresholding methods, we aim to correct unphysical scores while minimizing unnecessary changes to the learned interaction. The choice between the two approaches depends on the characteristics of the data and the desired degree of preservation of the original interaction effects.

\begin{figure}[H]
    \centering
    \begin{subfigure}[b]{0.7\textwidth}
        \centering
        \includegraphics[width=\textwidth]{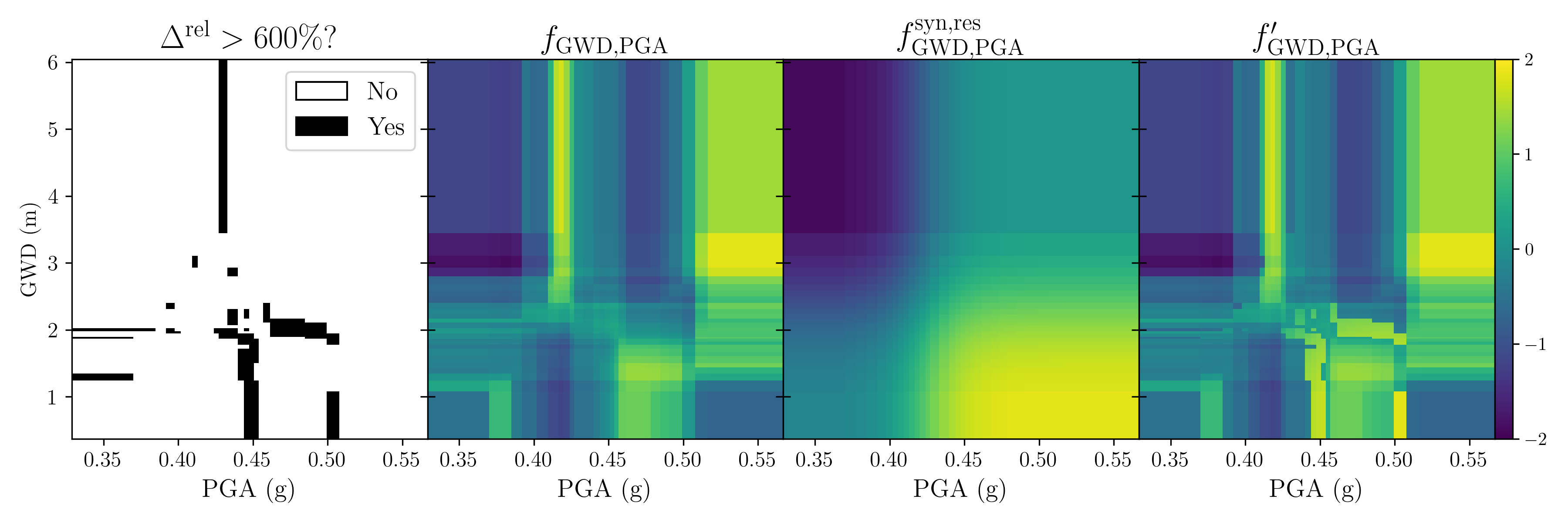}
        \caption{Relative error method with a threshold of 600\% for $f_{GWD,PGA}$.}
        \label{fig:mixing-map-gwd-pga-rel}
    \end{subfigure}
    \vfill
    \begin{subfigure}[b]{0.7\textwidth}
        \centering
        \includegraphics[width=\textwidth]{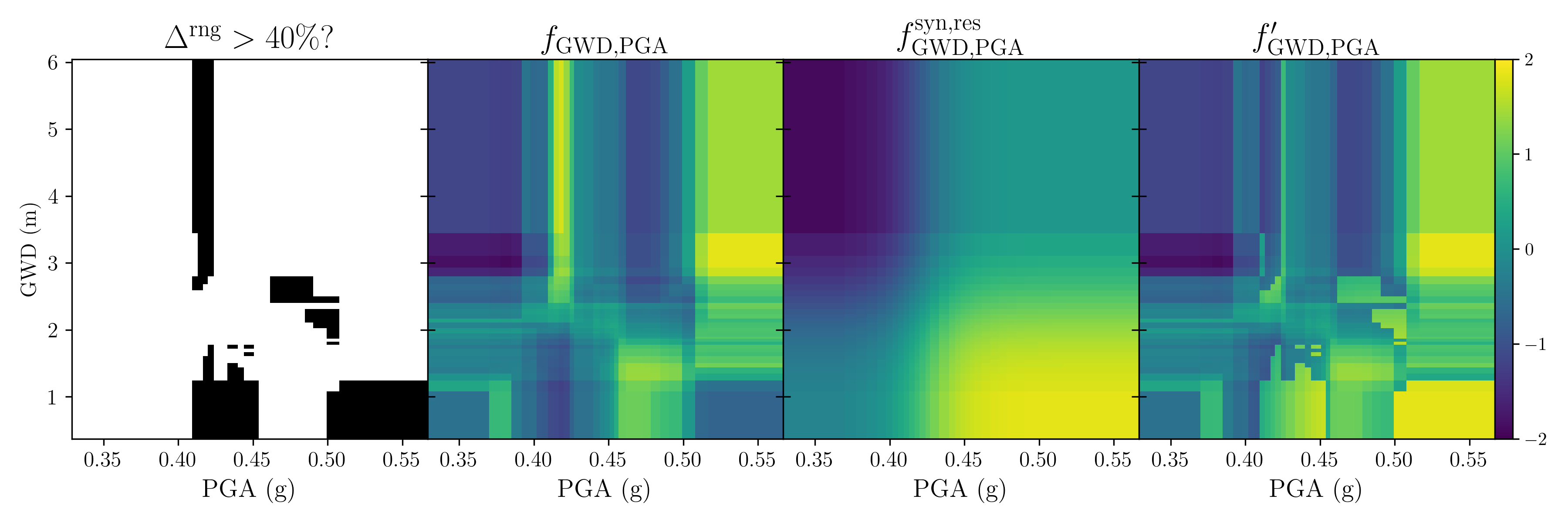}
        \caption{Range-based method with a threshold of 40\% for $f_{GWD,PGA}$.}
        \label{fig:mixing-map-gwd-pga}
    \end{subfigure}
    \vfill
    \begin{subfigure}[b]{0.7\textwidth}
        \centering
        \includegraphics[width=\textwidth]{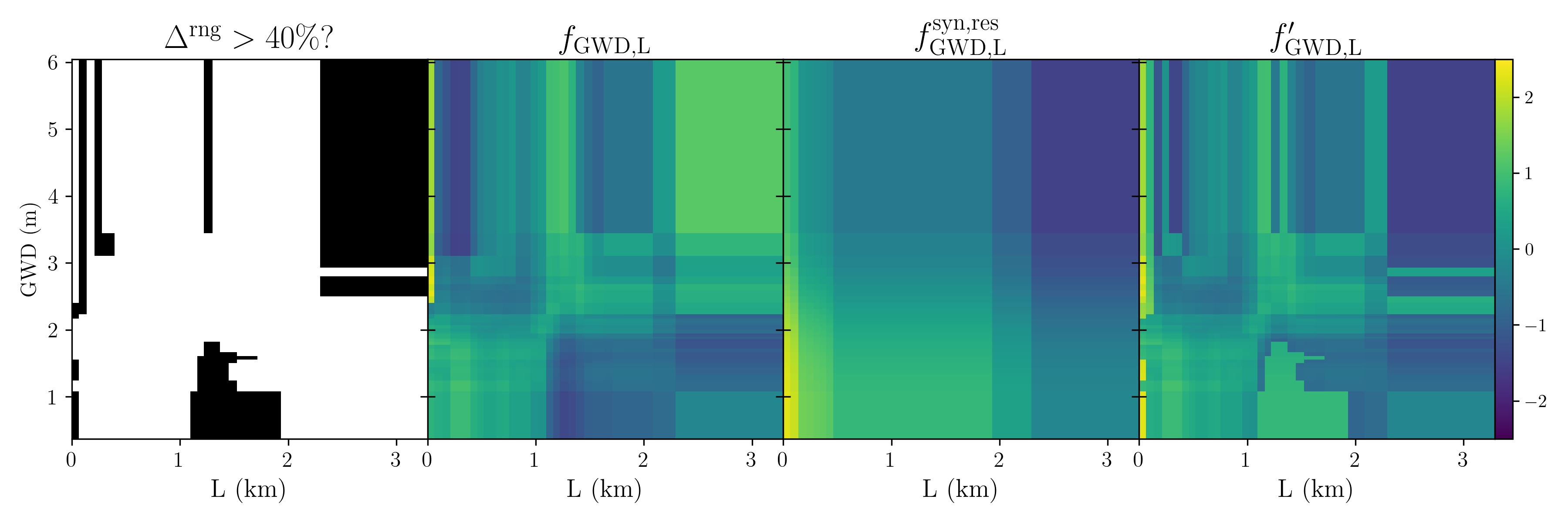}
        \caption{Range-based method with a threshold of 40\% for $f_{GWD,L}$.}
        \label{fig:mixing-map-gwd-l}
    \end{subfigure}
    \vfill
    \begin{subfigure}[b]{0.7\textwidth}
        \centering
        \includegraphics[width=\textwidth]{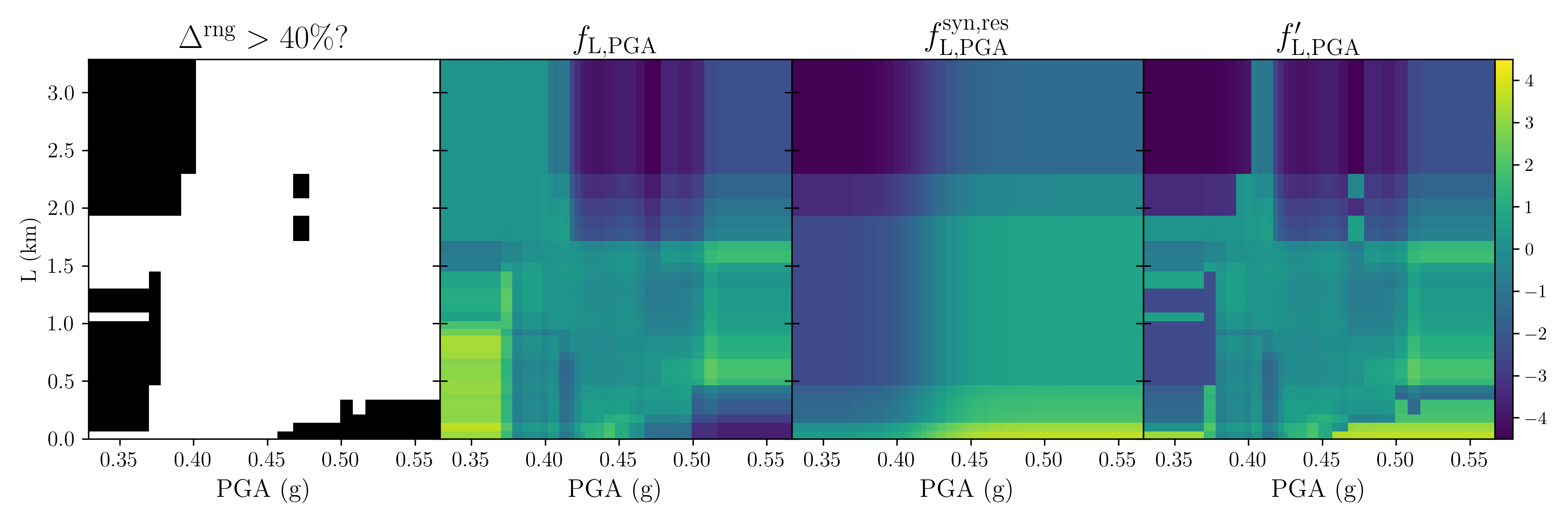}
        \caption{Range-based method with a threshold of 40\% for $f_{L,PGA}$.}
        \label{fig:mixing-map-l-pga}
    \end{subfigure}
    \caption{Mixing the original interaction map and the synthetic map.}
    \label{fig:mixing-map}
\end{figure}

\Cref{fig:mixing-map-gwd-pga-rel,fig:mixing-map-gwd-pga} illustrate the results of both thresholding methods applied to the GWD–PGA bivariate function. In each case, the first panel highlights the matrix elements that exceed the replacement threshold (shown in black). The second and third panels show the original EBM interaction function and the synthesized function, respectively. The difference between these two functions is used to determine which elements will be replaced. The final panel on the right displays the resulting bivariate function after selective replacement.

Using a relative error threshold of 600 \% and a range-based threshold of 40 \%, both thresholds modify approximately 12 \% of the 30 × 30 interaction matrix elements. This replacement proportion was intentionally controlled to enable a fair comparison of how effectively each method corrects the previously identified unphysical regions (see~\Cref{fig:interaction-example}) while preserving most of the learned interaction. 

We observe that the relative error method is highly sensitive and therefore requires a very large threshold (e.g., 600\%). This sensitivity is because the discrepancy is normalized by the original value, which makes the criterion easily triggered when the original value is close to zero. In contrast, the range-based method is normalized by the range of the original function, giving all elements similar sensitivity regardless of their individual magnitudes.

Although the amount of replaced elements is similar for the two approaches, their patterns are different. The relative error criterion primarily change regions where the original scores are small and therefore does not effectively remove the unphysical behavior we mentioned in~\Cref{fig:interaction-example}. In contrast, the range-based method concentrates replacements within the problematic regions. The final bivariate function that is more consistent with physical expectations while preserving most learned interactions.

Based on this comparison, we selected the range-based method and applied it to the GWD–L and L–PGA bivariate functions using the same 40\% threshold (see~\Cref{fig:mixing-map-gwd-l,fig:mixing-map-l-pga}). The results show that the method effectively corrects the unphysical positive scores in regions with high $L$ and high GWD, while increasing the positive scores in regions with very low $L$ and shallow GWD. Additionally, the method corrects the unphysical positive scores in low PGA regions and the unphysical negative scores in regions with high PGA and low $L$.

After modifying both univariate and bivariate functions, we construct a hybrid model, referred to as the domain-informed EBM. The domain-informed EBM retains data-driven patterns while better aligning with domain knowledge. In the following section, we evaluate the performance and reliability of the domain-informed model.

\section{Results} \label{sect:results}
\subsection{Performance Degradation}

To evaluate the impact of incorporating domain knowledge, we compare the predictive performance of the original EBM with the domain-informed EBM. We use the same evaluation metrics as described in~\Cref{sect:meth-eval}, including accuracy, precision, recall, F1-score, and AUC.

As shown in~\Cref{tab:performance_compare}, the domain-informed EBM exhibits a modest degradation in predictive performance across most metrics: accuracy decreases by 0.05, precision by 0.10, F1-score by 0.04, and AUC by 0.08. This performance drop is expected, as the modifications prioritize alignment with domain knowledge rather than optimizing predictive loss.

\begin{figure}[h]
    \centering
    \begin{subfigure}[b]{0.45\textwidth}
        \centering
        \includegraphics[width=\textwidth]{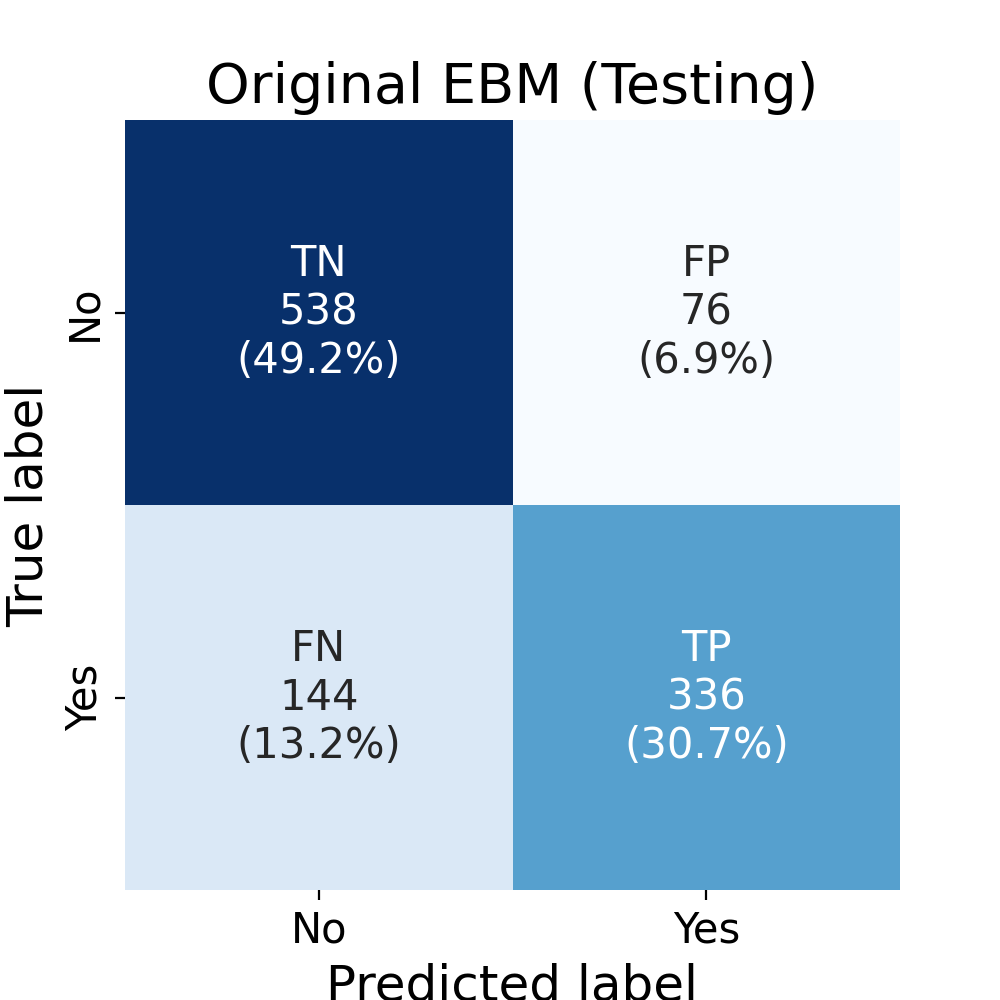}
        \caption{Original EBM.}
        \label{fig:confusion-orig}
    \end{subfigure}
    \begin{subfigure}[b]{0.45\textwidth}
        \centering
        \includegraphics[width=\textwidth]{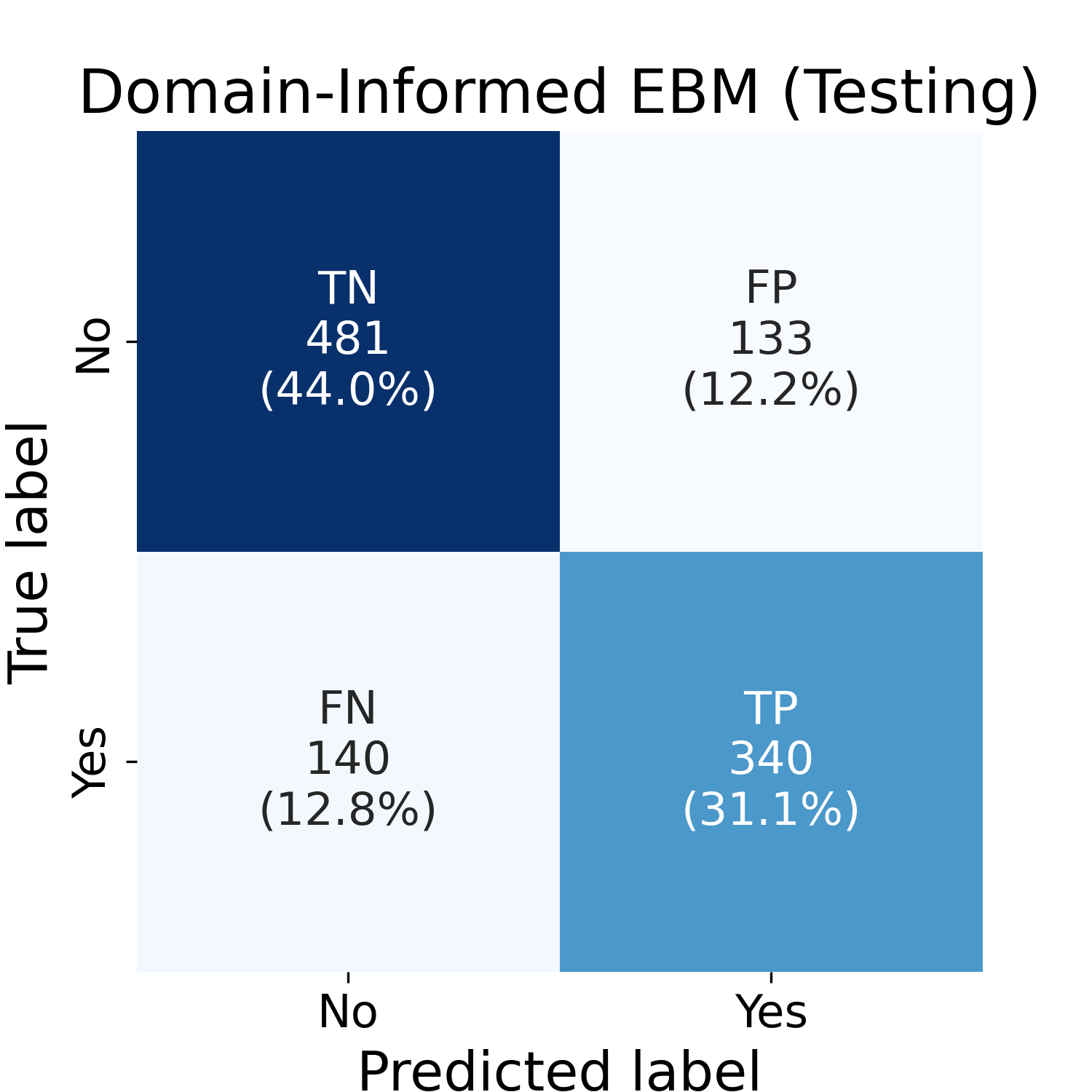}
        \caption{Domain-Informed EBM.}
        \label{fig:confusion-di}
    \end{subfigure}
    \caption{ Confusion matrices of the original EBM and the domain-informed EBM on the testing dataset.}
    \label{fig:confusion}
\end{figure}

To better understand the source of this degradation, we examine the confusion matrices for both models using the testing dataset (\Cref{fig:confusion}). The primary cause of the performance drop is an increase in false positives, which rise from 6.9\% to 12.2\%. In contrast, false negatives slightly decrease from 13.2\% to 12.8\%. 

These results highlight a fundamental trade-off: while the Domain-Informed EBM sacrifices some predictive accuracy, it gains consistency with known physical behavior. In the following sections, we investigate whether the predictions of the modified model are more meaningful through global and local explanation analysis.

\begin{table}[h]
\caption{EBM models performance on validation and testing dataset.}
\label{tab:performance_compare}
\centering
    \begin{tabular}{lcccc}
    \toprule
     & \multicolumn{2}{c}{Original EBM} & \multicolumn{2}{c}{Domain-Informed EBM} \\
     \cmidrule{2-5}
    Dataset & Validation & Testing & Validation & Testing \\ \midrule
    Accuracy & 0.80 & 0.80 & 0.76 & 0.75 \\
    Precision & 0.80 & 0.82 & 0.72 & 0.72 \\
    Recall & 0.71 & 0.70 & 0.70 & 0.71 \\
    F1-score & 0.76 & 0.75 & 0.71 & 0.71 \\
    AUC & 0.87 & 0.88 & 0.80 & 0.80\\
    \bottomrule
    \end{tabular}
\end{table}

\subsection{Global Explanation}
\begin{figure}[h]
    \centering
    \begin{subfigure}[b]{0.45\textwidth}
        \centering
        \includegraphics[width=\textwidth]{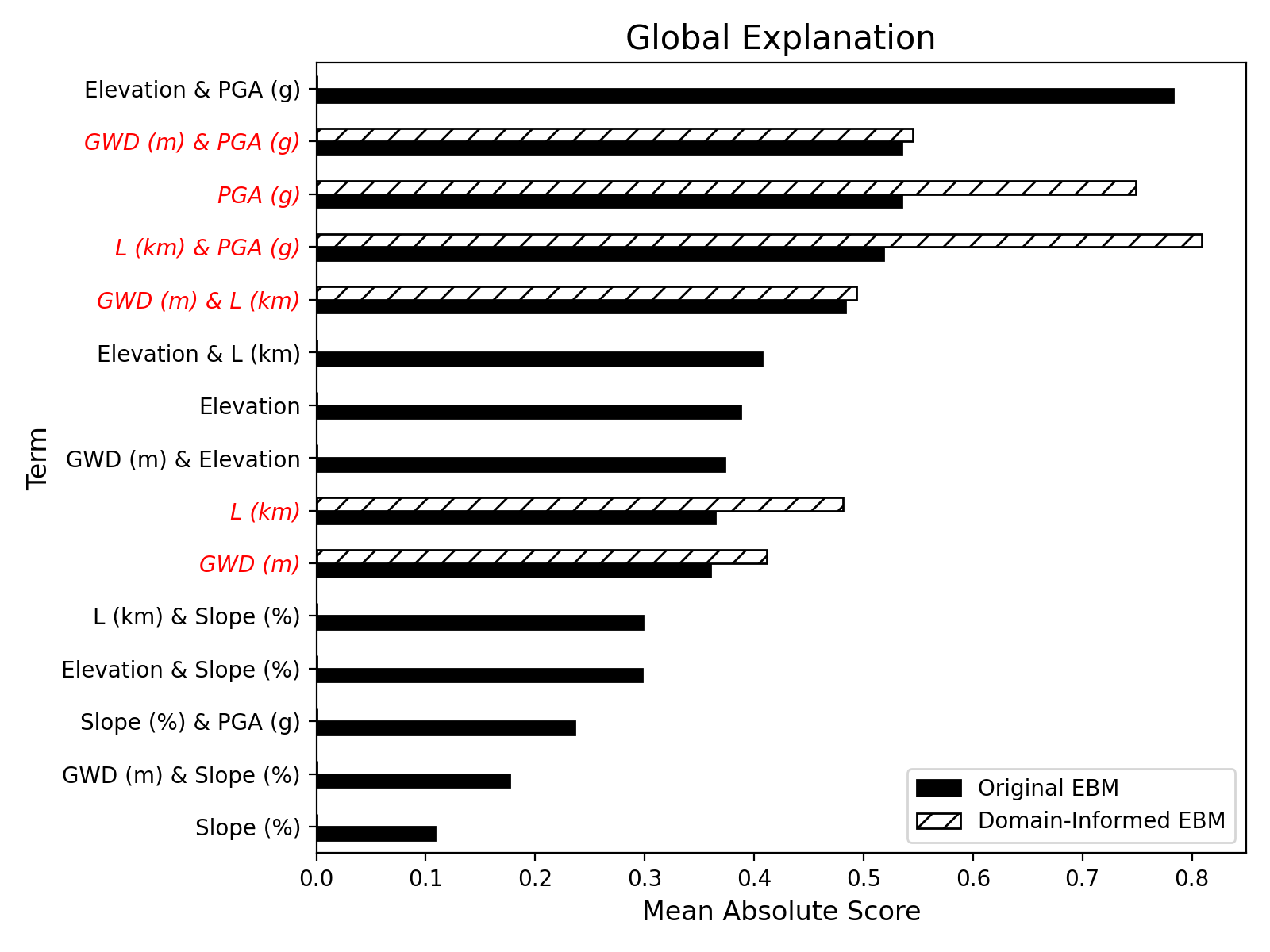}
        \caption{Mean Absolute Score Comparison (sorted by original EBM scores).}
        \label{fig:global-exp-compare}
    \end{subfigure}
    \begin{subfigure}[b]{0.45\textwidth}
        \centering
        \includegraphics[width=\textwidth]{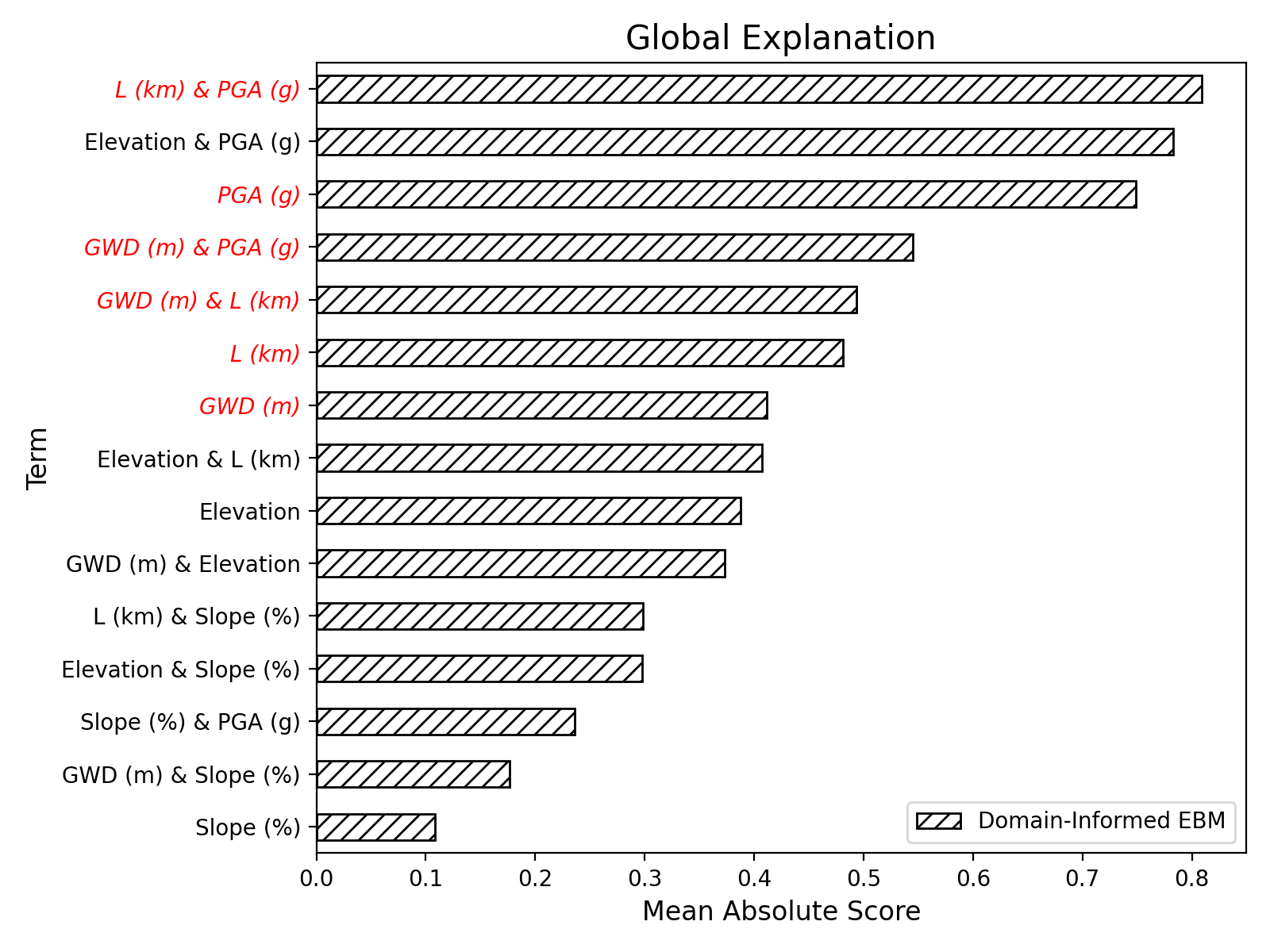}
        \caption{Mean Absolute Score of Domain-Informed EBM.}
        \label{fig:global-exp-di}
    \end{subfigure}
    \caption{Feature importance in the training data.}
    \label{fig:global-exp}
\end{figure}

In the global explanation, we rank features based on their importance in predicting the result across the entire training set. The feature importance is computed by averaging the absolute scores of a feature across the predictions. The higher rank in feature importance means the feature has more impact on predictions.~\Cref{fig:global-exp-compare} shows the feature importance in the original and domain-informed EBM, sorted by feature importance in the original EBM. Additionally, we plot the importance of the domain-informed EBM separately in \Cref{fig:global-exp-di}, showing the ranking in the domain-informed EBM.

In \Cref{fig:global-exp-compare}, the original EBM ranks the elevation-PGA interaction as the most important, with a 0.78 mean absolute score. This means the elevation-PGA interaction term, on average, contributes +0.78 or -0.78 to each prediction. We also notice a significant gap in feature importance between elevation-PGA and the second important feature, GWD-PGA. This indicates elevation-PGA is a very decisive feature in the original model. However, the relationship between lateral spreading occurrence and elevation is not well understood. This raises the concern of using the model as the predictions are highly affected by an obscure feature. 

On the other hand, the domain-informed EBM lessens this concern as the importance of the features incorporating domain knowledge increase after modification. For example, the feature importance of the PGA increases from 0.53 to 0.75. Among the modified features, the importance of the $L$-PGA interaction changes most significantly, from 0.52 to 0.81, becoming the most important feature (see \Cref{fig:global-exp-di}).

\begin{figure}[h]
    \centering
    \includegraphics[width=\textwidth]{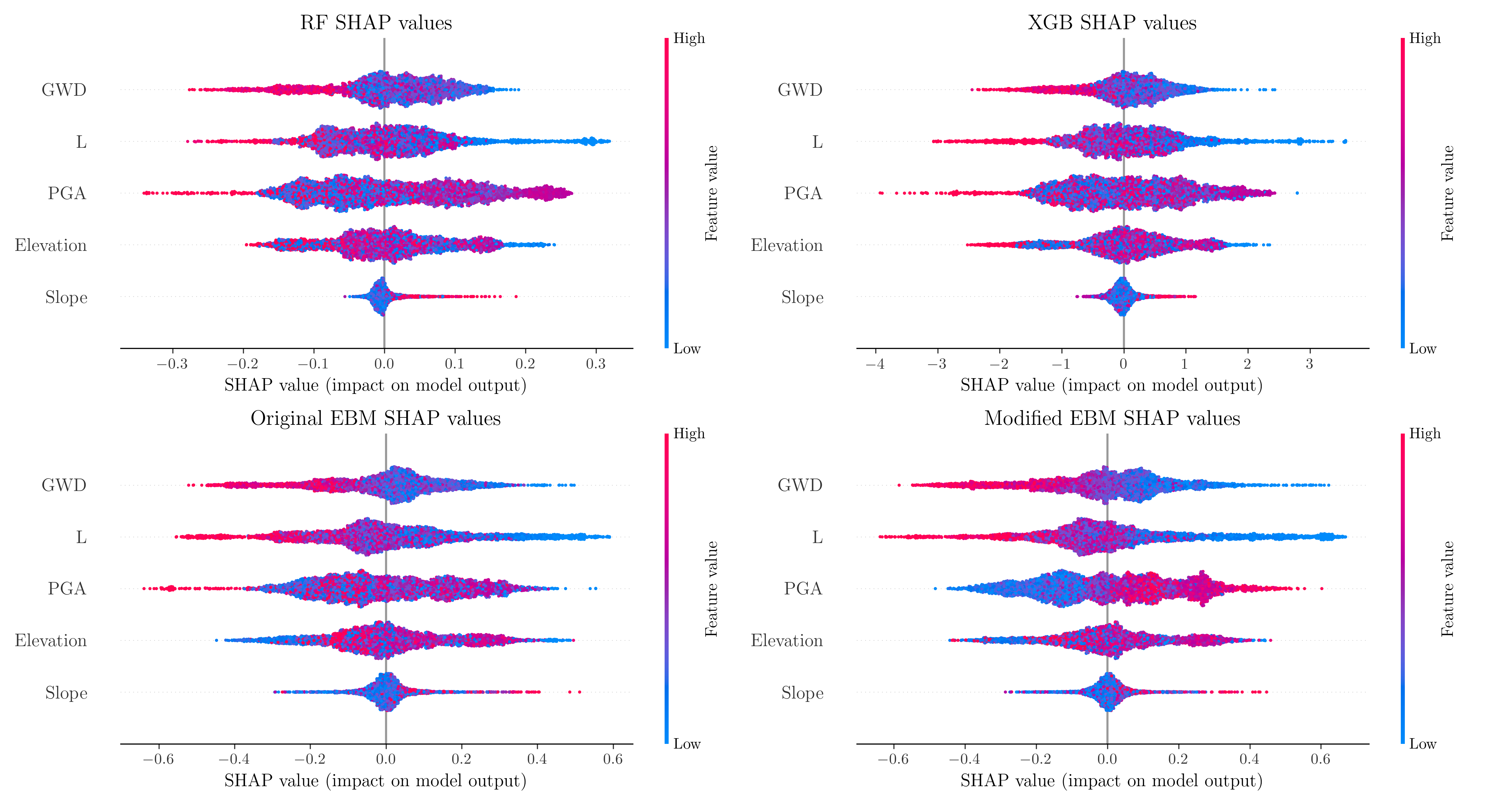}
    \caption{Beeswarm plot for RF (top left), XGB (top right), Original EBM (bottom left), and Domain-informed EBM (bottom right). Each dot in the plot represents a datapoint from the training data. The color denotes the feature values, and the x-axis denotes the SHAP values}
    \label{fig:shap-beeswarm}
\end{figure}

To more intuitively compare the global explanations of black-box models and the developed EBMs, we apply SHAP analysis to the black-box models (RF and XGB), as well as to both the original and domain-informed EBMs. SHAP values provide a model-agnostic approach to estimating the contribution of each feature to a prediction. We visualize these values using beeswarm plots (see~\Cref{fig:shap-beeswarm}), which show how feature values influence their SHAP contributions across all training datapoints. In this context, high SHAP values indicate a positive contribution to the likelihood of lateral spreading, analogous to high EBM scores.

In the PGA beeswarm plots, we observe unphysical behavior in both RF and XGB models: datapoints with high PGA (colored red) often contribute negatively to the prediction, while some datapoints with low PGA (colored blue) contribute positively. This contradicts domain knowledge, which expects higher PGA to increase the likelihood of lateral spreading.

We observe similar unphysical patterns in the PGA SHAP values computed for the original EBM. While EBM is inherently transparent, SHAP summarizes the combined effect of each feature, incorporating both its own score curve and any relevant interaction terms. In contrast, the domain-informed EBM produces a more physically consistent SHAP distribution: PGA contributions increase monotonically with PGA values, as expected. This demonstrates that the domain-informed modifications help correct previously unphysical patterns in the model's learned behavior.

\begin{figure}[h]
    \centering
    \begin{subfigure}[b]{0.45\textwidth}
        \centering
        \includegraphics[width=\textwidth]{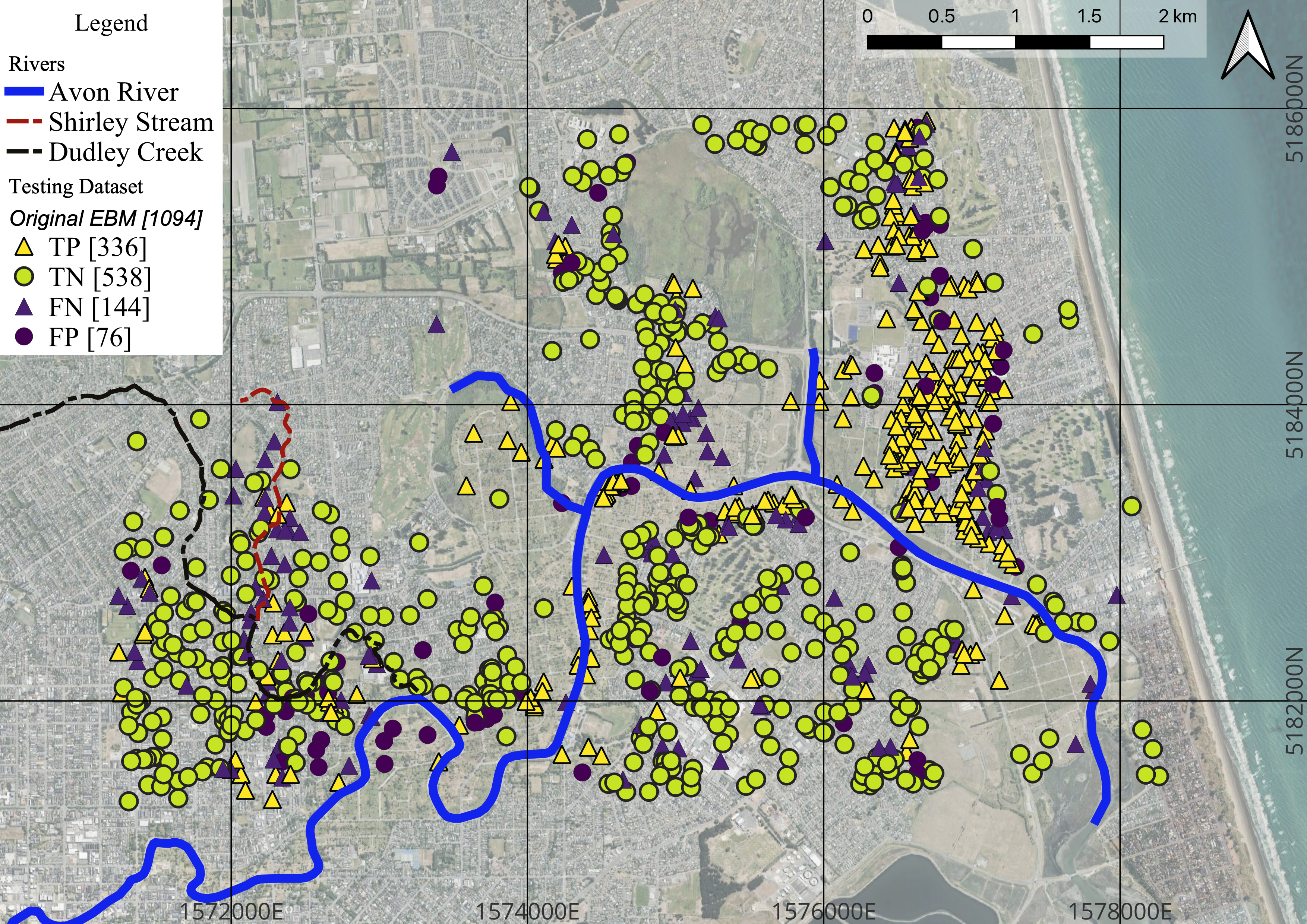}
        \caption{Original EBM.}
        \label{fig:prediction-map-orig}
    \end{subfigure}
    \hfill
    \begin{subfigure}[b]{0.45\textwidth}
        \centering
        \includegraphics[width=\textwidth]{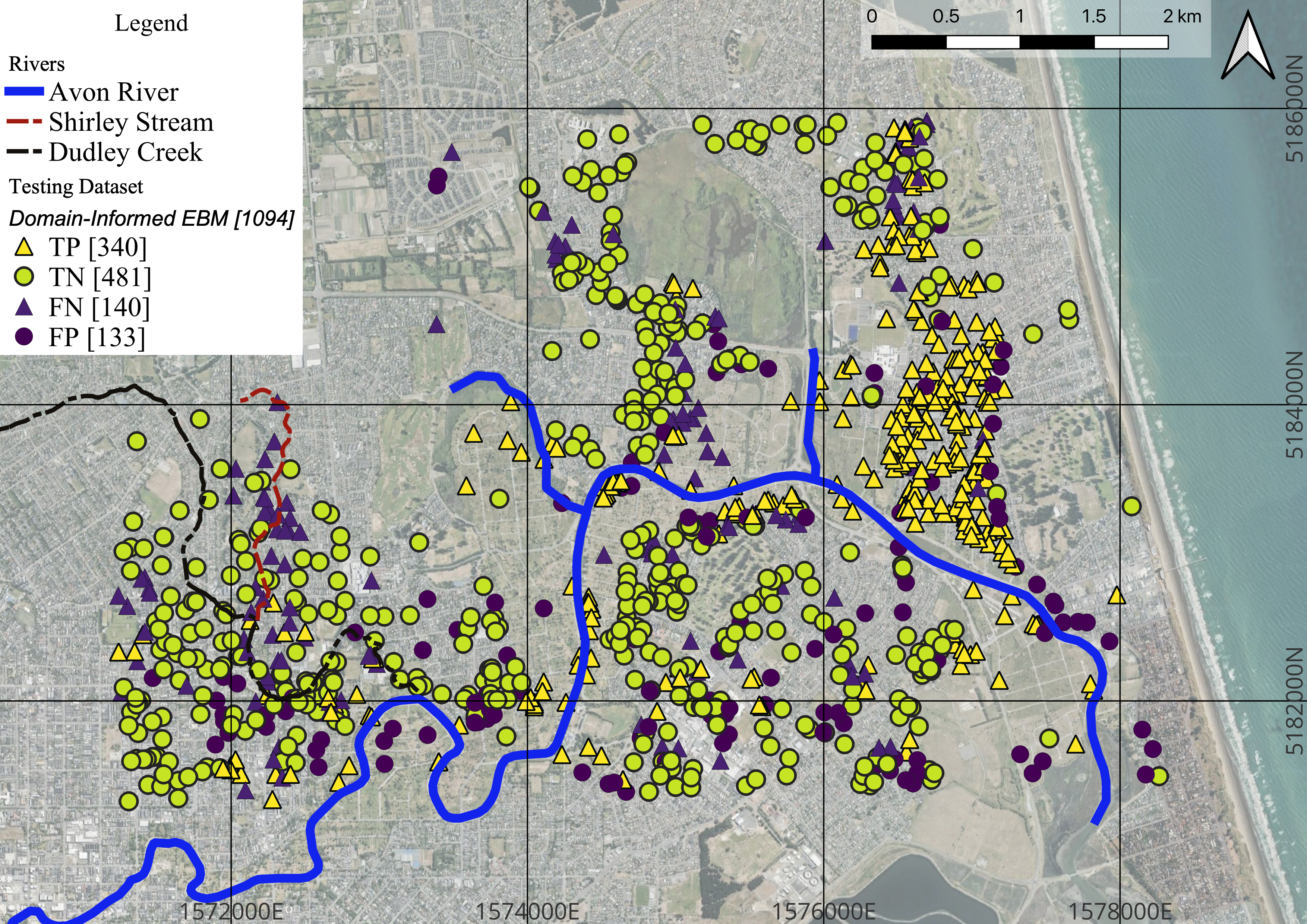}
        \caption{Domain-Informed EBM.}
        \label{fig:prediction-map-di}
    \end{subfigure}
    \vfill
    \begin{subfigure}[b]{0.45\textwidth}
        \centering
        \includegraphics[width=\textwidth]{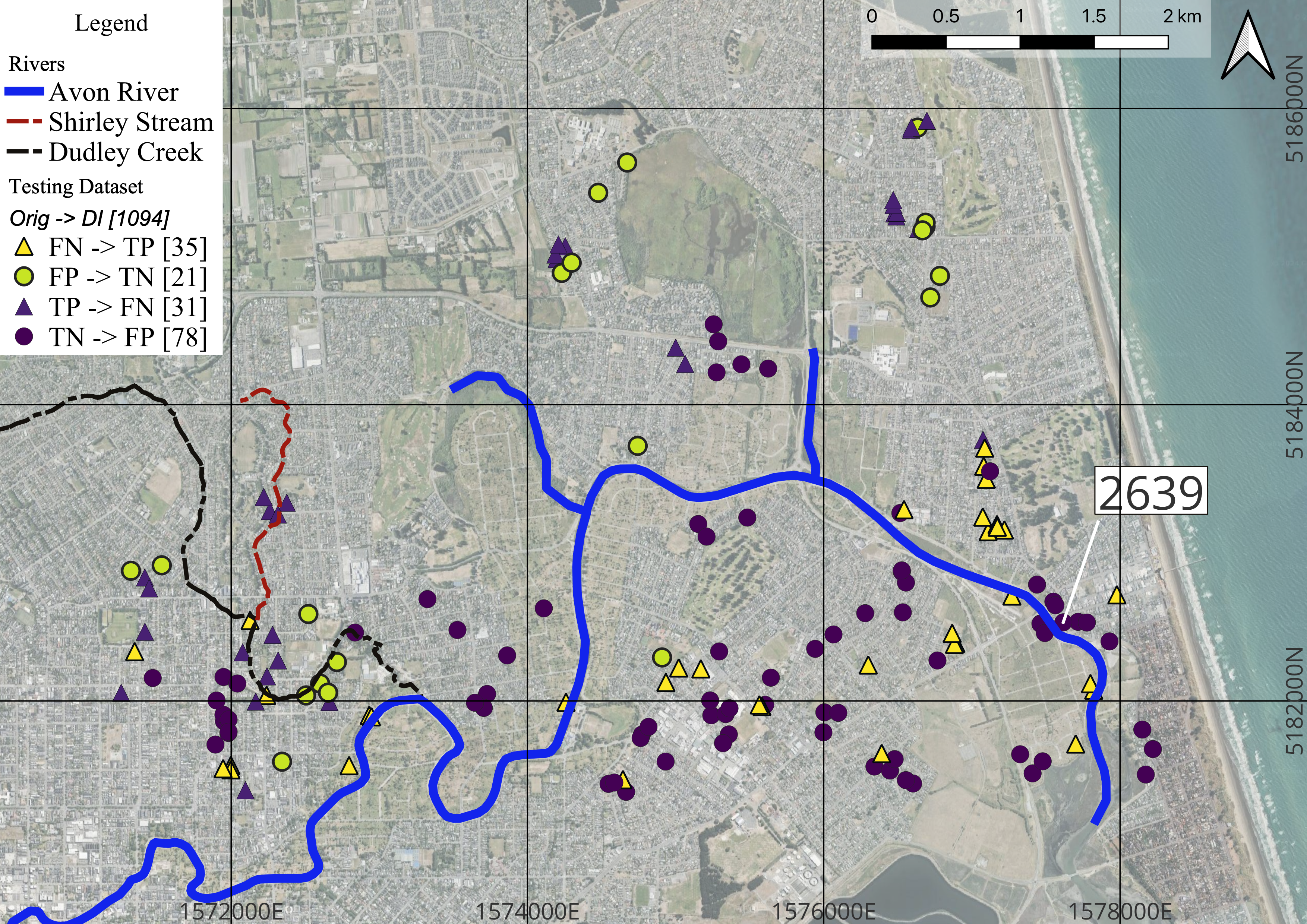}
        \caption{Difference between two models.}
        \label{fig:prediction-map-change}
    \end{subfigure}
    \caption{The distribution of true positive (TP), true negative (TN), false positive (FP), and false negative (FN) results for (a) original EBM and (b) domain-informed EBM. (c) The improved and degraded cases after the modification. The label XX-\>YY denotes the prediction changing from XX to YY after the modification (e.g., an FP change to TN is FP -\> TN.}
    \label{fig:prediction-map}
\end{figure}

Additionally, we map the prediction distributions of the testing data for the original and domain-informed EBM in \Cref{fig:prediction-map-orig,fig:prediction-map-di}. Triangles represent positive cases, while circles denote negatives. Light colors indicate correct predictions, and dark colors indicate errors. Overall, the domain-informed EBM's predictions are similar to the original EBM.

The modification corrects some predictions and introduces new errors.
Among sites originally misclassified by the original EBM, 35 false negatives (FN) become true positives (TP), and 21 false positives (FP) become true negatives (TN).
However, among sites originally classified correctly, 31 true positives (TP) become false negatives (FN), and 78 of true negatives (TN) become false positives (FP).
\Cref{fig:prediction-map-change} highlights these changes.
Triangles represent positive cases and circles represent negatives.
Light colors denote improved predictions; dark colors indicate degraded predictions.

This analysis provides a broad overview of the model's performance changes. However, we need to explore the local explanation to gain a deeper understanding of the reasons behind these prediction changes and errors. This next step will allow us to examine the factors influencing the model's decisions and how the domain-informed modifications have altered these influences.

\subsection{Local Explanation}
With local explanation, we analyze a site that is classified differently by the original EBM and the domain-informed EBM. Site 2639, highlighted in~\Cref{fig:prediction-map-change}, did not experience lateral spreading during the earthquake. However, it is located very close to the river, with an $L$ of 0.09 km, and was subjected to strong ground motion, with a PGA of 0.52 g. The original EBM correctly classifies it as a no-lateral-spreading site, while the domain-informed EBM incorrectly classifies it as a lateral-spreading site. 

\begin{figure}[h]
    \centering
    \includegraphics[width=0.55\textwidth]{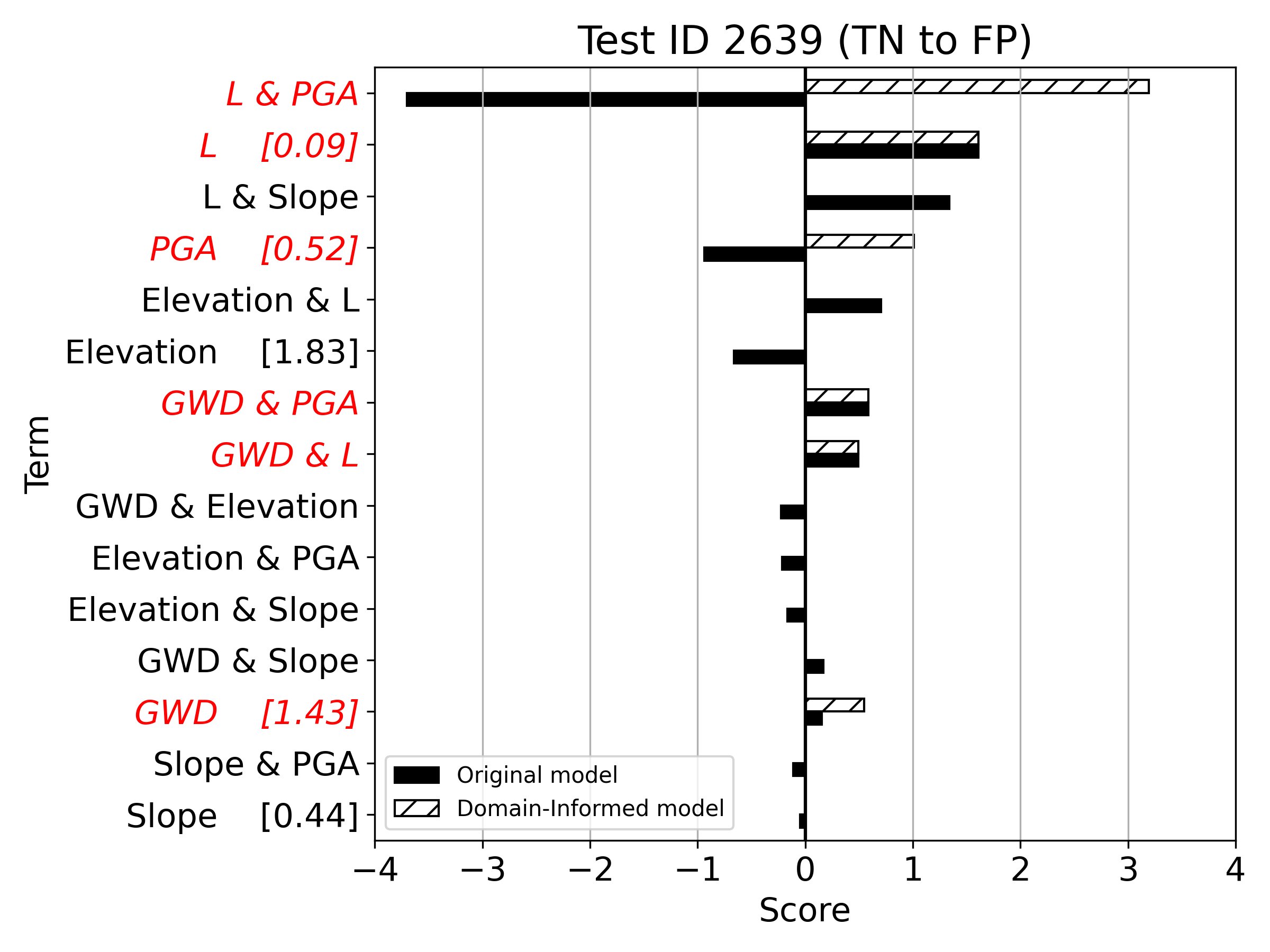}
    \caption{Local explanation for Site 2639. The modified features are marked in \textit{\textcolor{red}{italic}} style, and the feature values are shown in the brackets.}
    \label{fig:local-explanation}
\end{figure}

To understand why the models make these predictions, we present the feature scores for this site in~\Cref{fig:local-explanation}. The solid bars represent the feature scores of the original EBM, while the textured bars represent the feature scores of the domain-informed EBM. Note that we simplify the bar chart by omitting the unmodified feature scores of the domain-informed EBM, as they remain identical to those in the original EBM. The features are sorted by their importance in the original EBM, with the most influential feature appearing at the top.

In the original EBM, the most important feature is the $L$-PGA interaction, with a score of $-3.7$. This significant negative value suggests that a site experiencing high PGA is very unlikely to experience lateral spreading. However, this explanation contradicts our understanding of physics. In contrast, the domain-informed EBM also identifies the $L$-PGA interaction as the most important feature but assigns it a positive score of $+3.2$. This indicates that the domain-informed EBM considers proximity to the river and high PGA as indicative of a high likelihood of lateral spreading.

Through this local explanation, we observe that the original EBM exhibits physically inconsistent behavior, which is identified and addressed by incorporating domain knowledge.
Although the domain-informed EBM's prediction for this site is incorrect, its reasoning aligns with domain knowledge, improving interpretability and trust.

We argue that the site may be governed by other factors not included in the model inputs, such as soil types. Since this information is unavailable to the model, the original EBM can only minimize residual error using the available features, which may produce a correct prediction for the wrong physical reason.

Similar compensation effects can occur in black-box models. However, their feature-prediction relationship is not directly observable, preventing us from identifying such inconsistencies or direct correcting the learned relationships. This limitation highlights the importance of this study and the value of domain-informed and interpretable model.

\section{Conclusion}
\label{sect:conclusion}
This study demonstrates a novel approach to improve the trustworthiness and interpretability of machine learning models in natural hazard research, specifically for predicting lateral spreading using the Explainable Boosting Machine (EBM). The EBM's additive structure enables a full human understanding of its decision-making logic, addressing the `black-box' issue that is common in many ML models.

While the original EBM achieved 79.9\% testing accuracy, it exhibited non-physical behaviors in certain feature ranges. To address this, we developed a framework to integrate physics knowledge into the EBM. This involved applying monotonicity to univariate functions (GWD, L, PGA), synthesizing bivariate functions from modified univariate function, and selectively replacing regions in the original bivariate function that significantly deviate from domain-knowledge expectations. The domain-informed model maintained 75.0\% testing accuracy while aligning better with domain knowledge.

The study's enhanced explainability provided valuable insights into model behavior. Global explanation revealed the increased influence of modified functions and raised concerns about the physically opaque yet decisive feature in the original EBM. Local explanations showed how incorporating domain knowledge corrected non-physical behavior and provided a more logical explanation for individual predictions.

These results highlight the potential of interpretable glass-box models like EBM. They can fully explain decision-making processes. This allows domain experts to evaluate and refine the model, improving trustworthiness.

While limitations exist, including reduced accuracy from modifications and incomplete capture of expert knowledge nuances, this initial knowledge integration demonstrates the approach's potential. Future work should focus on refining the method through function optimization. Overall, integrating physics constraints into transparent ML models represents a promising direction for developing reliable and trustworthy models in natural hazard research, enhancing their applicability and acceptance in critical decision-making scenarios.

\section*{Data Availability}
The code and data used in this study are publicly available at DesignSafe Data Depot (\url{https://www.designsafe-ci.org/data/browser/public/designsafe.storage.published/PRJ-6270}) and GitHub (\url{https://github.com/chhsiao93/di-ebm-lateral-spreading}).
\section*{Acknowledgments}
The authors thank the support of NHERI DesignSafe-CI for computing resources and data storage~\citep{RathjeDesignSafe}.
\bibliographystyle{apalike}  
\bibliography{main}

\begin{thebibliography}{}

\bibitem[Breiman, 2001]{breiman-2001}
Breiman, L. (2001).
\newblock {Random Forests}.
\newblock {\em Machine Learning}, 45(1):5--32.

\bibitem[Can et~al., 2021]{can-2021}
Can, R., Kocaman, S., and Gokceoglu, C. (2021).
\newblock {A comprehensive assessment of XGBOOST algorithm for landslide susceptibility mapping in the upper basin of Ataturk Dam, Turkey}.
\newblock {\em Applied sciences}, 11(11):4993.

\bibitem[Chen and Guestrin, 2016]{Chen_2016}
Chen, T. and Guestrin, C. (2016).
\newblock {XGBoost}: A scalable tree boosting system.
\newblock In {\em Proceedings of the 22nd ACM SIGKDD International Conference on Knowledge Discovery and Data Mining}, pages 785--794. ACM.

\bibitem[Demir and Şahin, 2022]{demir-2022}
Demir, S. and Şahin, E.~K. (2022).
\newblock {Liquefaction prediction with robust machine learning algorithms (SVM, RF, and XGBoost) supported by genetic algorithm-based feature selection and parameter optimization from the perspective of data processing}.
\newblock {\em Environmental Earth Sciences}, 81(18).

\bibitem[Durante and Rathje, 2021]{durante-2021}
Durante, M.~G. and Rathje, E.~M. (2021).
\newblock {An exploration of the use of machine learning to predict lateral spreading}.
\newblock {\em Earthquake Spectra}, 37(4):2288--2314.

\bibitem[Durante and Rathje, 2022]{LateralSpreadingDataset}
Durante, M.~G. and Rathje, E.~M. (2022).
\newblock Machine learning models for the evaluation of the lateral spreading hazard in the avon river area following the 2011 christchurch earthquake.

\bibitem[Geyin et~al., 2022]{geyin_ai_2022}
Geyin, M., Maurer, B.~W., and Christofferson, K. (2022).
\newblock An {AI} driven, mechanistically grounded geospatial liquefaction model for rapid response and scenario planning.
\newblock {\em Soil Dynamics and Earthquake Engineering}, 159:107348.

\bibitem[Hastie and Tibshirani, 1987]{hastie-1987}
Hastie, T. and Tibshirani, R. (1987).
\newblock {Generalized additive Models: some applications}.
\newblock {\em Journal of the American Statistical Association}, 82(398):371--386.

\bibitem[Hsiao et~al., 2024]{hsiao2024explainable}
Hsiao, C.-H., Kumar, K., and Rathje, E.~M. (2024).
\newblock Explainable ai models for predicting liquefaction-induced lateral spreading.
\newblock {\em Frontiers in Built Environment}, 10:1387953.

\bibitem[Huang and Marques-Silva, 2024]{huang_failings_2024}
Huang, X. and Marques-Silva, J. (2024).
\newblock On the failings of shapley values for explainability.
\newblock {\em International Journal of Approximate Reasoning}, 171:109112.

\bibitem[Lundberg and Lee, 2017]{lundberg-2017}
Lundberg, S. and Lee, S.-I. (2017).
\newblock {A unified approach to interpreting model predictions}.
\newblock {\em neural information processing systems}, 30:4768--4777.

\bibitem[Maxwell et~al., 2021]{maxwell-2021}
Maxwell, A.~E., Sharma, M., and Donaldson, K.~A. (2021).
\newblock {Explainable boosting machines for slope failure spatial predictive modeling}.
\newblock {\em Remote sensing}, 13(24):4991.

\bibitem[{New Zealand Geotechnical Database}, 2014]{nzgd-gwd}
{New Zealand Geotechnical Database} (2014).
\newblock Event specific groundwater surface elevations. {Map layer CGD0800 – 12 June 2014}.
\newblock accessed 5 May 2022.

\bibitem[{New Zealand Geotechnical Database}, 2015a]{nzgd-pga}
{New Zealand Geotechnical Database} (2015a).
\newblock Conditional pga for liquefaction assessment. {Map layer CGD5110 – 30 June 2015}.
\newblock accessed 5 May 2022.

\bibitem[{New Zealand Geotechnical Database}, 2015b]{nzgd-dem}
{New Zealand Geotechnical Database} (2015b).
\newblock Lidar and digital elevation models. {Maplayer CGD0500 – 30 June 2015}.
\newblock accessed 5 May 2022.

\bibitem[Nori et~al., 2019]{nori-2019}
Nori, H., Jenkins, S., Koch, P., and Caruana, R. (2019).
\newblock {InterpretML: a unified framework for machine learning interpretability}.
\newblock {\em arXiv (Cornell University)}.

\bibitem[Rateria and Maurer, 2022]{rateria-2022}
Rateria, G. and Maurer, B.~W. (2022).
\newblock Evaluation and updating of ishihara’s (1985) model for liquefaction surface expression, with insights from machine and deep learning.
\newblock {\em Soils and Foundations}, 62(3):101131.

\bibitem[Rathje et~al., 2017a]{RathjeDesignSafe}
Rathje, E.~M., Dawson, C., Padgett, J.~E., Pinelli, J.-P., Stanzione, D., Adair, A., Arduino, P., Brandenberg, S.~J., Cockerill, T., Dey, C., Esteva, M., Haan, F.~L., Hanlon, M., Kareem, A., Lowes, L., Mock, S., and Mosqueda, G. (2017a).
\newblock {DesignSafe: New cyberinfrastructure for Natural Hazards engineering}.
\newblock {\em Natural hazards review}, 18(3).

\bibitem[Rathje et~al., 2017b]{rathje-2017}
Rathje, E.~M., Secara, S.~S., Martin, J.~G., Van~Ballegooy, S., and Russell, J.~K. (2017b).
\newblock {Liquefaction-Induced Horizontal Displacements from the Canterbury Earthquake Sequence in New Zealand Measured from Remote Sensing Techniques}.
\newblock {\em Earthquake Spectra}, 33(4):1475--1494.

\bibitem[Ribeiro et~al., 2016]{ribeiro-2016}
Ribeiro, M.~T., Singh, S., and Guestrin, C. (2016).
\newblock {"Why should I trust you?": explaining the predictions of any classifier}.
\newblock {\em arXiv (Cornell University)}.

\end{thebibliography}
\begin{appendix}

\section{Bivariate Function Maps} \label{sect:appendix-bivariate}

\begin{figure}[H]
    \centering
    \begin{subfigure}[t]{0.3\textwidth}
        \includegraphics[width=\linewidth]{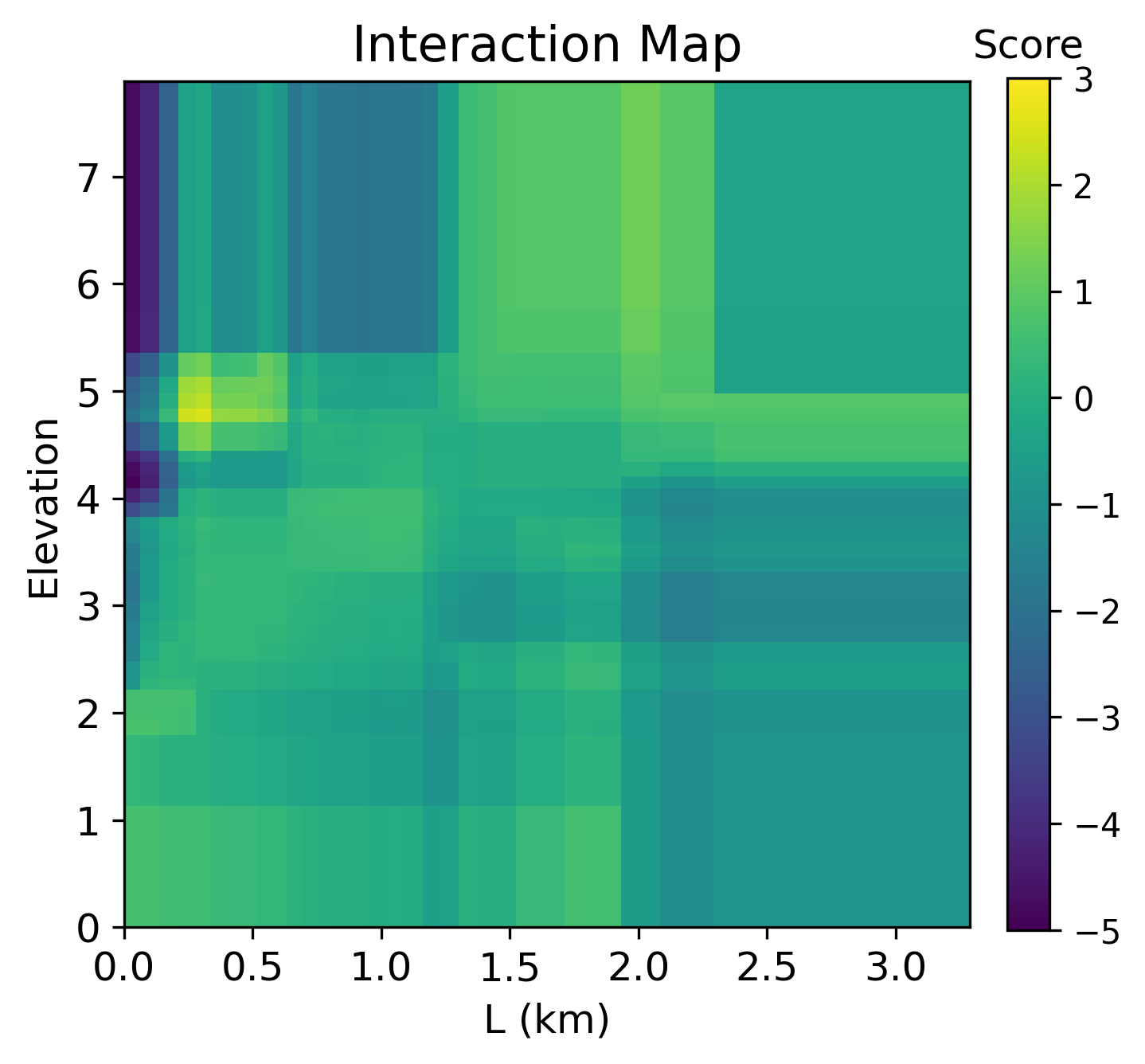}
        \caption{Elevation and L}
    \end{subfigure}
    \hfill
    \begin{subfigure}[t]{0.3\textwidth}
        \includegraphics[width=\linewidth]{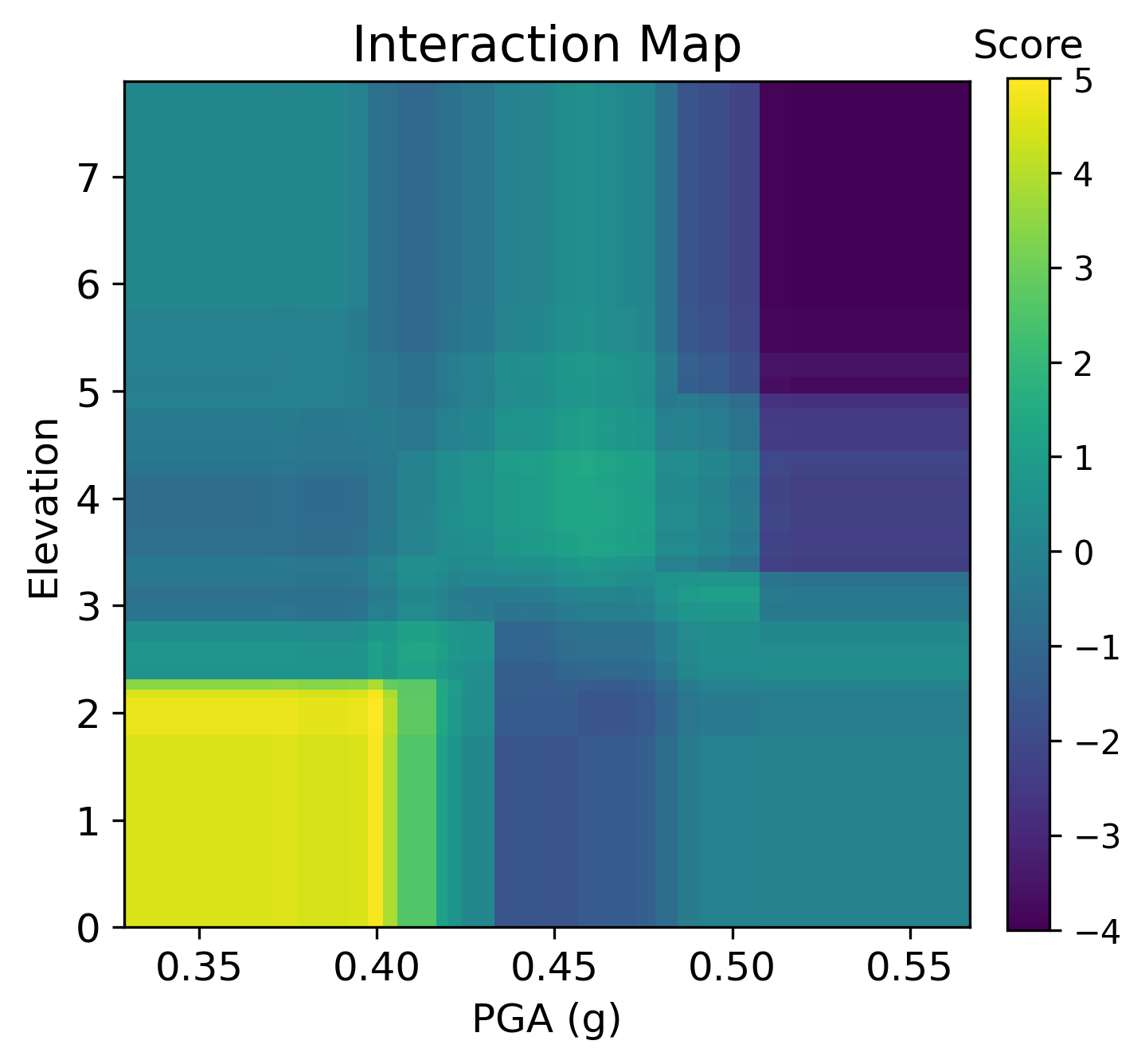}
        \caption{Elevation and PGA}
    \end{subfigure}
    \hfill
    \begin{subfigure}[t]{0.3\textwidth}
        \includegraphics[width=\linewidth]{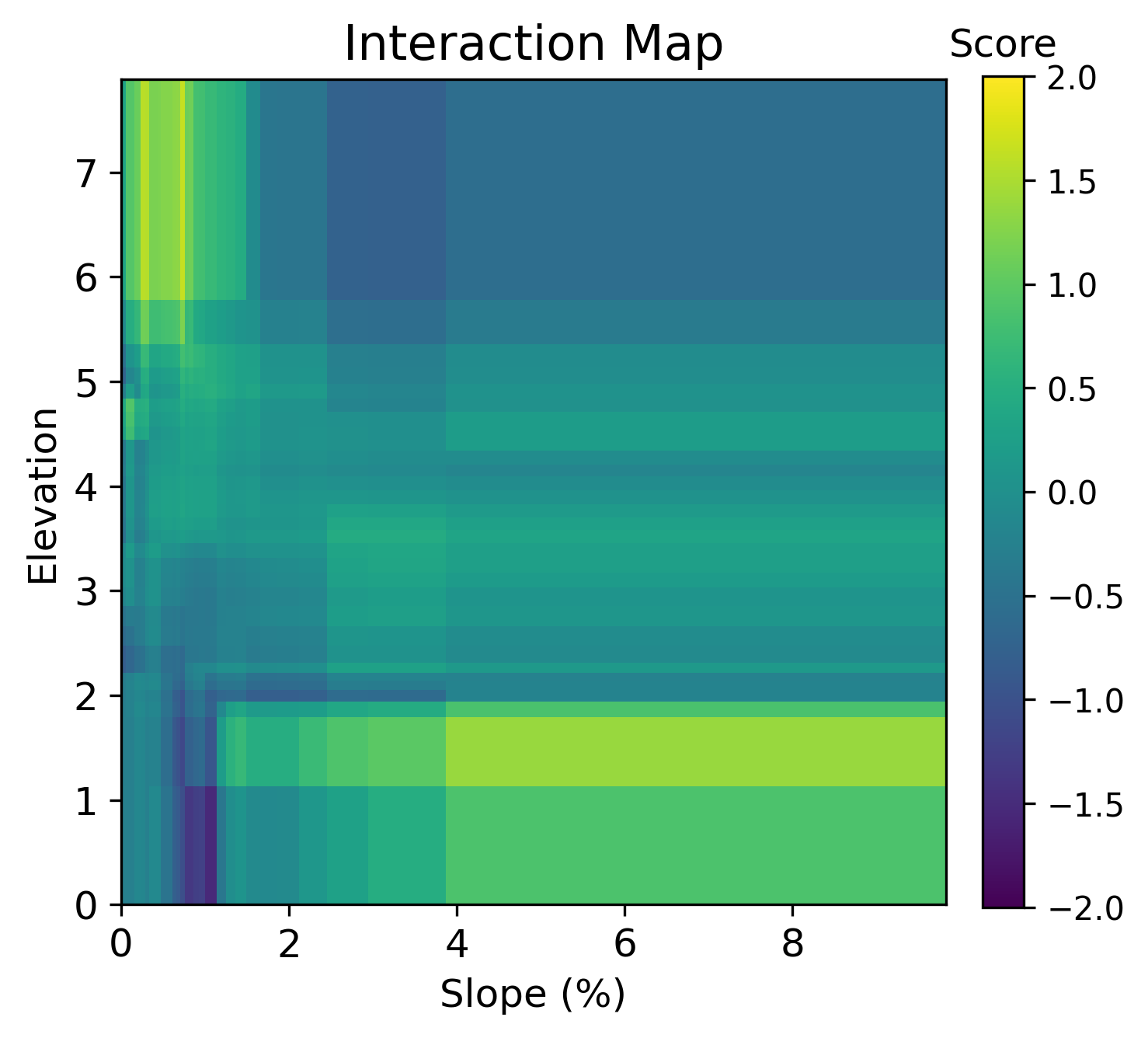}
        \caption{Elevation and slope}
    \end{subfigure}
    \vfill
    \begin{subfigure}[t]{0.3\textwidth}
        \includegraphics[width=\linewidth]{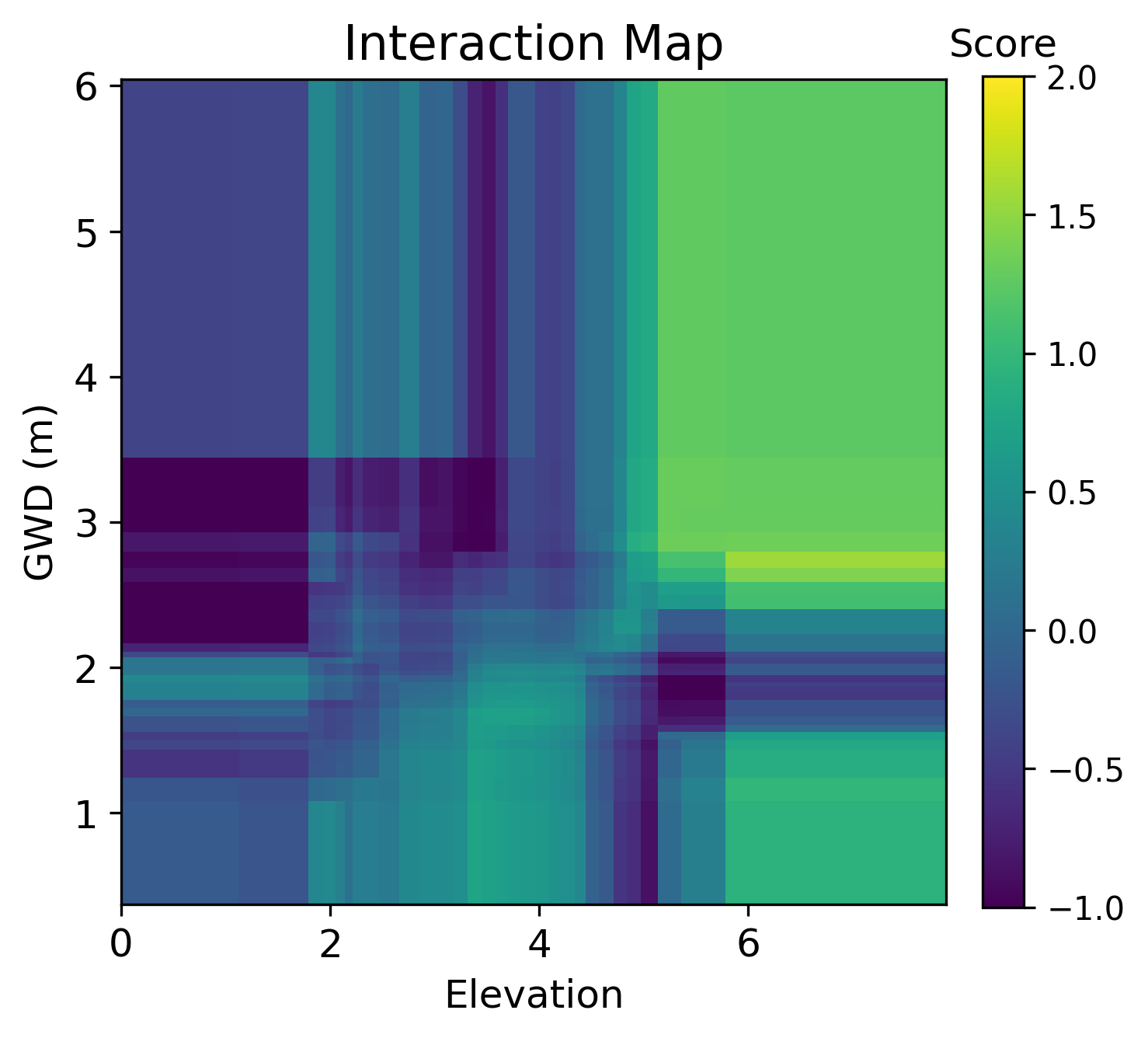}
        \caption{GWD and elevation}
    \end{subfigure}
    \hfill
    \begin{subfigure}[t]{0.3\textwidth}
        \includegraphics[width=\linewidth]{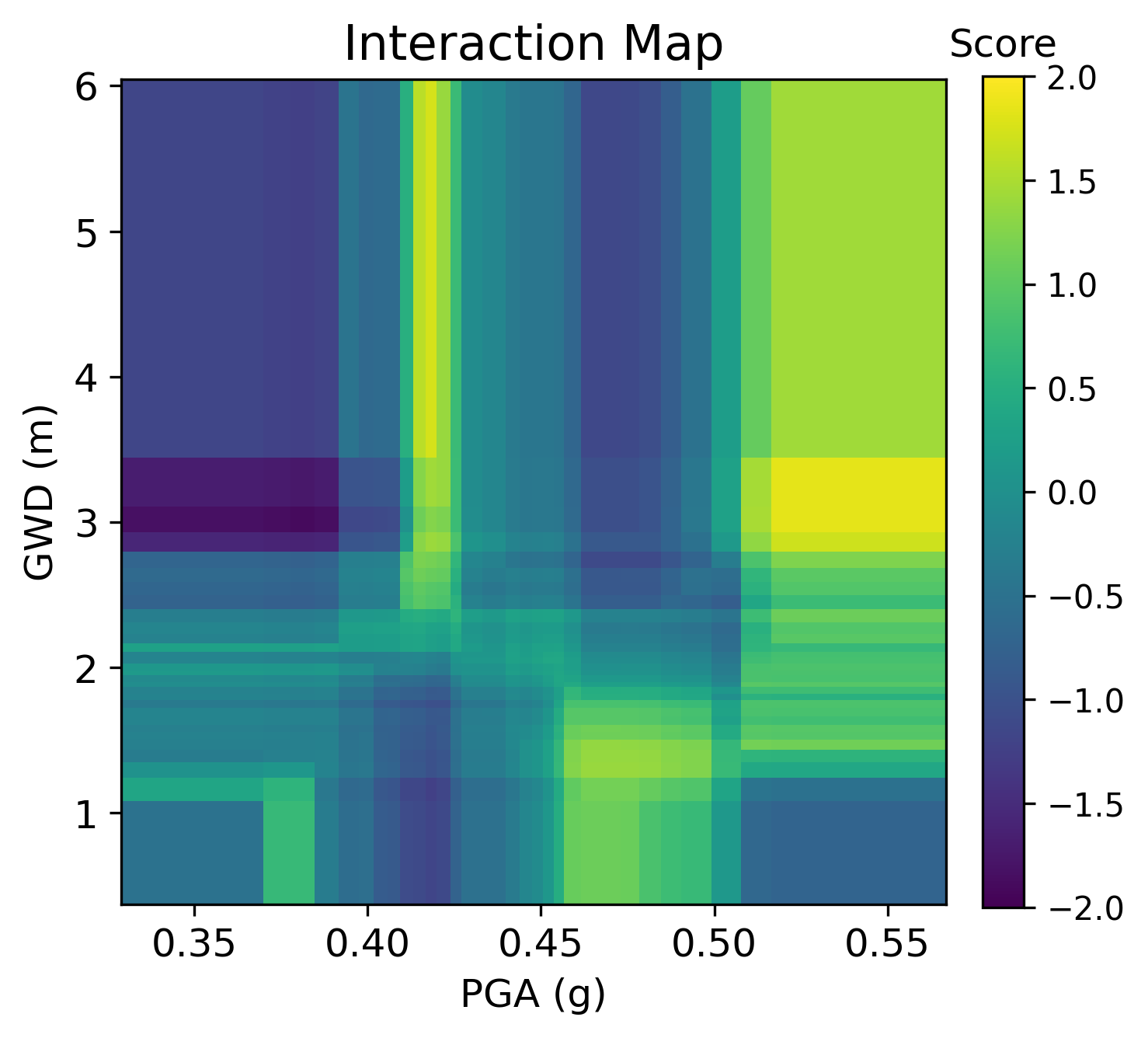}
        \caption{GWD and PGA}
    \end{subfigure}
    \hfill
    \begin{subfigure}[t]{0.3\textwidth}
        \includegraphics[width=\linewidth]{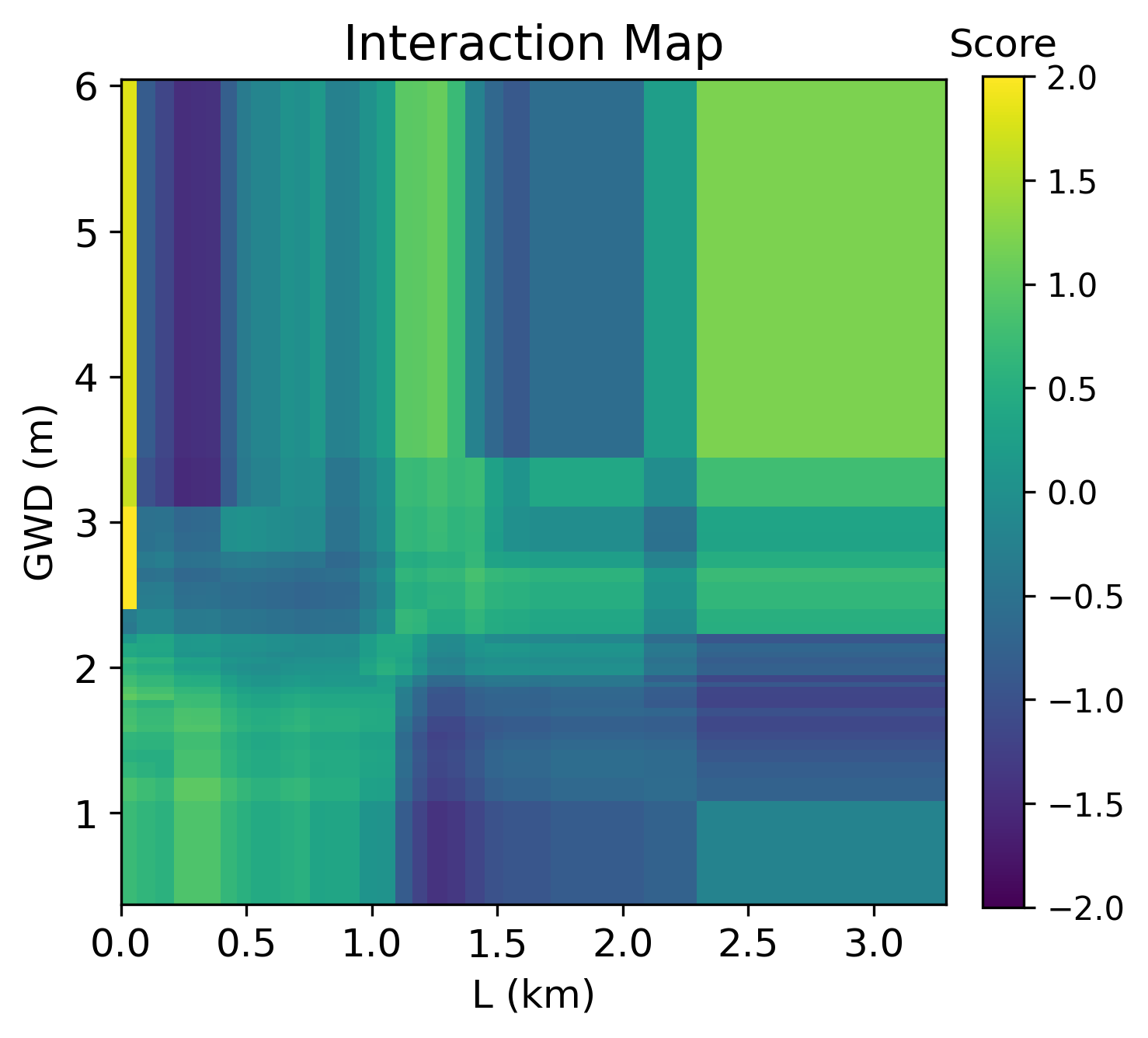}
        \caption{GWD and L}
    \end{subfigure}
    \vfill
    \begin{subfigure}[t]{0.3\textwidth}
        \includegraphics[width=\linewidth]{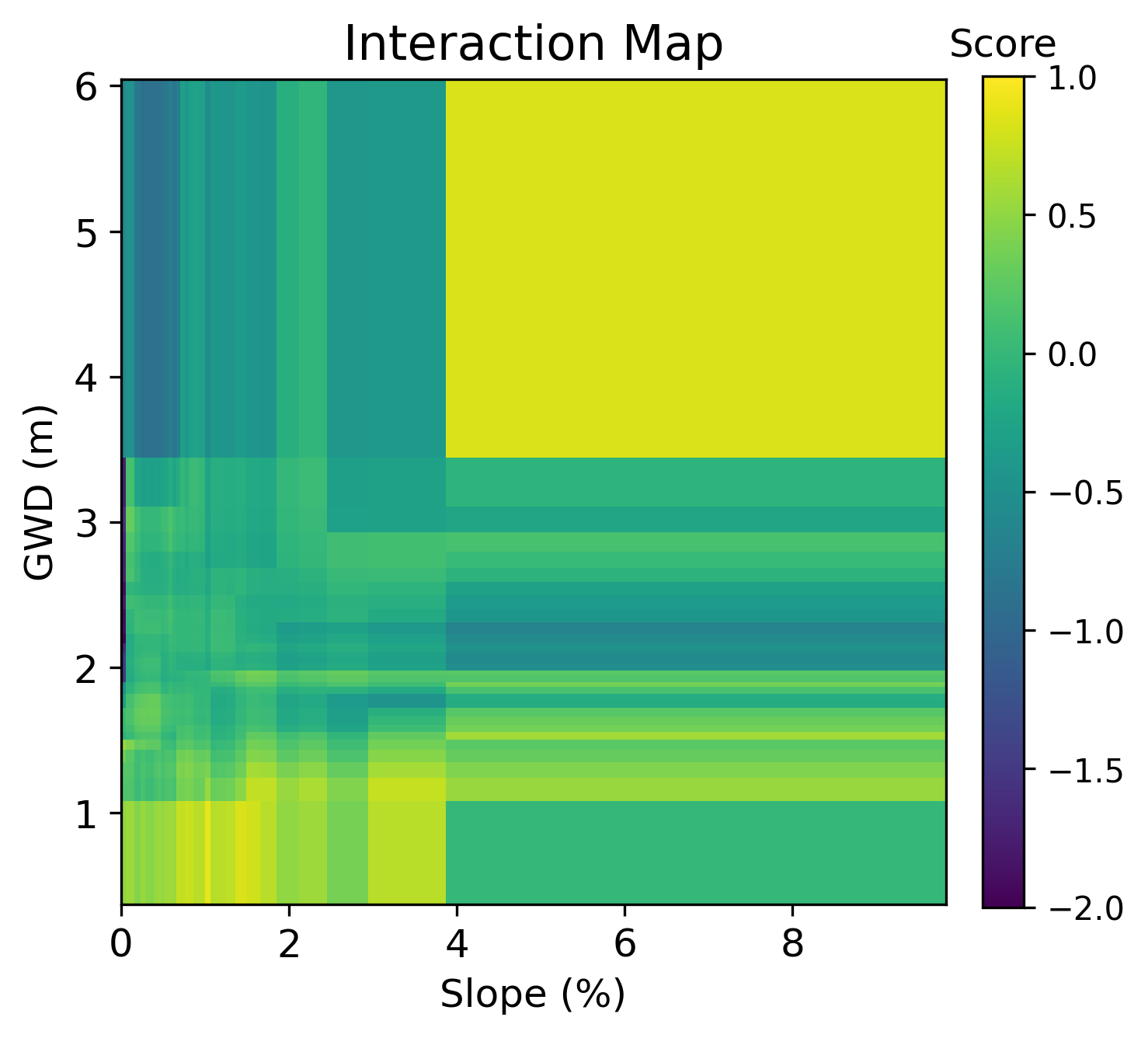}
        \caption{GWD and slope}
    \end{subfigure}
    \hfill
    \begin{subfigure}[t]{0.3\textwidth}
        \includegraphics[width=\linewidth]{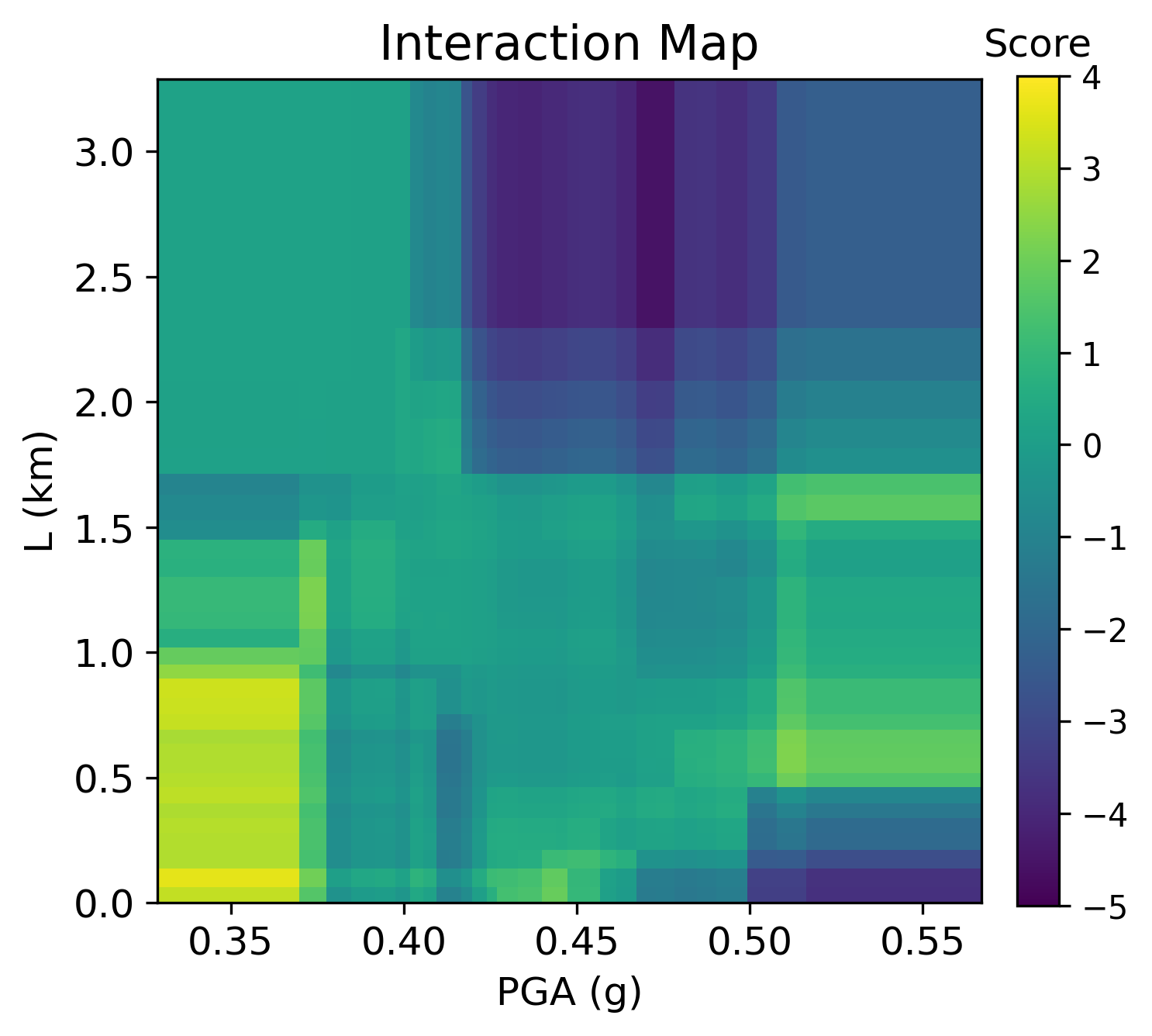}
        \caption{L and PGA}
    \end{subfigure}
    \hfill
    \begin{subfigure}[t]{0.3\textwidth}
        \includegraphics[width=\linewidth]{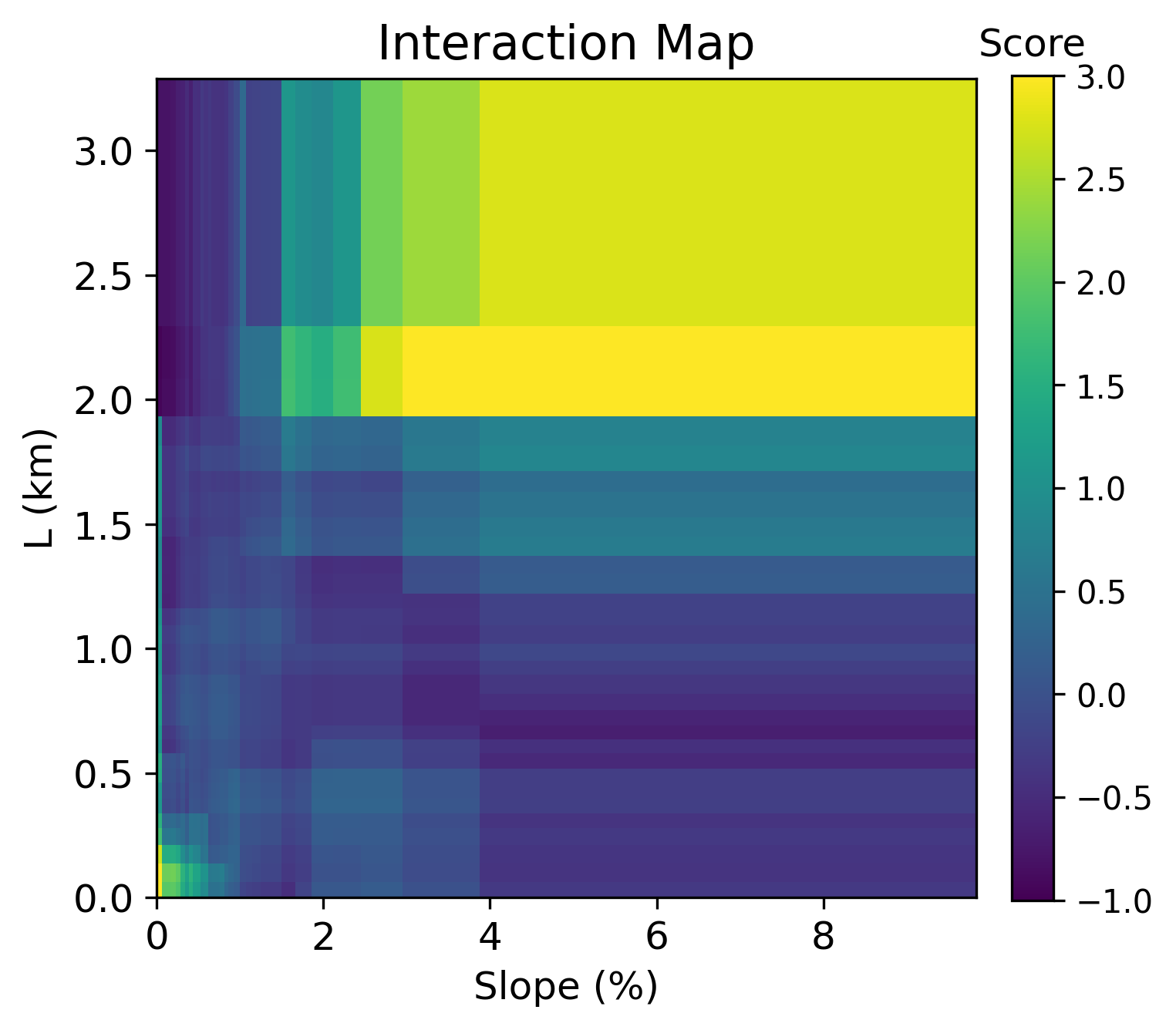}
        \caption{L and slope}
    \end{subfigure}
    \vfill
    \begin{subfigure}[t]{0.3\textwidth}
        \includegraphics[width=\linewidth]{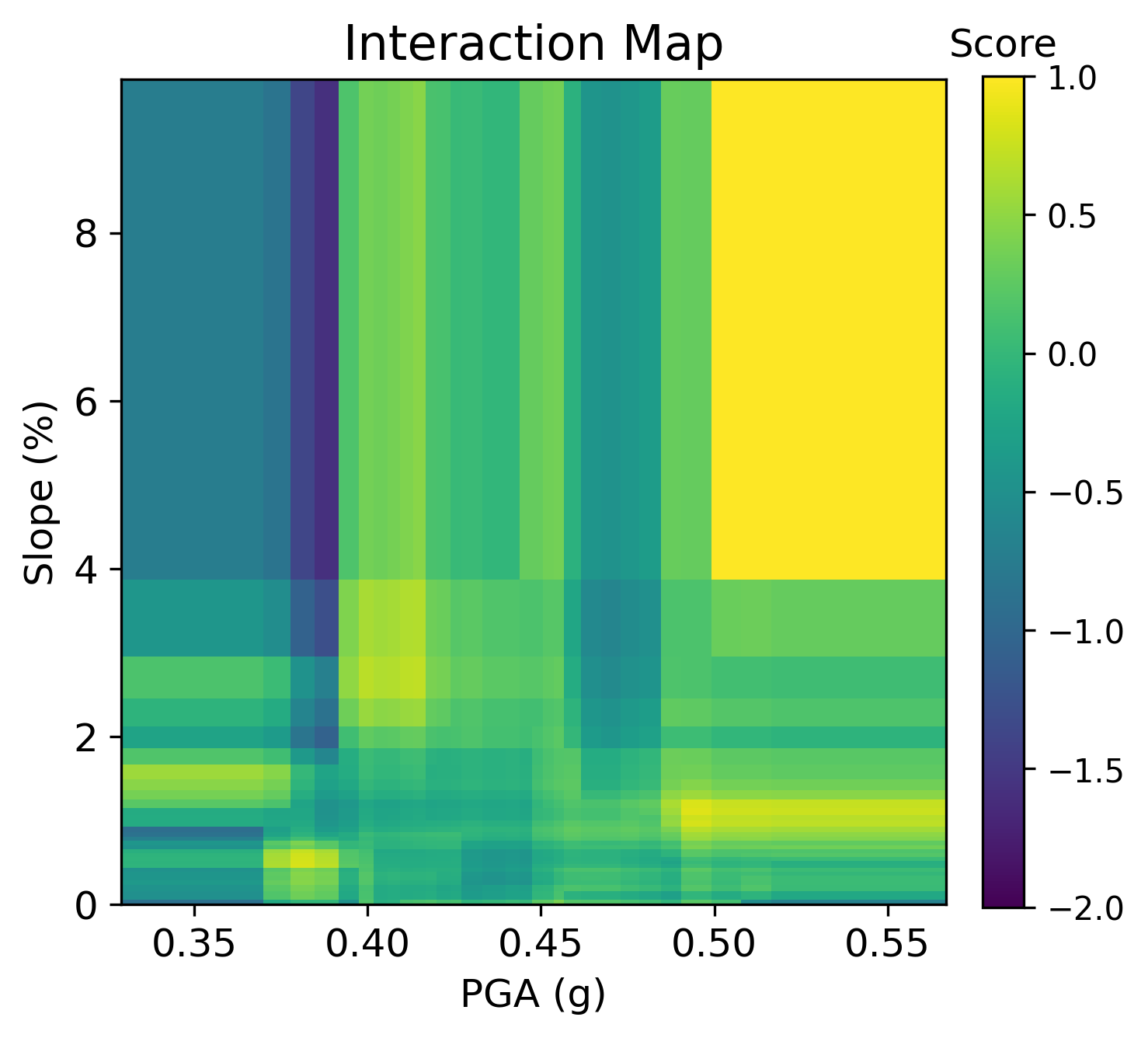}
        \caption{Slope and PGA}
    \end{subfigure}

    \caption{Bivariate function maps from the original EBM.}
    \label{fig:bivariate}

\end{figure}
\end{appendix}

\end{document}